\documentclass[nohyperref]{article}

\usepackage{microtype}
\usepackage{graphicx}
\usepackage{subfigure}
\usepackage{booktabs} 

\PassOptionsToPackage{hyphens}{url}\usepackage{hyperref}

\usepackage[accepted]{icml2022}

\usepackage{amsmath}
\usepackage{amssymb}
\usepackage{mathtools}
\usepackage{amsthm}

\usepackage[capitalize,noabbrev]{cleveref}

\theoremstyle{plain}
\newtheorem{theorem}{Theorem}[section]
\newtheorem{proposition}[theorem]{Proposition}

\theoremstyle{definition}

\theoremstyle{remark}

\usepackage[textsize=tiny]{todonotes}

\icmltitlerunning{XAI for Transformers: Better Explanations through Conservative Propagation}

\begin{document}

\twocolumn[
\icmltitle{XAI for Transformers: Better Explanations through Conservative Propagation}




\begin{icmlauthorlist}
\icmlauthor{Ameen Ali}{1}
\icmlauthor{Thomas Schnake}{2,3}
\icmlauthor{Oliver Eberle}{2,3}
\icmlauthor{Gr\'egoire Montavon}{2,3}
\icmlauthor{Klaus-Robert M\"uller}{2,3,4,5}
\icmlauthor{Lior Wolf}{1}

\end{icmlauthorlist}

\icmlaffiliation{1}{The School of Computer Science, Tel-Aviv University, Israel}
\icmlaffiliation{2}{Machine Learning Group, Technische Universit\"at Berlin, Berlin, Germany}
\icmlaffiliation{3}{BIFOLD -- Berlin Institute for the Foundations of Learning and Data, Berlin, Germany}
\icmlaffiliation{4}{Department of Artificial Intelligence, Korea University, Seoul, Korea}
\icmlaffiliation{5}{Max Planck Institute for Informatics, Saarbr\"ucken, Germany}

\icmlcorrespondingauthor{Ameen Ali}{ameenali@mail.tau.ac.il}

\icmlkeywords{Machine Learning, ICML}

\vskip 0.3in
]



\printAffiliationsAndNotice{}  

\begin{abstract}
    Transformers have become an important workhorse of machine learning, with numerous applications. This necessitates the development of reliable methods for increasing their transparency. Multiple interpretability methods, often based on gradient information, have been proposed. 
We show that the gradient in a Transformer reflects the function only locally, and thus fails to reliably identify the contribution of input features to the prediction. We identify Attention Heads and LayerNorm as the main reasons for such unreliable explanations and propose a more stable way for propagation through these layers. Our proposal, which can be seen as a proper extension of the well-established LRP method to Transformers, is shown both theoretically and empirically to overcome the deficiency of a simple gradient-based approach, and achieves state-of-the-art explanation performance on a broad range of Transformer models and datasets. 
\end{abstract}

\section{Introduction}
Transformer models \cite{vaswani2017attention} have attracted increasing interest and shown excellent performance in domains such as natural language processing (NLP) \cite{vaswani2017attention, devlin2019bert,radford2019language}, vision \cite{dosovitskiy2021image} or graphs \cite{graphormer,yun2019GTN}. Yet, their typically very high complexity (up to billions of parameters \cite{10.1145/3458817.3476209}) makes these models notoriously intransparent and their predictions inaccessible to the user. Since Transformer models have heavy application in potentially sensitive domains, e.g.\ as support in recruiting processes \cite{sussex97276}, the development of methods that explain their decisions is essential since they allow one to verify whether the model makes fair decisions and does not discriminate protected classes \cite{bolukbasi2016man, gonen2019lipstick}. 
In this work, we ask how to bring Explainable AI to Transformers in a theoretically sound manner, focusing in particular on the axiom of conservation, which underlies a number of popular explanation techniques \cite{bach-plos15,DBLP:conf/nips/LundbergL17,DBLP:conf/icml/SundararajanTY17}. We embed our analysis in the Layer-wise Relevance Propagation (LRP) framework \cite{bach-plos15}, which allows us to analyze conservation at the level of individual modules and layers of the Transformer model. Our analysis reveals that the conservation properties of existing XAI techniques can be severely impaired when extended to Transformer models. We propose a theoretically grounded, and easy to implement countermeasure, consisting of strategically `detaching' part of the Transformer model's forward computation before computing explanations.

We provide qualitative and quantitative experiments on different data domains, such as natural language understanding, computer vision and graph analysis. 
Comparison is performed with state-of-the-art baseline explanation methods for Transformers. For the quantitative analysis, we perform different input perturbation schemes, in which we track the behavior of the model when relevant or irrelevant features are added to or removed from the input data sample.
The proposed explanation method for Transformers presents excellent performance on the qualitative and quantitative experiments

and outperforms all existing baseline methods in most of the tasks. Hence, our fairly intuitive explanation method provides both theoretical soundness in its formal derivation and leading performance on many important data domains.  Our code is publicly available.\footnote{\url{https://github.com/AmeenAli/XAI_Transformers}}

\section{Related Work}

The approaches for Transformer explanation can be divided into (1) methods that extract the attention heads from each Transformer block, using them for interpretation, (2) gradient-based methods, and (3) perturbation-based methods. We would like to mention a few of these previous works.

The extraction of the attention heads from an attention module, to understand or visualize the prediction strategy of the model, has a long history \cite{bahdanau2015nmtranslation}, also for non-Transformer architectures, yet it is known to be myopic and unreliable \cite{jain-wallace-2019-attention}. Besides the simple extraction of the raw attention weights, there have been attempts to use the attention heads for defining more elaborate explanation mechanisms, such as `Attention Rollout' and `Attention Flow' \cite{abnar-zuidema-2020-quantifying}.

Other authors incorporate gradient methods to explain Transformer models, such as integrated gradient \cite{wallace-etal-2019-allennlp} or input gradients \cite{atanasova-etal-2020-diagnostic}. The gradient methods Saliency, Gradient$\,\times\,$Input  (GI) or Guided Backpropagation  has already been applied in numerous models and domains, found their use in Transformers, as well \citet{atanasova-etal-2020-diagnostic}. 

In particular, there have been multiple attempts to implement the layer-wise relevance propagation (LRP) method \cite{bach-plos15} in Transformers \cite{voita-etal-2019-analyzing} and other attention-based models \cite{ding-etal-2017-visualizing}. In addition, LRP has been applied to explain predictions of other models on NLP tasks \cite{wu-ong-2020-explain}, such as BERT \cite{devlin2019bert}. Other approaches to LRP / gradient propagation in Transformer blocks can be found in  \cite{Chefer_2021_ICCV,chefer2021transformer}, where the relevancy scores are obtained by combining attention scores with LRP or attention gradients. 

%
There are also a few examples in which perturbation-based methods used input reductions \cite{feng-etal-2018-pathologies, DBLP:conf/emnlp/PrabhakaranHM19} to determine the most relevant parts of the input, by observing change in model confidence or Shapley values \cite{DBLP:conf/nips/LundbergL17,atanasova-etal-2020-diagnostic}. 

The experiments we conduct employ both the conventional Transformer, as well as a Transformers on graphs. Specifically, we employ the recent Graphormer model of~\citet{graphormer}. Other graph Transformers include \cite{yun2019GTN,yoo2020graph,maziarka2020molecule,zhao2021gophormer}. A simple explainability procedure based on the obtained attention maps is presented by \citet{yun2019GTN}.

Previous works on explainability in graph neural networks include gradient-based methods \cite{pope2019explainability}, graph decomposition \cite{schnake2020higher} graph perturbations \cite{ying2019gnnexplainer,luo2020parameterized}, and a local approximation using simpler models \cite{huang2020graphlime}.

For a more general review of explainable AI, we refer to~\cite{
DBLP:journals/dsp/MontavonSM18,DBLP:journals/csur/GuidottiMRTGP19,DBLP:series/lncs/11700,DBLP:conf/ijcnlp/DanilevskyQAKKS20,
BARREDOARRIETA202082,samek2021xaireview}.

\section{A Theoretical View on Explaining Transformers} 
\label{section:theory}

Very recently, techniques such as Gradient$\,\times\,$Input (GI) \cite{DBLP:journals/corr/ShrikumarGSK16,DBLP:series/lncs/AnconaCOG19} or Layer-wise Relevance Propagation (LRP) \cite{bach-plos15} have been extended to explain Transformers \cite{voita-etal-2019-analyzing,wu-ong-2020-explain}. New explanation techniques, such as attention rollouts \cite{abnar-zuidema-2020-quantifying}, or other generic ways to aggregate attention information~\cite{Chefer_2021_ICCV,chefer2021transformer} have also been developed for a similar purpose. These recent works have provided empirical evidence that Transformers can be made explainable to a significant extent.

In this section, we aim to gain a better theoretical understanding of the problem of explaining Transformers, drawing on the same `axiomatic approach' \cite{Shapley1953,DBLP:journals/jmlr/StrumbeljK10,DBLP:conf/icml/SundararajanTY17,DBLP:series/lncs/Montavon19,hesse2021fast} used for analyzing and developing Explainable AI in the context of standard deep neural networks. A main axiom employed in Explainable AI, especially for the task of attribution, is \textit{conservation}. The conservation axiom (also known as \textit{completeness} or \textit{efficiency}) states that scores assigned to input variables and forming the explanation must sum to the output of the network, in other words, each input variable contributes a share of the predicted score at the output. Workhorses of XAI, such as Layer-wise Relevance Propagation \cite{bach-plos15}, Gradient$\,\times\,$Input \cite{DBLP:journals/corr/ShrikumarGSK16, DBLP:conf/nips/SrinivasF19, hesse2021fast}), Integrated Gradients \cite{DBLP:conf/icml/SundararajanTY17} or Shapley Values \cite{DBLP:journals/jmlr/StrumbeljK10, DBLP:conf/nips/LundbergL17}, are either designed to satisfy this conservation axiom, have been shown to satisfy it for certain models, or can be derived from it directly.

The Layer-wise Relevance Propagation framework \cite{bach-plos15} considers a particularly strong form of conservation, where each layer, component, or even neuron in the network is subject to the conservation axiom. In particular, the `relevance' received by a given layer/component/neuron from the layer above must be fully redistributed  to the layer below. Many relevance propagation rules have been developed within the LRP framework to address the specificities of different data and architectures. Interestingly, the rudimentary but popular Gradient$\,\times\,$Input method, although commonly viewed as a gradient-based method, can also be embedded in the LRP framework \cite{DBLP:journals/corr/ShrikumarGSK16, DBLP:series/lncs/AnconaCOG19, DBLP:series/lncs/Montavon19}, and potentially be used as a starting point (or default choice) for setting propagation rules. Furthermore, with this embedding into LRP, one can perform a detailed analysis of the GI explanation procedure, identifying layers or components where the conservation breaks, and possibly derive better propagation rules for these layers in a second step.

\begin{figure*}
\centering
\begin{tabular}{c|c}
(a) & (b)\\
\includegraphics[scale=.7]{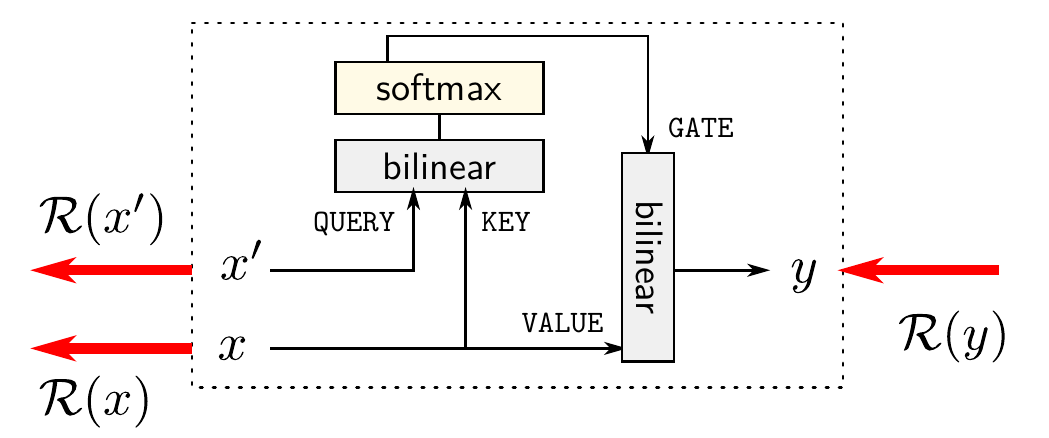} &
\includegraphics[scale=.7]{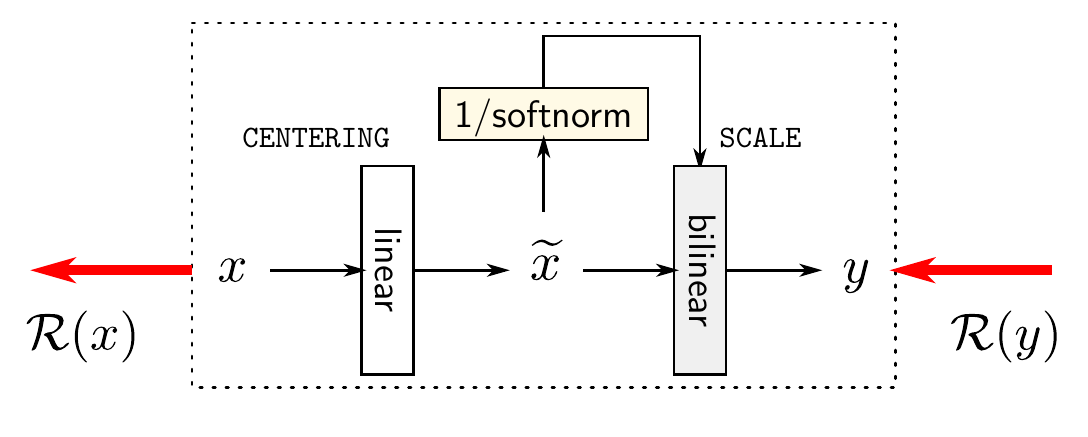} 
\end{tabular}
\vspace{-1mm}
\caption{Diagram of specific modules found in Transformer architectures: (a) attention head, (b) LayerNorm. Modules are depicted with the types of nonlinearities they incorporate (bilinear form, softmax, division by the norm), and annotated with the variables used in this paper. Red arrows denote the relevance propagation process under study.}
\end{figure*}

Denote by $(x_i)_i$ and $(y_j)_j$ the vectors of neurons representing the input and output of some layer or component of interest in a neural network, and by $f$ the output of the neural network. Gradient$\,\times\,$Input attributions on these two vector representations can be computed as:
\begin{align}
\mathcal{R}(x_i) &= x_i \cdot ({\partial f}/{\partial x_i}) \label{eq:Rx}\\ \mathcal{R}(y_j) &= y_j \cdot ({\partial f}/{\partial y_j})\label{eq:Ry}
\end{align}
respectively. Recall that the gradients at different layers of a neural network are related via the chain rule as follows:
\begin{align}
\frac{\partial f}{\partial x_i} = \sum_j \frac{\partial y_j}{\partial x_i} \frac{\partial f}{\partial y_j} 
\label{eq:gradprop}
\end{align}
Injecting \eqref{eq:Rx} and \eqref{eq:Ry} into \eqref{eq:gradprop}, the gradient propagation rule can be converted into an equivalent relevance propagation rule:
\begin{align}
\mathcal{R}(x_i) =  \sum_j \frac{\partial y_j}{\partial x_i} \frac{x_i}{y_j} \mathcal{R}(y_j)\,,
\label{eq:girule}
\end{align}
using the convention $0/0=0$. This is the LRP view on GI. With such interpretation, we can study whether GI is conservative layer-wise or at the level of the given component, by testing whether $\sum_i \mathcal{R}(x_i) = \sum_j \mathcal{R}(y_j)$. If this holds for every component of the neural network, then conservation also holds globally. We next show that two components of the Transformer cause a significant break of conservation and therefore require an improved propagation rule.

\subsection{Propagation in Attention Heads}
\label{section:attention}

Let us consider the attention head, which makes use of a Query-Key-Value mechanism, and which is a core component of Transformers. As defined in \cite{vaswani2017attention}, attention heads have the structure
\begin{align}
Y = \text{softmax}\Big(\frac{1}{\sqrt{d_K}}(X' W_Q) (X W_K)^\top\Big) X
\label{eq:attention-original}
\end{align}
where $X = (x_i)_i$ and $X' = (x'_j)_j$ are the input sequences of token embeddings, $Y = (y_j)_j$ is the sequence of output embeddings, $d_K$ 
is the dimensionality of the Key-vector,
and $W_{\{Q,K,V\}}$ are learned projection matrices. In the equation above, we omitted the multiplication by the embedding $W_V$, since the latter can be viewed as a subsequent linear layer and can therefore be treated separately. For the purposes of our analysis, we rewrite Eq.\ \eqref{eq:attention-original} as:
\begin{align}
y_j = \sum_{i} x_i p_{ij}
\label{eq:attention}
\end{align}
where $$p_{ij} = \frac{\exp(q_{ij})}{\sum_{i'} \exp(q_{i'j})}$$ is the softmax computation and $q_{ij} = \frac{1}{\sqrt{d_K}}x_i^\top W_K W_Q^\top \,x'_j$ is the matching function between tokens of the two input sequences. Note that the output of Eq.\ \eqref{eq:attention} depends on the input tokens both explicitly, via the term $x_i$, and via the gating term $p_{ij}$, which itself depends on $x$ and $x'$. Also, we observe that for each index $j$, we have an associated distribution $p_j = (p_{ij})_i$, and we denote by $\mathbb{E}_j[\cdot]$ and $\text{Cov}_j(\cdot,\cdot)$, respectively, the expectation and covariance over this distribution, e.g.\ $\mathbb{E}_j[x] = \sum_i x_i p_{ij}$.

We analyze the relevance propagation associated with applying Gradient$\,\times\,$Input to the Transformer model. For this, we define input token relevance as $\mathcal{R}(x_i) = x_i^\top (\partial f / \partial x_i)$ and $\mathcal{R}(x_j') ={x_j'}^{\top} (\partial f / \partial x_j')$, and output token relevance as $\mathcal{R}(y_j) = y_j^\top (\partial f / \partial y_j)$.
\begin{proposition}
\label{prop:1}
Assume that the inputs to the attention head and its softmax gate are centered, i.e.\ $\mathbb{E}_j[q_{:j}] = 0$ and $\mathbb{E}_j[x] = 0$ under all probability distributions $p_j$. The relevance propagation associated with Gradient$\,\times\,$Input is characterized by the conservation equation:
\begin{align}
&\sum_i \mathcal{R}(x_i) + \sum_j \mathcal{R}(x'_j)\nonumber\\
&\qquad = \sum_j \mathcal{R}(y_j) + 2 \sum_j \mathrm{Cov}_j \big(q_{:j}, x\big)^\top \frac{\partial f}{\partial y_j}
\end{align}
This implies that conservation breaks if the rightmost term is non-zero.
\end{proposition}
The proof can be found in Appendix \ref{appendix:ah}. Prop~\ref{prop:1} implies that conservation between layers may not hold in the presence of covariates between $q_{:j}$ and $x$, something that is likely to occur, since the former is a function of the latter. Note that the centering assumption we made in the Proposition above mainly serves to arrive at a simple and intuitive conservation equation. We provide the more general equation in Appendix~\ref{appendix:ah}.

To address the lack of conservation in Gradient$\,\times\,$Input (and the resulting consequence that some attention heads may be over- or underrepresented in the explanation), we will propose in Section \ref{section:lrp} an alternative propagation rule that retains the conservation property and also works better empirically.

\subsection{Propagation in LayerNorm}\label{sect:prop_layern}

\begin{figure}[h]
    \centering
     \hfill
    \includegraphics[width=.45\linewidth]{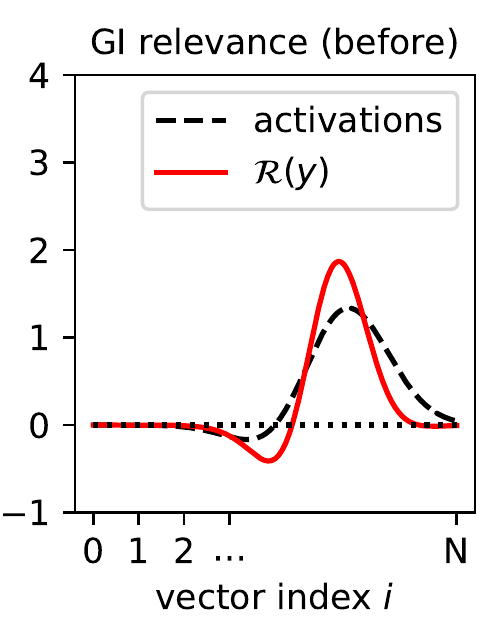} \hfill \includegraphics[width=.45\linewidth]{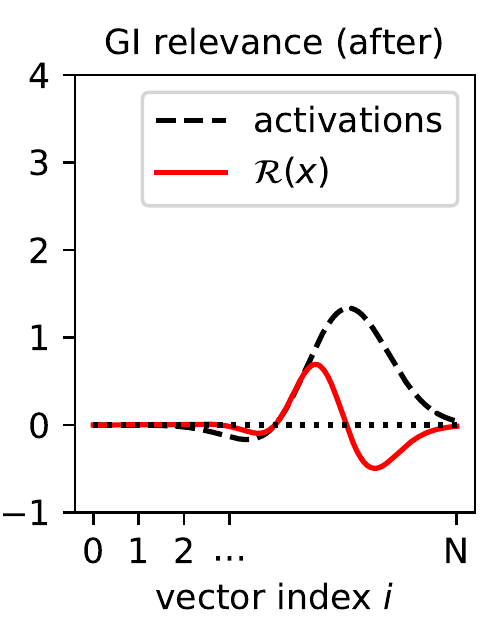}
     \hfill \vspace{-2mm}
    \caption{Comparison of the relevance associated with Gradient$\,\times\,$Input before and after propagation in LayerNorm. We can observe a `relevance collapse' (visible on the right of the peak activation).}
    \label{fig:normalization}
\end{figure}

Another layer that is rather specific to the Transformer is `LayerNorm'. For our analysis, we focus on the core part of LayerNorm, consisting of centering and standardization (and ignore the subsequent affine transformation):
\begin{align}
y_i &= \frac{x_i - \mathbb{E}[x]}{\sqrt{\epsilon + \text{Var}[x]}} \label{eq:layernorm}
\end{align}
Here, $\mathbb{E}[\cdot]$ and $\text{Var}[\cdot]$ denote the mean and variance over all activations of the corresponding channel (and potentially minibatch).
\begin{proposition}
The relevance propagation associated with Gradient$\,\times\,$Input is characterized by the conservation equation:
\begin{align}
\sum_i \mathcal{R}(x_i) &=
\Big(1 - \frac{\mathrm{Var}[x]}{\epsilon + \mathrm{Var}[x]}\Big) \sum_i \mathcal{R}(y_i)
\label{eq:naive}
\end{align}
\end{proposition}
The proof is given in Appendix \ref{appendix:ln}. We observe that conservation is never satisfied. Conservation breaks especially strongly when $\epsilon$ is small compared to $\mathrm{Var}[x]$. This relevance collapse can be observed in Fig.\ \ref{fig:normalization}.

\section{Better LRP Rules for Transformers}\label{section:lrp}

We propose to address the deficiencies of Gradient$\,\times\,$Input revealed by our layer-wise analysis. Specifically, we take the LRP view on Gradient$\,\times\,$Input as a starting point (Eq.\ \eqref{eq:girule}) and replace the implicit propagation rules in attention heads and LayerNorm (two components we identified as breaking conservation) by ad-hoc propagation rules that are conservative by design.

Specifically, we propose at explanation time to make a locally linear expansion of the \textit{attention head} by viewing the gating terms $p_{ij}$ as constants. Consequently, these terms can be interpreted as the weights of a linear layer locally mapping the input sequence $x$ to the output sequence $y$. We can then use the canonical LRP rule for linear layers defined as
\begin{align*}
\mathcal{R}(x_i) &= \displaystyle \sum_j \frac{x_i p_{ij}}{\sum_{i'} x_{i'} p_{i'j}} \mathcal{R}(y_j)
\qquad \text{(AH-rule)}
\end{align*}
to propagate the relevance scores from the layer output to the layer input. With such a reformulation, we also note that the query sequence $x'$ appears disconnected, and consequently, we have implicitly $\mathcal{R}(x'_j) = 0$. This strategy has also been used and justified theoretically for LSTM blocks \cite{arras-lncs19}, and has been shown empirically to yield superior performance to gradient-based methods, in particular Gradient$\,\times\,$Input.

\medskip

Furthermore, to address the particularly severe break of conservation in \textit{LayerNorm}, we propose again a locally linear expansion at explanation time, by viewing the multiplicative factor $\alpha = (\sqrt{\epsilon + \mathrm{Var}[x]})^{-1}$ as constant. The entire LayerNorm operation can then be expressed by the linear transformation $\alpha C x$ where $C = I_N - \frac1N \boldsymbol{1}_{(N \times N)}$ is the `centering matrix' and where $N$ is the number of tokens in the input sequence. Using again the same canonical LRP rule for linear layers, this time with weights $\alpha C$, we obtain:
\begin{align*}
\mathcal{R}(x_i) &=  \sum_j\frac{x_i \cdot (\delta_{ij} - \frac1N)}{\sum_{i'} x_{i'} \cdot (\delta_{i'j} - \frac1N)} \mathcal{R}(y_j) \qquad  \text{(LN-rule)}
\end{align*}
where $(\delta_{ij} - \frac1N)$ is another way of writing $C_{ij}$ and the factor $\alpha$ present in the numerator and denominator cancels out.

In practice, these rules do not need to be implemented explicitly. We can simply observe that they are effectively the same rules as those induced by Gradient$\,\times\,$Input, with the gating and rescaling terms in their respective layers treated as constant (i.e.\ detached in the forward computation so that the gradient does not propagate through it). We therefore propose the following implementation trick for the rules above:

\begin{center}
\fbox{
\begin{minipage}{.925\linewidth}
\textbf{Implementation Trick:~}
To compute the proposed improved LRP explanation, rewrite Eq.\ \eqref{eq:attention} as
\begin{align*}
y_j = \sum_i x_i [p_{ij}]_\texttt{.detach()}
\end{align*}
in every attention head, and rewrite Eq.\ \eqref{eq:layernorm} as
$$
y_i = \frac{x_i - \mathbb{E}[x]}{\big[\sqrt{\epsilon + \text{Var}[x]}\big]_\texttt{.detach()}}
$$
in every LayerNorm. Then extract the LRP explanation by calling Gradient$\,\times\,$Input on the resulting function $f$.
\end{minipage}
}
\end{center}

This trick makes the method straightforward to implement, as it simply consists of adding \texttt{detach()} calls at the appropriate locations in the neural network code, and then running standard Gradient$\,\times\,$Input. Furthermore, the runtime is at least as good as Gradient$\,\times\,$Input, and typically better, due to the resulting simplification of the gradient computation.

\section{Experimental Setup}

The proposed approach is tested on several Transformer models trained on various datasets. We benchmark the performance of our method against a number of other approaches proposed in the literature and commonly used for explaining Transformer-type architectures.

\subsection{Datasets}

We use the following datasets from natural language processing, image classification, as well as molecular modeling to evaluate the different XAI approaches.
For the NLP experiments, we consider sentiment classification on the SST-2 \cite{socher2013-recursive} and IMDB datasets \cite{maas-EtAl:2011:ACL-HLT2011}  which contain 11,844 and 50,000 movie reviews, respectively, for binary classification into negative or positive sentiment. In addition, we use the TweetEval Dataset \cite{xiong2019tweetqa} for tweet classification on sentiment (59,899), hate detection (12,970) and emotion recognition (5,052). Furthermore, the SILICONE Dataset \cite{chapuis-etal-2020-hierarchical} is also used for emotion detection tasks (Semaine 13,708) and utterance sentiment analysis (Meld-S 5,627). 

For experiments with graph Transformers, the MNIST superpixels data \cite{monti2017geometric} is extracted from 70,000 samples for digit classification. Each node represents one image patch (superpixel), and is connected to its neighbors. Another graph dataset is based on the MoleculeNet \cite{wu2018moleculenet} benchmark: the BACE dataset contains 1,522 compounds, together with their structures and binary labels indicating a binding result for a set of inhibitors of human $\beta$-secretase 1 (BACE-1)\cite{subramanian2016computational}.

\subsection{Benchmark Methods}

First, we compare to a  `Gradient$\times$Input' \cite{Denil2014ExtractionOS, DBLP:journals/corr/ShrikumarGSK16, atanasova-etal-2020-diagnostic} baseline without considering any modifications to the individual gradient computations as described in Section \ref{section:theory}.

In addition, we compute averages over last-layer attention head vectors (`Attention-last')  as proposed in \citet{hollenstein-beinborn-2021-relative} as well as `Rollout' and attention flow (`A-flow') \cite{abnar-zuidema-2020-quantifying} which capture the layer-wise structure of deep Transformer models in comparison to raw attention head analysis.  

`Generic Attention Explainability' (GAE) by \citet{Chefer_2021_ICCV} propagates attention gradients together with gradients from other parts of the network, resulting in state-of-the art performance in explaining Transformer architectures. 

We consider three variants of our proposed LRP-based technique. First `LRP (AH)' where propagation through attention heads is handled via the AH-rule described in Section \ref{section:lrp}. For any other layers, we use the GI-equivalent propagation rule (i.e.\ in practice simply propagating gradient without detaching terms). Then `LRP (LN)' where propagation through LayerNorm is handled via the LN-rule. Lastly `LRP (AH+LN)' which applies both the AH-Rule and LN-Rule in respective layers.

 \section{Results}
We now turn to evaluating the LRP-based method we have proposed. We first validate that conservation is indeed maintained, and then continue with quantitative perturbation experiments.  Finally, we visualize the different explanations produced by these methods qualitatively.

 \subsection{Conservation}
 As a starting point, we would like to verify whether the insights produced in Section \ref{section:theory} on the conservation, or lack thereof, also hold empirically. We consider for this our two Transformer models trained on the SST-2 and MNIST datasets, and compute GI and LRP (AH+LN) explanations. We then compare the score produced at the output of the Transformer network against the sum of explanation scores over input features of the network. For this purpose, the input features are the positionally encoded embedding vectors present in the first layer.

\begin{figure}[h]
    \centering
    \includegraphics[width=\linewidth]{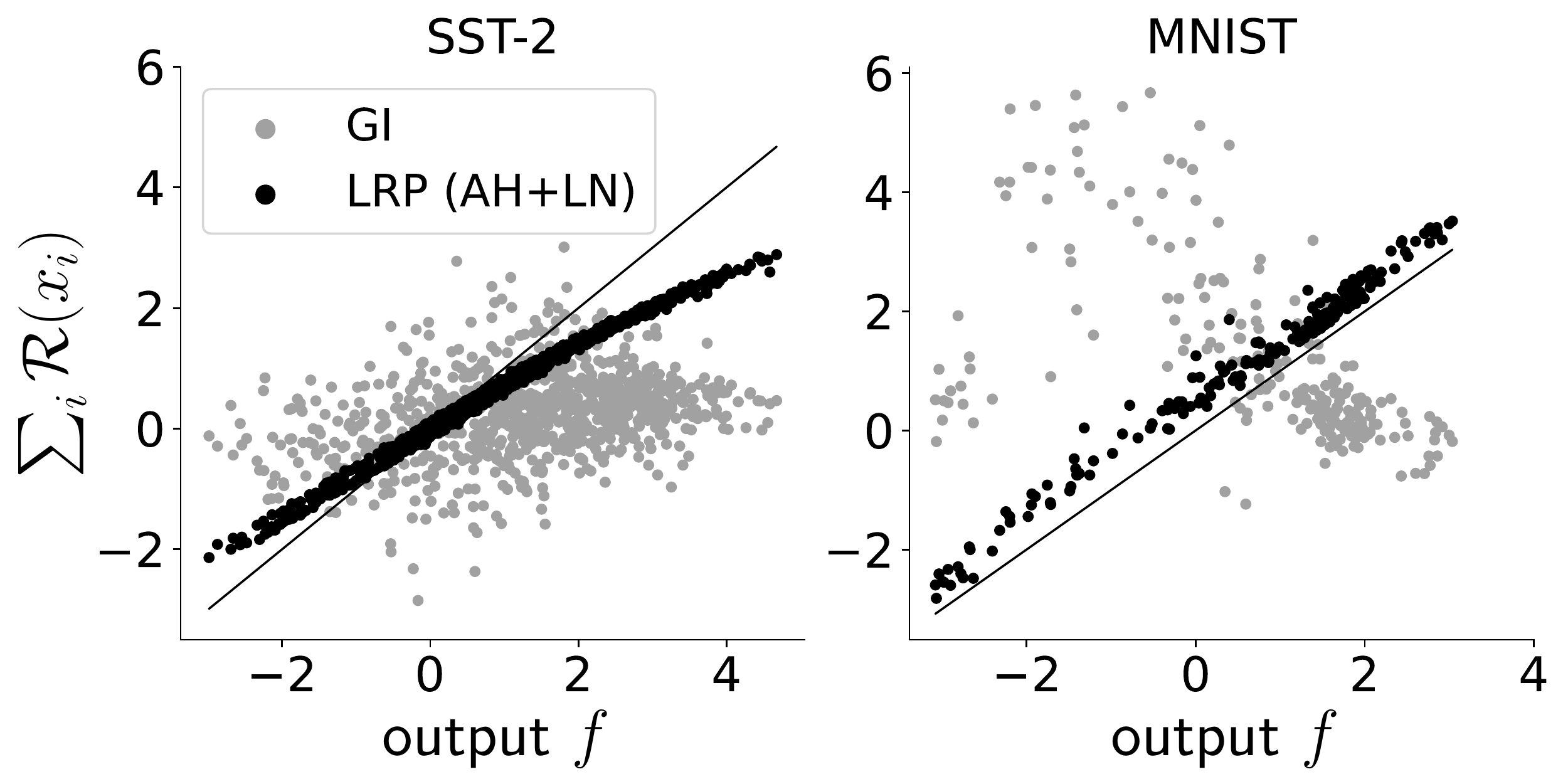} \vspace{-5mm}
    \caption{Conservation, or lack thereof, of GI and LRP attributions on two Transformer models. The two axes represent the output score and the sum of explanation scores over the neural network input variables. Each point in the scatter plot represents one example. The closer the points to the diagonal, the more conservative the explanation technique. }
    \label{fig:conservation}
\end{figure}

 The results are shown in Fig.~\ref{fig:conservation}. Evidently, our LRP approach produces explanations that reflect the output score much more closely than GI, although mild breaks of conservation still occur (likely due to the presence of non-attributable biases in linear layers). Surprisingly, GI explanation scores on MNIST are almost anticorrelated with the output of the model. This clearly highlights a major problem with interpreting GI explanations as attributions of the function output in the context of Transformers. In Appendix \ref{app:conservation} we also provided experiments that show if other explainability methods fulfill conservation.

\subsection{Quantitative Evaluation}

We now test the performance of different explanation methods using an input perturbation scheme in which the most or least relevant input nodes are considered \cite{schnake2020higher}. Two different settings are considered for the Graph Transformer experiments. For the activation task, a good explanation gives an ordering from most to least relevant nodes that when added to an empty graph activate the network output maximally and as quickly as possible. Thus, we observe the output probability $p_c(x)$ of the correct class $c$ and report the area under the activation curve (AUAC) with higher AUAC indicating a more faithful explanation with regards to the correct prediction. 

In the pruning task, we start with the original graph and remove nodes in the order from smallest to largest absolute values. We measure AU-MSE, which is the area under the mean squared error $(y_0-y_{m_t})^2$, with the model output logits of the unpruned model $y_0$ and $y_{m_t}$ representing the output graph after applying the masking $m_t$ at step $t$ to the input graph. A lower AU-MSE is desired and indicates that removing less relevant nodes has little effect on prediction. 

For the NLP experiments, we consider token sequences instead of graph nodes. The activation task starts with an empty sentence of ``UNK" tokens, which are then gradually replaced with the original tokens in the order of highest to lowest relevancy. In the pruning task, we remove tokens from lowest to highest absolute relevance by replacing them with ``UNK" tokens, similarly to the ablation experiments of~\citet{abnar-zuidema-2020-quantifying}.

\begin{figure}[t]
\centering
\includegraphics[width=0.99\linewidth]{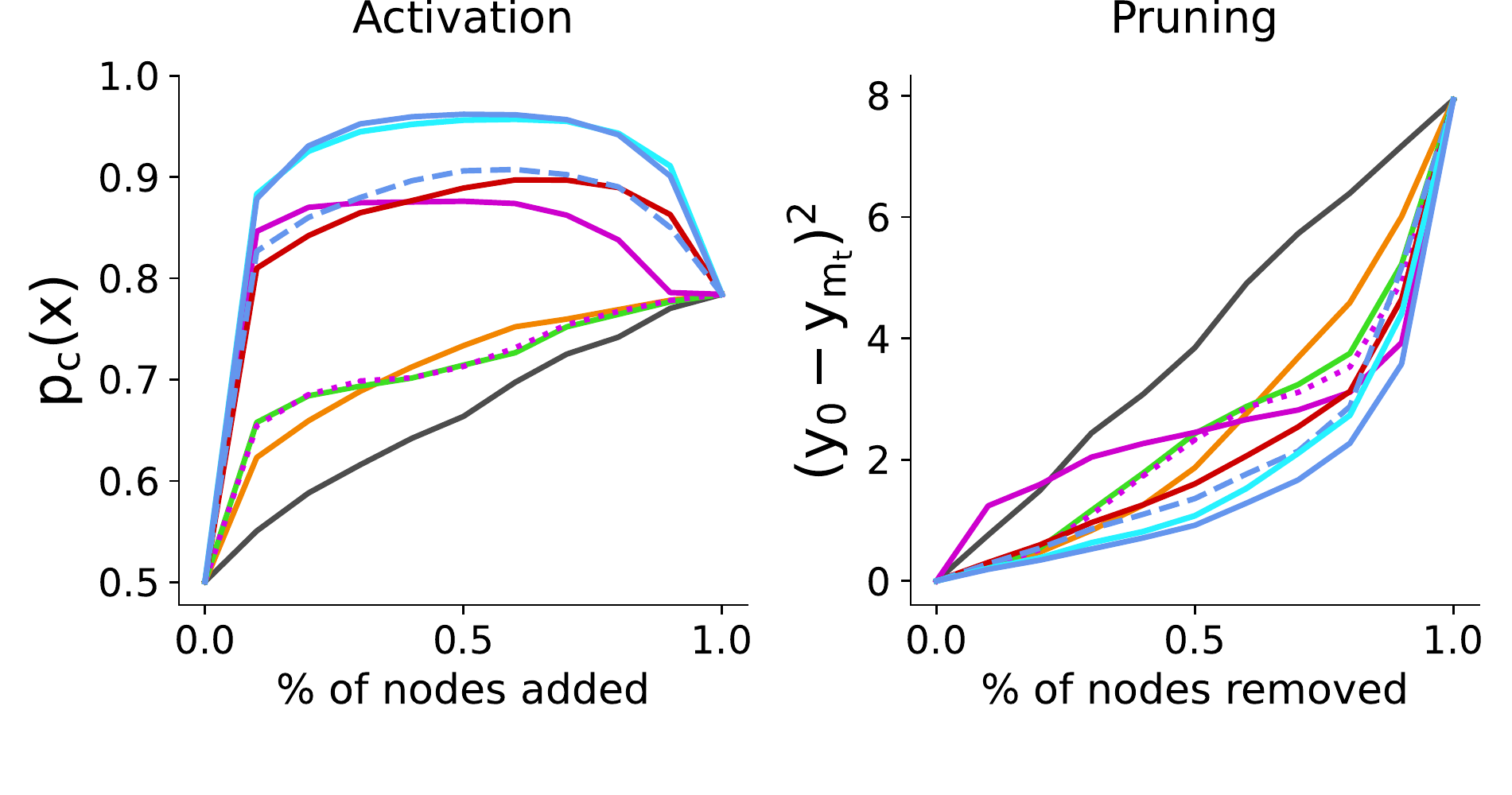}
\includegraphics[width=0.99\linewidth]{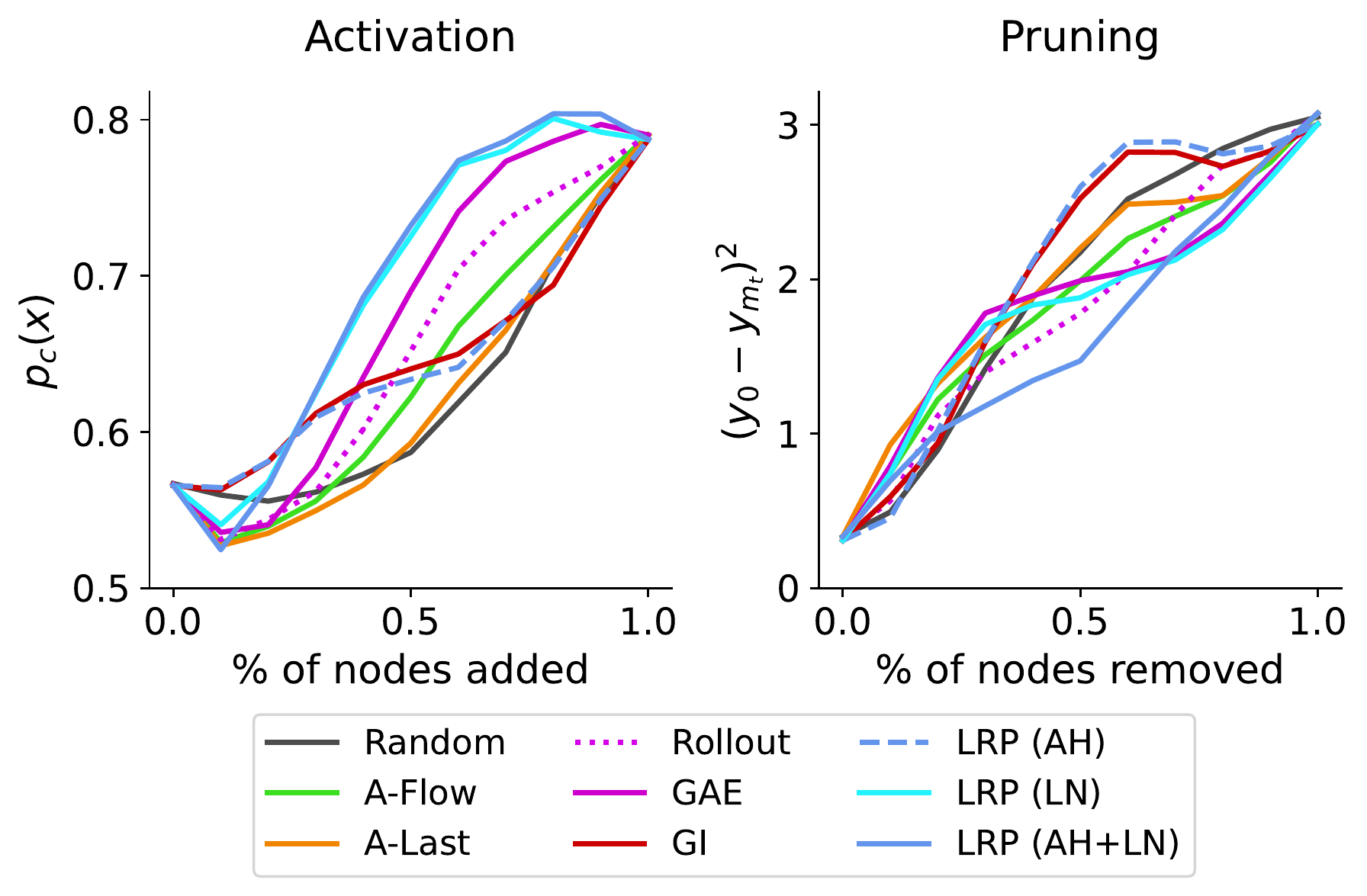}
\caption{Evaluation of explanations using input perturbations. Results for SST-2 (top row) and BACE (lower row) are shown for the activation task, in which the most relevant nodes are added first (left column), and for pruning, in which the nodes of least absolute relevance are removed first (right column).}
\label{fig:flippings}
\end{figure}

In Table \ref{tab:auac} and \ref{tab:aumse}, we report results for the activation and pruning tasks and observe that the proposed handling of the gradient in the attention head (AH) and in layer normalization (LN) during backpropagation indeed results in consistently better performance across all datasets. We see the greatest improvement when applying both the proposed detaching of gating and rescaling terms (AH+LN) together. In addition, explanations based on gradient information are superior to raw attention-based methods (A-Last, Rollout, A-Flow). In Figure \ref{fig:flippings} we show activation and pruning curves for the SST-2 dataset in a Transformer and the BACE dataset on a Graphormer model and include them for all other datasets in Appendix \ref{app:curves}. Additional experiments for inference time analysis over the different explanation methods are provided in Appendix \ref{app:timeanalysis}. We note that the application of the proposed gradient rules leads to a gradual improvement over naive gradient implementations of GI, especially during the transition from very relevant to rather relevant inputs for the activation task. This suggests that the improved LRP explanations are both sparser and more effective at determining the most relevant input nodes, while attributing low relevance to task-irrelevant input nodes. We will inspect this effect qualitatively in the next section.

\begin{table}[t]
\caption{AUAC - area under activation curve. Higher AUAC is better and indicates that adding the most relevant nodes strongly activates the correct model prediction. A-Flow is a computationally very expensive method and was therefore omitted for some methods.}
\vskip 2mm
\centering 
\scalebox{0.9}{
\begin{tabular}{@{}l@{~}c@{~}c@{~}c@{~}c@{~}c@{~}c@{~}c@{~}c@{~}c@{}}
\toprule
       Method       & \rotatebox{90}{IMDB} & \rotatebox{90}{SST-2} & \rotatebox{90}{BACE} &   \rotatebox{90}{MNIST} & \rotatebox{90}{T-Emotions} & \rotatebox{90}{T-Hate} & \rotatebox{90}{T-Sentiment}&
              \rotatebox{90}{Meld-S} & \rotatebox{90}{Semaine}\\
\midrule
Random      &                .673 &                 .664 & .624 & .324 & .516 & .640 & .484 & .460 & .432\\
A-Last      &                .708 &                 .712 & .620  & .862& .542  & .663 & .515 & .483 & .451\\
A-Flow      &                    - &                 .711 & .637  & - & -  & - &- & - & -\\
Rollout     &                .738 &                 .713 & .653  & .358 & .554 & .659 & .520 & .489 & .441\\
GAE &.872& .821 &.675&.426& .675& .762&.611&.548& .532 \\
GI          &                .920 &                 .847 & .646  & .942 & .652 & .772 & .651& .591 & .529\\
\midrule
LRP(AH)      &                .911 &                 .855 & .645 & .942 & .675 & .797 & .668 & .594 & .544\\
LRP (LN) &.935&.907&.702&.947&.735&.829&.710&.632&.593 \\
LRP(AH+LN) &        \textbf{.939} &        \textbf{.908} &   \textbf{.707}    & \textbf{.948} & \textbf{.750}  &  \textbf{.838}  & \textbf{.713}&\textbf{.635} & \textbf{.606}\\

\bottomrule
\end{tabular}
}
\label{tab:auac}
\end{table}
\begin{table}[t]
\caption{AU-MSE - area under the mean squared error $(y_0-y_{m_t})^2$ curve for pruning. Lower is better and indicates that removing less relevant nodes has little effect on the model prediction.}
\vskip 2mm

\centering 
\scalebox{0.9}{

\begin{tabular}{@{}l@{~}c@{~}c@{~}c@{~}c@{~}c@{~}c@{~}c@{~}c@{~}c@{}}
\toprule
Method & \rotatebox{90}{IMDB} & \rotatebox{90}{SST-2}& \rotatebox{90}{BACE} & \rotatebox{90}{MNIST} & \rotatebox{90}{T-Emotions} & \rotatebox{90}{T-Hate} & \rotatebox{90}{T-Sentiment}&
\rotatebox{90}{Meld-S}&
\rotatebox{90}{Semaine}
\\
\midrule
Random      &                2.16 &                 3.97 & 1.95 &   69.82 &4.25& 9.12 & 2.87 & 2.54 & 1.92\\
A-Last      &                1.65 &                 2.56 & 1.99 &  45.82 &3.73 & 7.77 & 1.90 & 1.74 & 1.42\\
A-Flow      &                    - &                 2.52 & 1.87 &  - & - & -  & - & - & -\\
Rollout     &                1.04 &                 2.43 & 1.77 &  115.2 & 2.85 & 6.55 & 1.71 & 1.53 & 1.40\\
GAE     &               1.63 &             2.26 & 1.66 & 59.81 & 2.21 & 7.40 & 1.61 & 1.56 & 1.37\\
GI          &                0.87 &                 2.10 & 2.06 &  18.06 & 2.09& 6.69  & 1.41 & 1.57 & 1.43\\
\midrule
LRP(AH)      &                0.77 &                 2.02 & 2.08 & 18.03 &1.83& 6.43 & 1.43 & 1.69 & 1.38\\
LRP (LN) &0.69&1.78&1.65&17.55&1.55&5.02&1.25&1.50& 1.13\\
LRP(AH+LN) &        \textbf{0.65} &        \textbf{1.56} &    \textbf{1.61}   &  \textbf{17.49} & \textbf{1.47}& \textbf{4.88}  & \textbf{1.23} & \textbf{1.48} & \textbf{1.08}\\

\bottomrule
\end{tabular}
}
\label{tab:aumse}
\end{table}

\subsection{Qualitative Results}
In this section we will examine  the explanations produced with different methods qualitatively for both Transformer and Graphormers.
\begin{figure}
    \centering
    \includegraphics[width=0.48\textwidth]{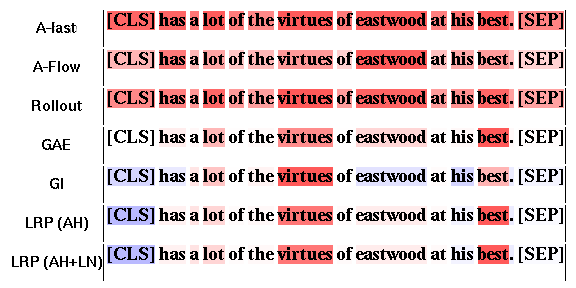} \vspace{-5mm}
    \caption{Visualisation of the attribution scores per word in a sentence of the SST-2 dataset when predicting the sentiment of the sentence. Opaque red color indicates that the word contributes positively to the prediction, blue indicates that the word has a negative contribution. The heatmaps of the methods `A-last', `A-flow' and `Rollout' are limited to non-negative scores.}
    \label{fig:heat_text_example1}
\end{figure}

In Figure \ref{fig:heat_text_example1} we see the relevance scores of different attribution methods when interpreting the prediction of the Transformer model on a sentence from the SST-2 dataset with positive sentiment. We provide additional examples in Appendix \ref{app:sentiment}. We can see that most of the methods show that the model focuses on the words ``best" and ``virtues", which is a reasonable strategy, since they contribute clearly to a positive sentiment.
It is also visible that  `A-last' has a more distinct positive or negative focus on the token ``eastwood" which suggests an over-confidence in names or entities (we will return to this in section \ref{sect:bias_trans}).
Our proposed methods, `LRP (AH)' and `LRP (AH+LN)', consider this entity token to be less relevant for the task (mild coloring) and instead highlight tokens that provide a more general strategy for detecting positive sentiment in a more pronounced manner.

\begin{figure}[t]
    \setlength{\tabcolsep}{1pt} 
     \renewcommand{\arraystretch}{0.9} 
    \centering
    \begin{tabular}{@{}cccc}
            A-Last & Rollout & GAE \\

        \includegraphics[width=0.3\linewidth]{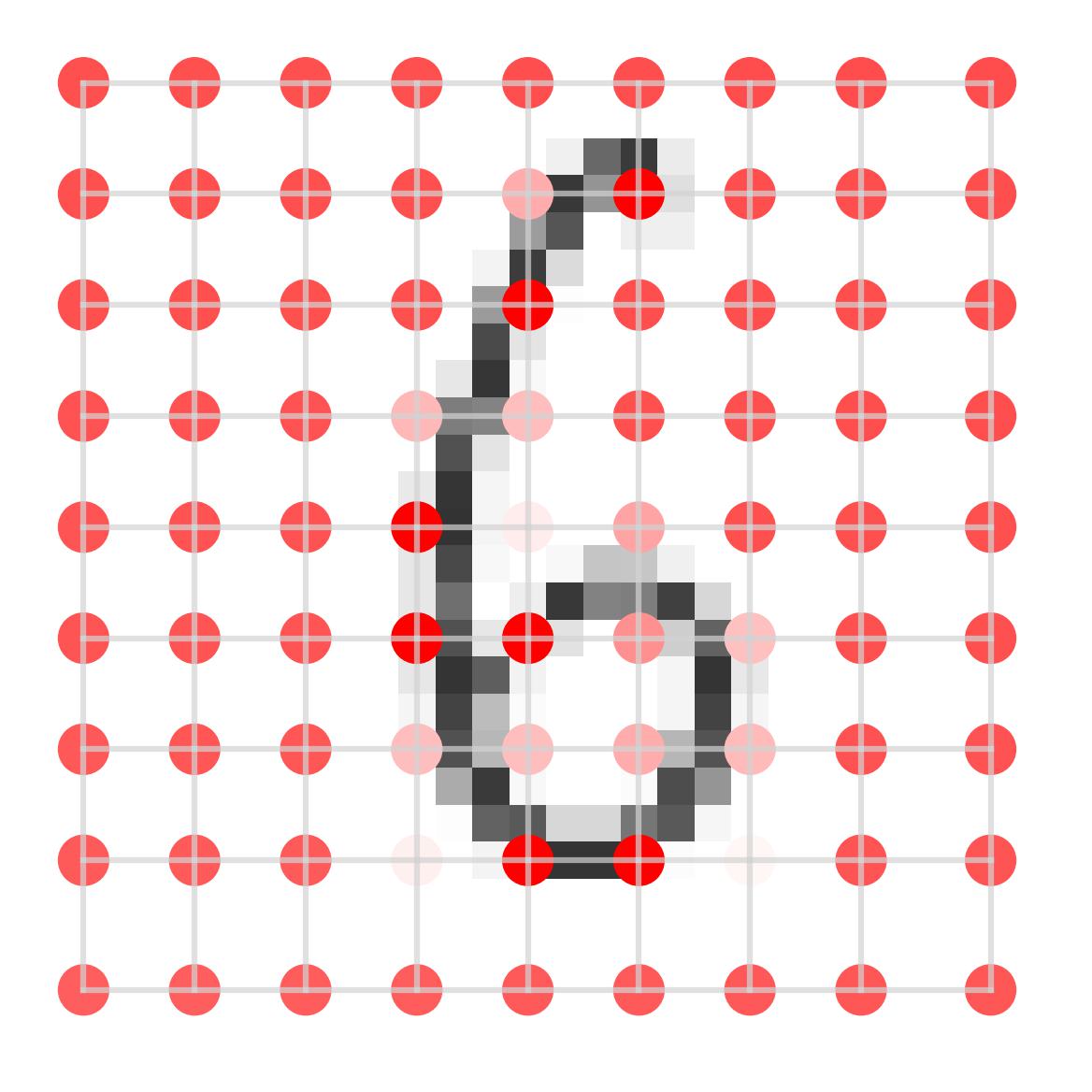}
        &
        \includegraphics[width=0.3\linewidth]{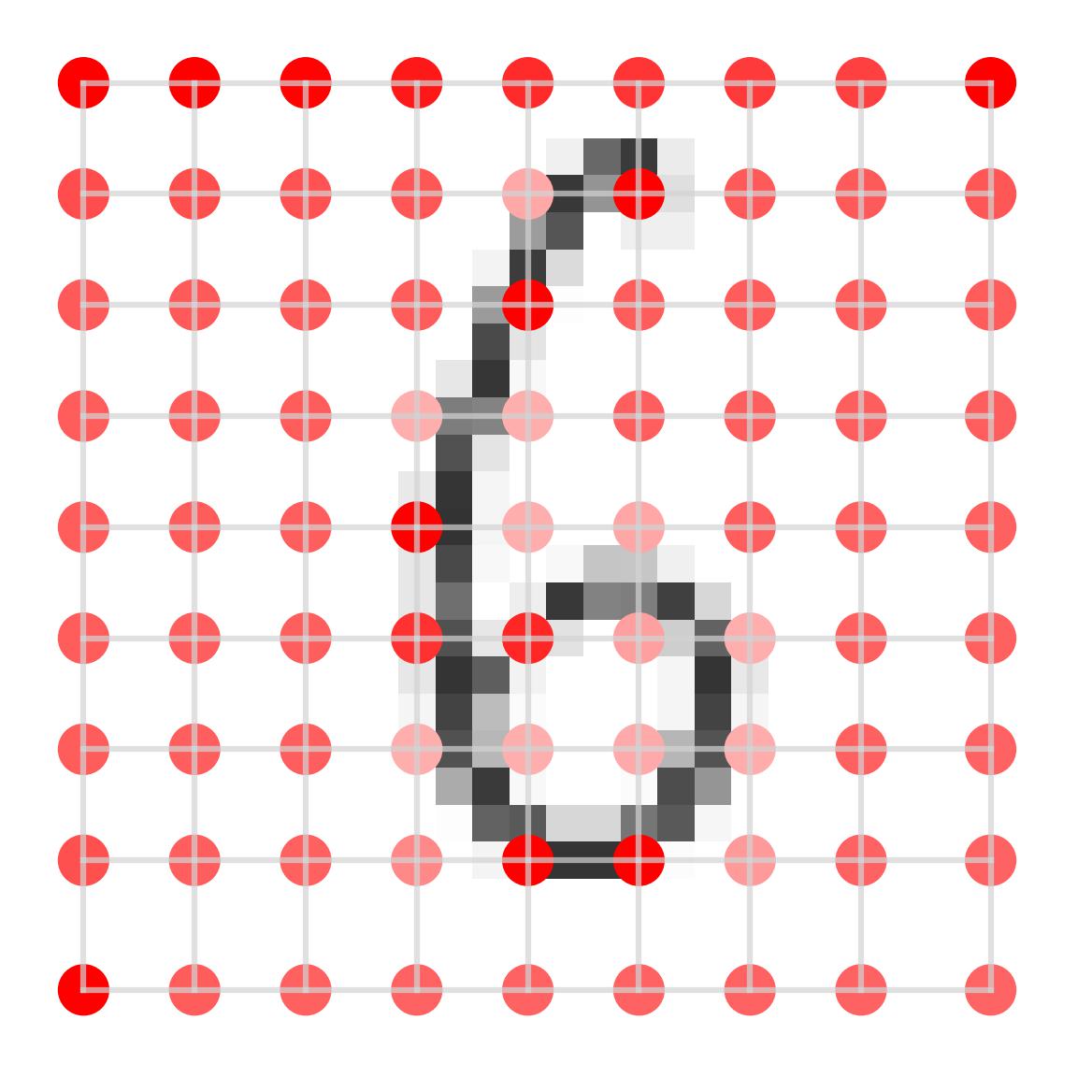}
        &
        \includegraphics[width=0.3\linewidth]{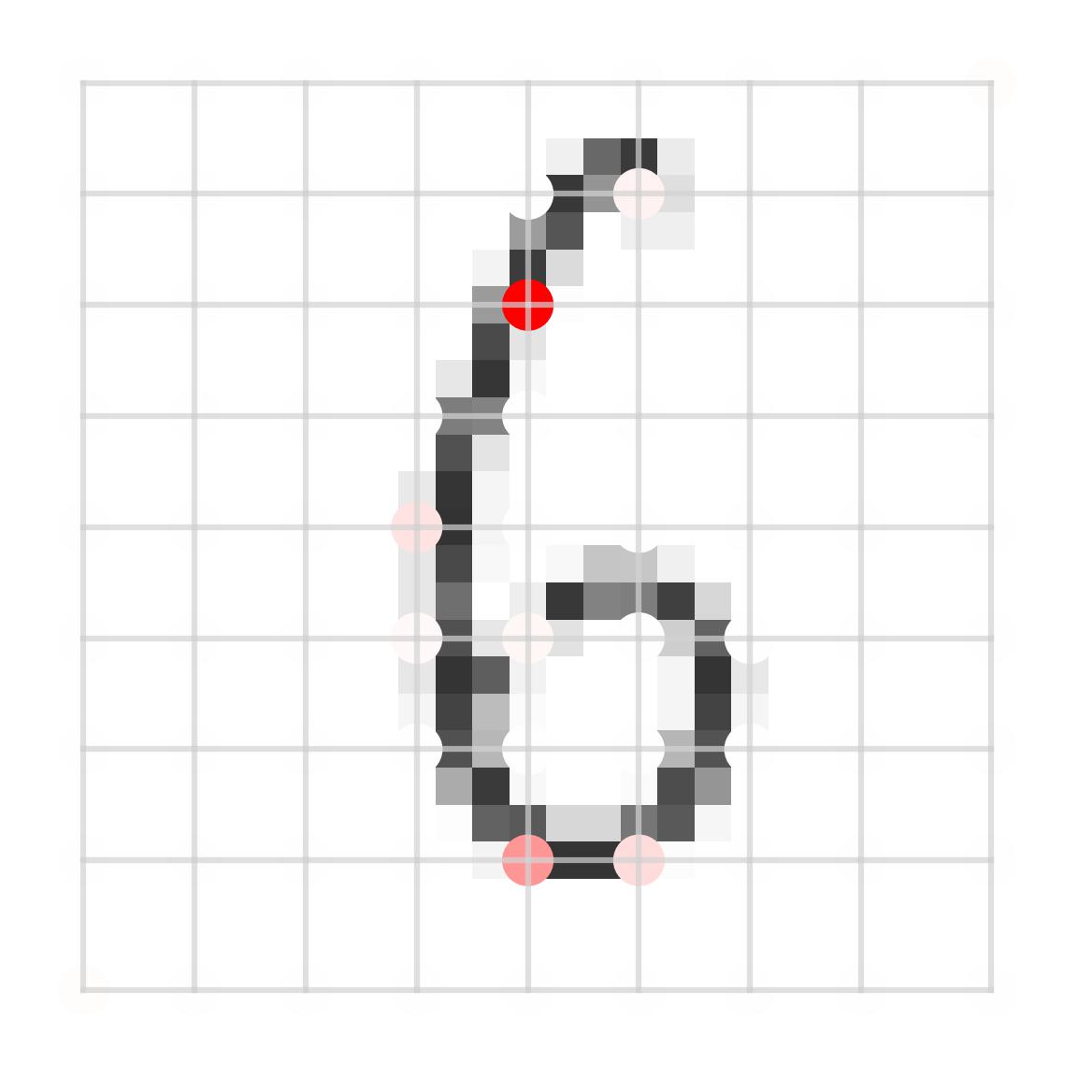}\\
        GI &  LRP (AH) & LRP (AH+LN)     \\
      \includegraphics[width=0.3\linewidth]{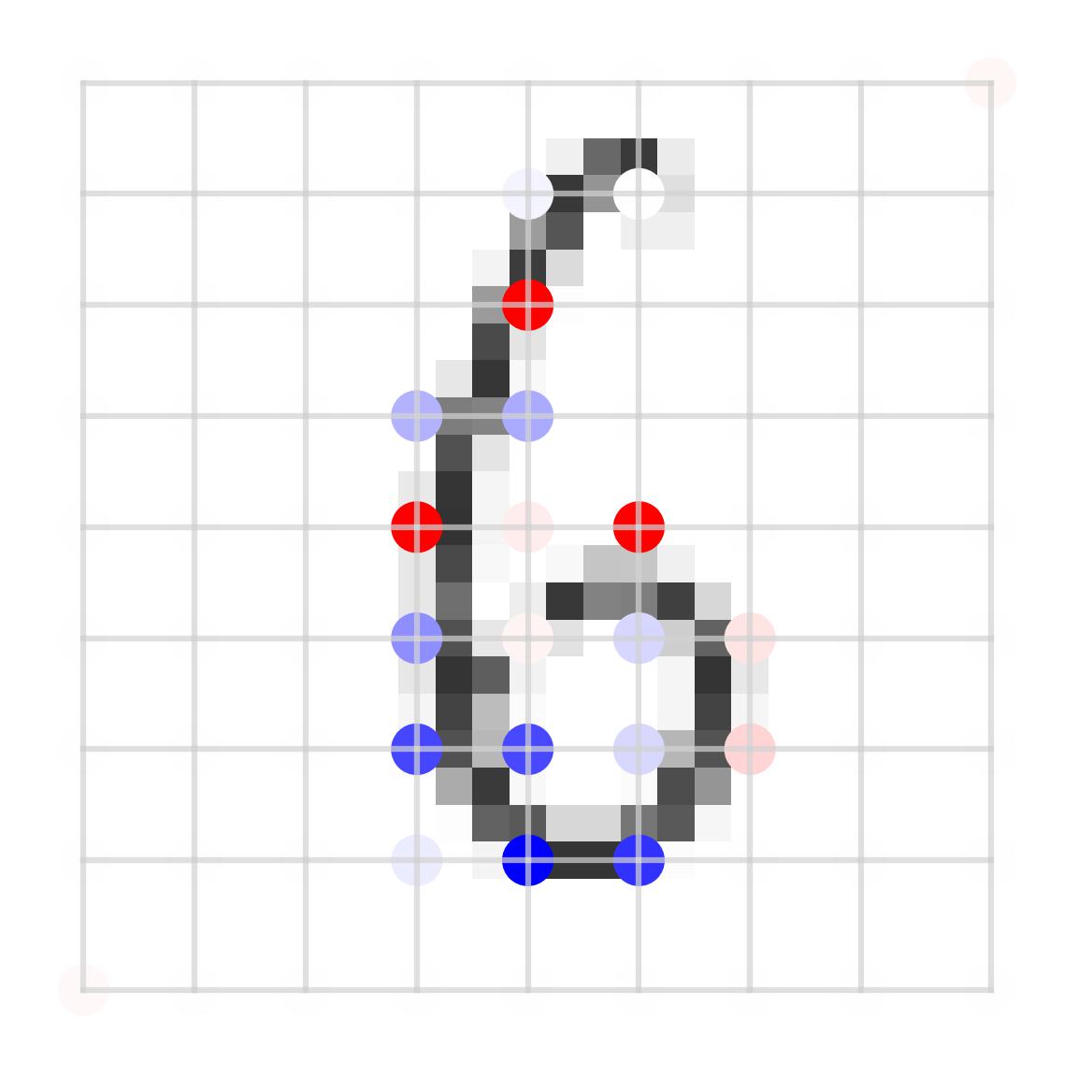}
        &
        \includegraphics[width=0.3\linewidth]{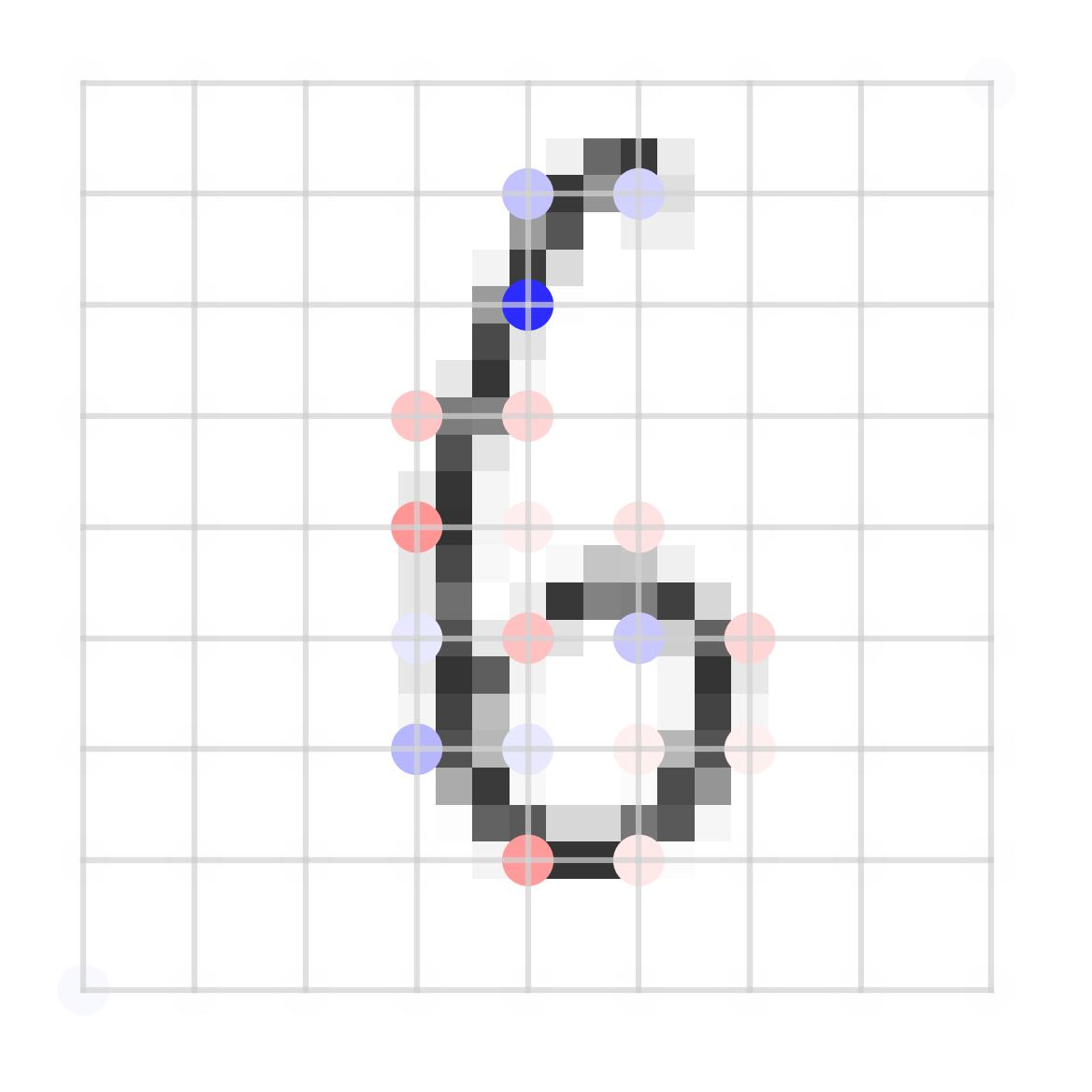} &
        \includegraphics[width=0.3\linewidth]{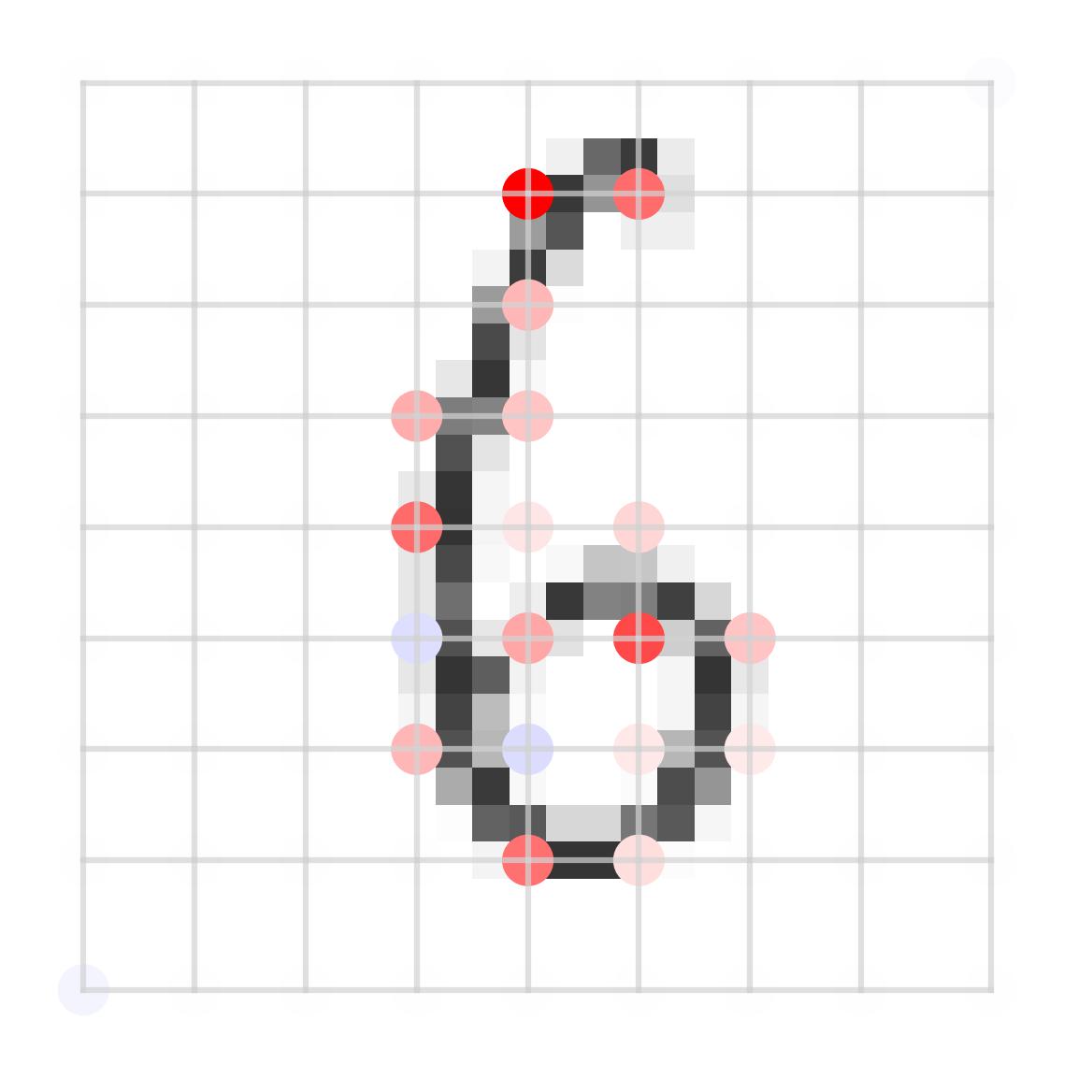} \\
    \end{tabular} \vspace{-2mm}
    \caption{Explanations of MNIST superpixels  on a Graphormer model for digit classification. Relevance of a node in the graph is indicated by color and opacity. Red/Blue contribute positively/negatively to the prediction and magnitude is encoded via color opacity.}
    \label{fig:resultsmnist}
\end{figure}

The explanations obtained for a Graphormer model trained on MNIST superpixels are depicted in Fig.~\ref{fig:resultsmnist} and we provide additional examples in Appendix \ref{app:mnist}. Evidently, the proposed improved explanations using LRP (AH) and LRP (AH+LN) indeed highlight the superpixels that contain digit information more reliably than the other approaches. The advantage of conserving information is clearly visible when comparing naive LRP with LRP (AH+LN) where the effect of distorted relevance propagation is apparent. The weaker performance of raw attention-based methods (A-Last, Rollout and GAE) in the perturbation experiments can now be explained by a lack of specificity to the salient digit in favor of attributing more relevance to uninformative background superpixels.

\subsection{Use Case: Analyzing Bias in Transformers}\label{sect:bias_trans}
\begin{figure*}
    \centering
    \includegraphics[width=0.87\linewidth]{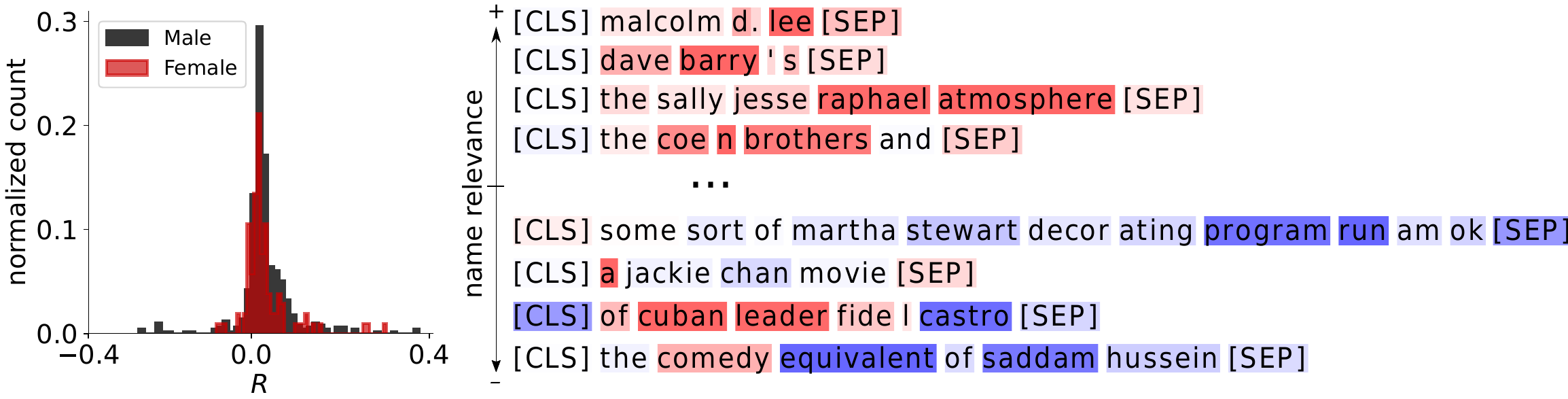}
    \vspace{-2mm}
    \caption{Investigating gender bias in a pre-trained DistilBERT model for sentiment classification on SST-2 movie reviews. Distribution of normalized name occurrences over relevance scores is shown along with sentence samples that contain names that most (top rows)/ least (lower rows) influence a classification towards positive/negative sentiment. }
    \label{fig:bias}
\end{figure*}

We now use our method on a popular Transformer architecture, DistilBERT \cite{Sanh2019DistilBERTAD}, to study the detection of systematic bias in machine learning systems through XAI. We download the publicly available checkpoint for sentiment classification on SST-2 from HuggingFace\footnote{\url{https://huggingface.co/distilbert-base-uncased-finetuned-sst-2-english}} and apply the implementation trick introduced in Section \ref{section:lrp}. In order to detect such bias, template-based approaches, i.e. ``{\textless}name{\textgreater} is a successful {\textless}job\_title{\textgreater}", have been used to test the behavior of the model regarding different systematic relations between, for example, demographics and most likely model predictions \cite{kiritchenko-mohammad-2018-examining, DBLP:conf/emnlp/PrabhakaranHM19,10.1145/3287560.3287572,2020-check-list, ousidhoum-etal-2021-probing}. While this is a flexible approach, it involves the risk of producing model inputs that are out of the training distribution and thus can cause unstable predictions.

Instead, we study relevance attribution to bias-sensitive groups of tokens that are of interest. In this example, we explore the possible gender bias in sentiment analysis using the DistilBERT model. For this, we explain the \textit{difference} between positive and negative model outputs for sentiment classification to observe which entities and related gender may exhibit a tendency to be more/less relevant to change the classification towards a positive/negative sentiment.

In Figure \ref{fig:bias} (left) we first of all observe that there is no consistent bias for female or male names. Overall, there are more male than female names in the dataset, but the distributions of positive/negative sentiment attributed to them are similar.
However, we do observe biased model responses towards certain entity categories: After ranking entities based on their assigned relevance from most to least relevant in Figure \ref{fig:bias} (right) we observe that common Western male names such as  ``lee", ``barry" or ``coen"  can modulate sentiment the strongest towards positive. Interestingly, the first female entity, ``sally jesse raphael", is ranked high because of her typically male family name ``raphael". At the other extreme, among the names with the strongest negative impact on sentiment, we find a non-Western family name (``chan") and male political figures (``saddam hussein", ``castro").

Because our XAI-based approach disentangles the contribution of individual words from that of other words in the sentence, our approach is more immune to confounders than a naive approach that would simply look at the correlation between name occurrence and predicted sentiment.

\section{Conclusion}

Transformers are a major development in machine learning, with a strong uptake in practical applications, in particular NLP. Consequently, there is a need to obtain transparency regarding the decisions made by these models.

In this work, we have shown that one cannot take for granted that common XAI methods will necessarily continue to work well on Transformer models, as they do on standard deep neural networks. In particular, within the framework of Layer-wise Relevance Propagation, we have shown that GI fails to implement conservation, a common property of attribution techniques. Furthermore, our analysis pinpointed that attention heads and LayerNorms need to be addressed specifically. We also proposed specific and easily to implement LRP rules for these layers.

Empirically, our method systematically achieves state-of-the-art results on Transformer and Graphormer models over a broad range of datasets. Lastly, we have showcased our explanation technique on the problem of detecting biases in a sentiment model. Our explanation technique was able to characterize model bias in a detailed manner, without having to generate counterfactual examples and the risk of stepping out of the data manifold. In conclusion, our proposed novel XAI framework for Transformers will ultimately contribute to a safer, more transparent and fair use of state-of-the-art machine learning techniques. 

\section*{Acknowledgements}
KRM was partially supported by the German Ministry for Education and Research under Grant 01IS14013A-E, Grant 01GQ1115, Grant 01GQ0850, as BIFOLD (ref.\ 01IS18025A and ref.\ 01IS18037A) and  the Institute of Information \& Communications Technology
Planning \& Evaluation (IITP) grants funded by the Korea Government under
Grants 2017-0-00451 and 
2019-0-00079.\\
This project has received funding from the European Research Council (ERC) under the European Unions Horizon 2020 research, innovation programme (grant ERC CoG 725974).


\bibliography{arxiv_final}

\begin{thebibliography}{61}
\providecommand{\natexlab}[1]{#1}
\providecommand{\url}[1]{\texttt{#1}}
\expandafter\ifx\csname urlstyle\endcsname\relax
  \providecommand{\doi}[1]{doi: #1}\else
  \providecommand{\doi}{doi: \begingroup \urlstyle{rm}\Url}\fi

\bibitem[Abnar \& Zuidema(2020)Abnar and
  Zuidema]{abnar-zuidema-2020-quantifying}
Abnar, S. and Zuidema, W.~H.
\newblock Quantifying attention flow in transformers.
\newblock In Jurafsky, D., Chai, J., Schluter, N., and Tetreault, J.~R. (eds.),
  \emph{Proceedings of the 58th Annual Meeting of the Association for
  Computational Linguistics, {ACL} 2020}, pp.\  4190--4197, 2020.

\bibitem[Aleisa et~al.(2021)Aleisa, Beloff, and White]{sussex97276}
Aleisa, M.~A., Beloff, N., and White, M.
\newblock Airm: a new ai recruiting model for the saudi arabia labor market.
\newblock In Arai, K. (ed.), \emph{Intelligent Systems Conference (IntelliSys)
  2021}, volume 3: 296 of \emph{Lecture Notes in Networks and Systems}, pp.\
  105--124, Cham, September 2021. Springer.

\bibitem[Ancona et~al.(2019)Ancona, Ceolini, {\"{O}}ztireli, and
  Gross]{DBLP:series/lncs/AnconaCOG19}
Ancona, M., Ceolini, E., {\"{O}}ztireli, C., and Gross, M.~H.
\newblock Gradient-based attribution methods.
\newblock In \emph{Explainable {AI:} Interpreting, Explaining and Visualizing
  Deep Learning}, volume 11700 of \emph{Lecture Notes in Computer Science},
  pp.\  169--191. Springer, 2019.

\bibitem[Arras et~al.(2019)Arras, Arjona-Medina, Widrich, Montavon, Gillhofer,
  M{\"u}ller, Hochreiter, and Samek]{arras-lncs19}
Arras, L., Arjona-Medina, J., Widrich, M., Montavon, G., Gillhofer, M.,
  M{\"u}ller, K.-R., Hochreiter, S., and Samek, W.
\newblock {Explaining and Interpreting LSTMs}.
\newblock In Samek, W., Montavon, G., Vedaldi, A., Hansen, L.~K., and
  M{\"u}ller, K.-R. (eds.), \emph{Explainable AI: Interpreting, Explaining and
  Visualizing Deep Learning}, volume 11700 of \emph{Lecture Notes in Computer
  Science}, pp.\  211--238. Springer, Cham, 2019.

\bibitem[Arrieta et~al.(2020)Arrieta, Rodr{\'{\i}}guez, Ser, Bennetot, Tabik,
  Barbado, Garc{\'{\i}}a, Gil{-}Lopez, Molina, Benjamins, Chatila, and
  Herrera]{BARREDOARRIETA202082}
Arrieta, A.~B., Rodr{\'{\i}}guez, N.~D., Ser, J.~D., Bennetot, A., Tabik, S.,
  Barbado, A., Garc{\'{\i}}a, S., Gil{-}Lopez, S., Molina, D., Benjamins, R.,
  Chatila, R., and Herrera, F.
\newblock Explainable artificial intelligence {(XAI):} concepts, taxonomies,
  opportunities and challenges toward responsible {AI}.
\newblock \emph{Information Fusion}, 58:\penalty0 82--115, 2020.

\bibitem[Atanasova et~al.(2020)Atanasova, Simonsen, Lioma, and
  Augenstein]{atanasova-etal-2020-diagnostic}
Atanasova, P., Simonsen, J.~G., Lioma, C., and Augenstein, I.
\newblock A diagnostic study of explainability techniques for text
  classification.
\newblock In Webber, B., Cohn, T., He, Y., and Liu, Y. (eds.),
  \emph{Proceedings of the 2020 Conference on Empirical Methods in Natural
  Language Processing, {EMNLP} 2020, Online, November 16-20, 2020}, pp.\
  3256--3274. Association for Computational Linguistics, 2020.

\bibitem[Bach et~al.(2015)Bach, Binder, Montavon, Klauschen, M{\"u}ller, and
  Samek]{bach-plos15}
Bach, S., Binder, A., Montavon, G., Klauschen, F., M{\"u}ller, K.-R., and
  Samek, W.
\newblock On pixel-wise explanations for non-linear classifier decisions by
  layer-wise relevance propagation.
\newblock \emph{PLoS ONE}, 10\penalty0 (7):\penalty0 e0130140, 2015.

\bibitem[Bahdanau et~al.(2015)Bahdanau, Cho, and
  Bengio]{bahdanau2015nmtranslation}
Bahdanau, D., Cho, K., and Bengio, Y.
\newblock Neural machine translation by jointly learning to align and
  translate.
\newblock In Bengio, Y. and LeCun, Y. (eds.), \emph{3rd International
  Conference on Learning Representations, {ICLR} 2015, San Diego, CA, USA, May
  7-9, 2015, Conference Track Proceedings}, 2015.

\bibitem[Bolukbasi et~al.(2016)Bolukbasi, Chang, Zou, Saligrama, and
  Kalai]{bolukbasi2016man}
Bolukbasi, T., Chang, K.-W., Zou, J., Saligrama, V., and Kalai, A.
\newblock Man is to computer programmer as woman is to homemaker? debiasing
  word embeddings.
\newblock In \emph{Proceedings of the 30th International Conference on Neural
  Information Processing Systems}, pp.\  4356–4364, Red Hook, NY, USA, 2016.
  Curran Associates Inc.

\bibitem[Chapuis et~al.(2020)Chapuis, Colombo, Manica, Labeau, and
  Clavel]{chapuis-etal-2020-hierarchical}
Chapuis, E., Colombo, P., Manica, M., Labeau, M., and Clavel, C.
\newblock Hierarchical pre-training for sequence labelling in spoken dialog.
\newblock In \emph{Findings of the Association for Computational Linguistics:
  {EMNLP} 2020}, pp.\  2636--2648, 2020.

\bibitem[Chefer et~al.(2021{\natexlab{a}})Chefer, Gur, and
  Wolf]{Chefer_2021_ICCV}
Chefer, H., Gur, S., and Wolf, L.
\newblock Generic attention-model explainability for interpreting bi-modal and
  encoder-decoder transformers.
\newblock In \emph{Proceedings of the IEEE/CVF International Conference on
  Computer Vision (ICCV)}, pp.\  397--406, 2021{\natexlab{a}}.

\bibitem[Chefer et~al.(2021{\natexlab{b}})Chefer, Gur, and
  Wolf]{chefer2021transformer}
Chefer, H., Gur, S., and Wolf, L.
\newblock Transformer interpretability beyond attention visualization.
\newblock In \emph{Proceedings of the IEEE/CVF Conference on Computer Vision
  and Pattern Recognition}, pp.\  782--791, 2021{\natexlab{b}}.

\bibitem[Danilevsky et~al.(2020)Danilevsky, Qian, Aharonov, Katsis, Kawas, and
  Sen]{DBLP:conf/ijcnlp/DanilevskyQAKKS20}
Danilevsky, M., Qian, K., Aharonov, R., Katsis, Y., Kawas, B., and Sen, P.
\newblock A survey of the state of explainable {AI} for natural language
  processing.
\newblock In \emph{{AACL/IJCNLP}}, pp.\  447--459. Association for
  Computational Linguistics, 2020.

\bibitem[De-Arteaga et~al.(2019)De-Arteaga, Romanov, Wallach, Chayes, Borgs,
  Chouldechova, Geyik, Kenthapadi, and Kalai]{10.1145/3287560.3287572}
De-Arteaga, M., Romanov, A., Wallach, H., Chayes, J., Borgs, C., Chouldechova,
  A., Geyik, S., Kenthapadi, K., and Kalai, A.~T.
\newblock Bias in bios: A case study of semantic representation bias in a
  high-stakes setting.
\newblock In \emph{Proceedings of the Conference on Fairness, Accountability,
  and Transparency}, pp.\  120–128, New York, NY, USA, 2019. Association for
  Computing Machinery.

\bibitem[Denil et~al.(2014)Denil, Demiraj, and
  de~Freitas]{Denil2014ExtractionOS}
Denil, M., Demiraj, A., and de~Freitas, N.
\newblock Extraction of salient sentences from labelled documents.
\newblock \emph{ArXiv}, abs/1412.6815, 2014.

\bibitem[Devlin et~al.(2019)Devlin, Chang, Lee, and Toutanova]{devlin2019bert}
Devlin, J., Chang, M., Lee, K., and Toutanova, K.
\newblock {BERT:} pre-training of deep bidirectional transformers for language
  understanding.
\newblock In \emph{Proceedings of the 2019 Conference of the North American
  Chapter of the Association for Computational Linguistics: Human Language
  Technologies, {NAACL-HLT} 2019}, pp.\  4171--4186, 2019.

\bibitem[Ding et~al.(2017)Ding, Liu, Luan, and Sun]{ding-etal-2017-visualizing}
Ding, Y., Liu, Y., Luan, H., and Sun, M.
\newblock Visualizing and understanding neural machine translation.
\newblock In \emph{Proceedings of the 55th Annual Meeting of the Association
  for Computational Linguistics}, pp.\  1150--1159, 2017.

\bibitem[Dosovitskiy et~al.(2021)Dosovitskiy, Beyer, Kolesnikov, Weissenborn,
  Zhai, Unterthiner, Dehghani, Minderer, Heigold, Gelly, Uszkoreit, and
  Houlsby]{dosovitskiy2021image}
Dosovitskiy, A., Beyer, L., Kolesnikov, A., Weissenborn, D., Zhai, X.,
  Unterthiner, T., Dehghani, M., Minderer, M., Heigold, G., Gelly, S.,
  Uszkoreit, J., and Houlsby, N.
\newblock An image is worth 16x16 words: Transformers for image recognition at
  scale.
\newblock In \emph{9th International Conference on Learning Representations,
  {ICLR} 2021}, 2021.

\bibitem[Feng et~al.(2018)Feng, Wallace, Grissom~II, Iyyer, Rodriguez, and
  Boyd-Graber]{feng-etal-2018-pathologies}
Feng, S., Wallace, E., Grissom~II, A., Iyyer, M., Rodriguez, P., and
  Boyd-Graber, J.
\newblock Pathologies of neural models make interpretations difficult.
\newblock In \emph{Proceedings of the 2018 Conference on Empirical Methods in
  Natural Language Processing}, pp.\  3719--3728, Brussels, Belgium,
  October-November 2018. Association for Computational Linguistics.

\bibitem[Gonen \& Goldberg(2019)Gonen and Goldberg]{gonen2019lipstick}
Gonen, H. and Goldberg, Y.
\newblock Lipstick on a pig: {D}ebiasing methods cover up systematic gender
  biases in word embeddings but do not remove them.
\newblock In \emph{Proceedings of the 2019 Conference of the North {A}merican
  Chapter of the Association for Computational Linguistics: Human Language
  Technologies}, pp.\  609--614, 2019.

\bibitem[Guidotti et~al.(2019)Guidotti, Monreale, Ruggieri, Turini, Giannotti,
  and Pedreschi]{DBLP:journals/csur/GuidottiMRTGP19}
Guidotti, R., Monreale, A., Ruggieri, S., Turini, F., Giannotti, F., and
  Pedreschi, D.
\newblock A survey of methods for explaining black box models.
\newblock \emph{{ACM} Comput. Surv.}, 51\penalty0 (5):\penalty0 93:1--93:42,
  2019.

\bibitem[Hesse et~al.(2021)Hesse, Schaub-Meyer, and Roth]{hesse2021fast}
Hesse, R., Schaub-Meyer, S., and Roth, S.
\newblock Fast axiomatic attribution for neural networks.
\newblock In Beygelzimer, A., Dauphin, Y., Liang, P., and Vaughan, J.~W.
  (eds.), \emph{Advances in Neural Information Processing Systems}, 2021.

\bibitem[Hollenstein \& Beinborn(2021)Hollenstein and
  Beinborn]{hollenstein-beinborn-2021-relative}
Hollenstein, N. and Beinborn, L.
\newblock Relative importance in sentence processing.
\newblock In \emph{Proceedings of the 59th Annual Meeting of the Association
  for Computational Linguistics and the 11th International Joint Conference on
  Natural Language Processing}, pp.\  141--150, 2021.

\bibitem[Huang et~al.(2020)Huang, Yamada, Tian, Singh, Yin, and
  Chang]{huang2020graphlime}
Huang, Q., Yamada, M., Tian, Y., Singh, D., Yin, D., and Chang, Y.
\newblock Graphlime: Local interpretable model explanations for graph neural
  networks.
\newblock \emph{arXiv preprint arXiv:2001.06216}, 2020.

\bibitem[Jain \& Wallace(2019)Jain and Wallace]{jain-wallace-2019-attention}
Jain, S. and Wallace, B.~C.
\newblock {A}ttention is not {E}xplanation.
\newblock In \emph{Proceedings of the 2019 Conference of the North {A}merican
  Chapter of the Association for Computational Linguistics: Human Language
  Technologies}, pp.\  3543--3556, Minneapolis, Minnesota, June 2019.
  Association for Computational Linguistics.

\bibitem[Kiritchenko \& Mohammad(2018)Kiritchenko and
  Mohammad]{kiritchenko-mohammad-2018-examining}
Kiritchenko, S. and Mohammad, S.
\newblock Examining gender and race bias in two hundred sentiment analysis
  systems.
\newblock In \emph{Proceedings of the Seventh Joint Conference on Lexical and
  Computational Semantics}, pp.\  43--53, 2018.

\bibitem[Lundberg \& Lee(2017)Lundberg and Lee]{DBLP:conf/nips/LundbergL17}
Lundberg, S.~M. and Lee, S.
\newblock A unified approach to interpreting model predictions.
\newblock In \emph{Advances in Neural Information Processing Systems 30: Annual
  Conference on Neural Information Processing Systems 2017}, pp.\  4765--4774,
  2017.

\bibitem[Luo et~al.(2020)Luo, Cheng, Xu, Yu, Zong, Chen, and
  Zhang]{luo2020parameterized}
Luo, D., Cheng, W., Xu, D., Yu, W., Zong, B., Chen, H., and Zhang, X.
\newblock Parameterized explainer for graph neural network.
\newblock In \emph{Advances in Neural Information Processing Systems 33: Annual
  Conference on Neural Information Processing Systems 2020}, volume~33, 2020.

\bibitem[Maas et~al.(2011)Maas, Daly, Pham, Huang, Ng, and
  Potts]{maas-EtAl:2011:ACL-HLT2011}
Maas, A.~L., Daly, R.~E., Pham, P.~T., Huang, D., Ng, A.~Y., and Potts, C.
\newblock Learning word vectors for sentiment analysis.
\newblock In \emph{Proceedings of the 49th Annual Meeting of the Association
  for Computational Linguistics: Human Language Technologies}, pp.\  142--150,
  2011.

\bibitem[Maziarka et~al.(2020)Maziarka, Danel, Mucha, Rataj, Tabor, and
  Jastrz{k{e}}bski]{maziarka2020molecule}
Maziarka, {\L}., Danel, T., Mucha, S., Rataj, K., Tabor, J., and
  Jastrz{k{e}}bski, S.
\newblock Molecule attention transformer.
\newblock \emph{arXiv preprint arXiv:2002.08264}, 2020.

\bibitem[Montavon(2019)]{DBLP:series/lncs/Montavon19}
Montavon, G.
\newblock Gradient-based vs. propagation-based explanations: An axiomatic
  comparison.
\newblock In \emph{Explainable AI: Interpreting, Explaining and Visualizing
  Deep Learning}, pp.\  253--265. Springer International Publishing, 2019.

\bibitem[Montavon et~al.(2018)Montavon, Samek, and
  M{\"{u}}ller]{DBLP:journals/dsp/MontavonSM18}
Montavon, G., Samek, W., and M{\"{u}}ller, K.
\newblock Methods for interpreting and understanding deep neural networks.
\newblock \emph{Digit. Signal Process.}, 73:\penalty0 1--15, 2018.

\bibitem[Monti et~al.(2017)Monti, Boscaini, Masci, Rodol{\`{a}}, Svoboda, and
  Bronstein]{monti2017geometric}
Monti, F., Boscaini, D., Masci, J., Rodol{\`{a}}, E., Svoboda, J., and
  Bronstein, M.~M.
\newblock Geometric deep learning on graphs and manifolds using mixture model
  cnns.
\newblock In \emph{2017 {IEEE} Conference on Computer Vision and Pattern
  Recognition}, pp.\  5115--5124, 2017.

\bibitem[Narayanan et~al.(2021)Narayanan, Shoeybi, Casper, LeGresley, Patwary,
  Korthikanti, Vainbrand, Kashinkunti, Bernauer, Catanzaro, Phanishayee, and
  Zaharia]{10.1145/3458817.3476209}
Narayanan, D., Shoeybi, M., Casper, J., LeGresley, P., Patwary, M.,
  Korthikanti, V., Vainbrand, D., Kashinkunti, P., Bernauer, J., Catanzaro, B.,
  Phanishayee, A., and Zaharia, M.
\newblock Efficient large-scale language model training on {GPU} clusters using
  megatron-lm.
\newblock In \emph{{SC} '21: The International Conference for High Performance
  Computing, Networking, Storage and Analysis}, 2021.

\bibitem[Ousidhoum et~al.(2021)Ousidhoum, Zhao, Fang, Song, and
  Yeung]{ousidhoum-etal-2021-probing}
Ousidhoum, N., Zhao, X., Fang, T., Song, Y., and Yeung, D.-Y.
\newblock Probing toxic content in large pre-trained language models.
\newblock In \emph{Proceedings of the 59th Annual Meeting of the Association
  for Computational Linguistics and the 11th International Joint Conference on
  Natural Language Processing}, pp.\  4262--4274. Association for Computational
  Linguistics, 2021.

\bibitem[Pope et~al.(2019)Pope, Kolouri, Rostami, Martin, and
  Hoffmann]{pope2019explainability}
Pope, P.~E., Kolouri, S., Rostami, M., Martin, C.~E., and Hoffmann, H.
\newblock Explainability methods for graph convolutional neural networks.
\newblock In \emph{2019 IEEE/CVF Conference on Computer Vision and Pattern
  Recognition (CVPR)}, pp.\  10772--10781, 2019.

\bibitem[Prabhakaran et~al.(2019)Prabhakaran, Hutchinson, and
  Mitchell]{DBLP:conf/emnlp/PrabhakaranHM19}
Prabhakaran, V., Hutchinson, B., and Mitchell, M.
\newblock Perturbation sensitivity analysis to detect unintended model biases.
\newblock In \emph{Proceedings of the 2019 Conference on Empirical Methods in
  Natural Language Processing and the 9th International Joint Conference on
  Natural Language Processing, {EMNLP-IJCNLP}}, pp.\  5739--5744, 2019.

\bibitem[Radford et~al.(2019)Radford, Wu, Child, Luan, Amodei, Sutskever,
  et~al.]{radford2019language}
Radford, A., Wu, J., Child, R., Luan, D., Amodei, D., Sutskever, I., et~al.
\newblock Language models are unsupervised multitask learners.
\newblock \emph{OpenAI blog}, 2019.

\bibitem[Ribeiro et~al.(2020)Ribeiro, Wu, Guestrin, and Singh]{2020-check-list}
Ribeiro, M.~T., Wu, T.~S., Guestrin, C., and Singh, S.
\newblock Beyond accuracy: Behavioral testing of nlp models with checklist.
\newblock In \emph{Proc. Association for Computational Linguistics (ACL)}, pp.\
   4902--4912, 2020.

\bibitem[Samek et~al.(2019)Samek, Montavon, Vedaldi, Hansen, and
  M{\"{u}}ller]{DBLP:series/lncs/11700}
Samek, W., Montavon, G., Vedaldi, A., Hansen, L.~K., and M{\"{u}}ller, K.-R.
  (eds.).
\newblock \emph{Explainable {AI:} Interpreting, Explaining and Visualizing Deep
  Learning}, volume 11700 of \emph{Lecture Notes in Computer Science}.
\newblock Springer, 2019.

\bibitem[Samek et~al.(2021)Samek, Montavon, Lapuschkin, Anders, and
  Müller]{samek2021xaireview}
Samek, W., Montavon, G., Lapuschkin, S., Anders, C.~J., and Müller, K.-R.
\newblock Explaining deep neural networks and beyond: A review of methods and
  applications.
\newblock \emph{Proceedings of the IEEE}, 109\penalty0 (3):\penalty0 247--278,
  2021.

\bibitem[Sanh et~al.(2019)Sanh, Debut, Chaumond, and
  Wolf]{Sanh2019DistilBERTAD}
Sanh, V., Debut, L., Chaumond, J., and Wolf, T.
\newblock Distilbert, a distilled version of bert: smaller, faster, cheaper and
  lighter.
\newblock \emph{ArXiv}, abs/1910.01108, 2019.

\bibitem[Schnake et~al.(2021)Schnake, Eberle, Lederer, Nakajima, Sch\"utt,
  M\"uller, and Montavon]{schnake2020higher}
Schnake, T., Eberle, O., Lederer, J., Nakajima, S., Sch\"utt, K.~T., M\"uller,
  K.-R., and Montavon, G.
\newblock Higher-order explanations of graph neural networks via relevant
  walks.
\newblock \emph{IEEE Transactions on Pattern Analysis and Machine
  Intelligence}, pp.\  10.1109/TPAMI.2021.3115452, 2021.

\bibitem[Shapley(1953)]{Shapley1953}
Shapley, L.~S.
\newblock \emph{A Value for n-Person Games}, pp.\  307--318.
\newblock Princeton University Press, 1953.

\bibitem[Shrikumar et~al.(2017)Shrikumar, Greenside, and
  Kundaje]{DBLP:journals/corr/ShrikumarGSK16}
Shrikumar, A., Greenside, P., and Kundaje, A.
\newblock Learning important features through propagating activation
  differences.
\newblock In \emph{Proceedings of the 34th International Conference on Machine
  Learning - Volume 70}, ICML'17, pp.\  3145–3153, 2017.

\bibitem[Socher et~al.(2013)Socher, Perelygin, Wu, Chuang, Manning, Ng, and
  Potts]{socher2013-recursive}
Socher, R., Perelygin, A., Wu, J., Chuang, J., Manning, C.~D., Ng, A.~Y., and
  Potts, C.
\newblock Recursive deep models for semantic compositionality over a sentiment
  treebank.
\newblock In \emph{Proceedings of the 2013 Conference on Empirical Methods in
  Natural Language Processing}, pp.\  1631--1642, 2013.

\bibitem[Srinivas \& Fleuret(2019)Srinivas and
  Fleuret]{DBLP:conf/nips/SrinivasF19}
Srinivas, S. and Fleuret, F.
\newblock Full-gradient representation for neural network visualization.
\newblock In \emph{Advances in Neural Information Processing Systems 32: Annual
  Conference on Neural Information Processing Systems 2019}, pp.\  4126--4135,
  2019.

\bibitem[Strumbelj \& Kononenko(2010)Strumbelj and
  Kononenko]{DBLP:journals/jmlr/StrumbeljK10}
Strumbelj, E. and Kononenko, I.
\newblock An efficient explanation of individual classifications using game
  theory.
\newblock \emph{J. Mach. Learn. Res.}, 11:\penalty0 1--18, 2010.

\bibitem[Subramanian et~al.(2016)Subramanian, Ramsundar, Pande, and
  Denny]{subramanian2016computational}
Subramanian, G., Ramsundar, B., Pande, V., and Denny, R.~A.
\newblock Computational modeling of $\beta$-secretase 1 (bace-1) inhibitors
  using ligand based approaches.
\newblock \emph{Journal of chemical information and modeling}, 56\penalty0
  (10):\penalty0 1936--1949, 2016.

\bibitem[Sundararajan et~al.(2017)Sundararajan, Taly, and
  Yan]{DBLP:conf/icml/SundararajanTY17}
Sundararajan, M., Taly, A., and Yan, Q.
\newblock Axiomatic attribution for deep networks.
\newblock In \emph{{ICML}}, volume~70 of \emph{Proceedings of Machine Learning
  Research}, pp.\  3319--3328. {PMLR}, 2017.

\bibitem[Vaswani et~al.(2017)Vaswani, Shazeer, Parmar, Uszkoreit, Jones, Gomez,
  Kaiser, and Polosukhin]{vaswani2017attention}
Vaswani, A., Shazeer, N., Parmar, N., Uszkoreit, J., Jones, L., Gomez, A.~N.,
  Kaiser, {\L}., and Polosukhin, I.
\newblock Attention is all you need.
\newblock In \emph{Advances in neural information processing systems}, pp.\
  5998--6008, 2017.

\bibitem[Voita et~al.(2019)Voita, Talbot, Moiseev, Sennrich, and
  Titov]{voita-etal-2019-analyzing}
Voita, E., Talbot, D., Moiseev, F., Sennrich, R., and Titov, I.
\newblock Analyzing multi-head self-attention: Specialized heads do the heavy
  lifting, the rest can be pruned.
\newblock In \emph{Proceedings of the 57th Annual Meeting of the Association
  for Computational Linguistics}, pp.\  5797--5808, 2019.

\bibitem[Wallace et~al.(2019)Wallace, Tuyls, Wang, Subramanian, Gardner, and
  Singh]{wallace-etal-2019-allennlp}
Wallace, E., Tuyls, J., Wang, J., Subramanian, S., Gardner, M., and Singh, S.
\newblock {A}llen{NLP} interpret: A framework for explaining predictions of
  {NLP} models.
\newblock In \emph{Proceedings of the 2019 Conference on Empirical Methods in
  Natural Language Processing and the 9th International Joint Conference on
  Natural Language Processing (EMNLP-IJCNLP): System Demonstrations}, pp.\
  7--12, 2019.

\bibitem[Wu \& Ong(2021)Wu and Ong]{wu-ong-2020-explain}
Wu, Z. and Ong, D.~C.
\newblock On explaining your explanations of bert: An empirical study with
  sequence classification.
\newblock \emph{CoRR}, abs/2101.00196, 2021.

\bibitem[Wu et~al.(2018)Wu, Ramsundar, Feinberg, Gomes, Geniesse, Pappu,
  Leswing, and Pande]{wu2018moleculenet}
Wu, Z., Ramsundar, B., Feinberg, E.~N., Gomes, J., Geniesse, C., Pappu, A.~S.,
  Leswing, K., and Pande, V.
\newblock Moleculenet: a benchmark for molecular machine learning.
\newblock \emph{Chemical science}, 9\penalty0 (2):\penalty0 513--530, 2018.

\bibitem[Xiong et~al.(2019)Xiong, Wu, Wang, Kulkarni, Yu, Guo, Chang, and
  Wang]{xiong2019tweetqa}
Xiong, W., Wu, J., Wang, H., Kulkarni, V., Yu, M., Guo, X., Chang, S., and
  Wang, W.~Y.
\newblock Tweetqa: A social media focused question answering dataset.
\newblock In \emph{Proceedings of the 57th Annual Meeting of the Association
  for Computational Linguistics}, pp.\  5020--5031, 2019.

\bibitem[Ying et~al.(2021)Ying, Cai, Luo, Zheng, Ke, He, Shen, and
  Liu]{graphormer}
Ying, C., Cai, T., Luo, S., Zheng, S., Ke, G., He, D., Shen, Y., and Liu, T.-Y.
\newblock Do transformers really perform badly for graph representation?
\newblock In \emph{Thirty-Fifth Conference on Neural Information Processing
  Systems}, 2021.

\bibitem[Ying et~al.(2019)Ying, Bourgeois, You, Zitnik, and
  Leskovec]{ying2019gnnexplainer}
Ying, R., Bourgeois, D., You, J., Zitnik, M., and Leskovec, J.
\newblock Gnnexplainer: Generating explanations for graph neural networks.
\newblock \emph{Advances in neural information processing systems}, pp.\
  9240--9251, 2019.

\bibitem[Yoo et~al.(2020)Yoo, Kim, Lee, Jeong, Choi, Lee, and
  Choi]{yoo2020graph}
Yoo, S.-Y., Kim, Y.-S., Lee, K., Jeong, K., Choi, J., Lee, H., and Choi, Y.~S.
\newblock Graph-aware transformer: Is attention all graphs need?
\newblock \emph{ArXiv}, abs/2006.05213, 2020.

\bibitem[Yun et~al.(2019)Yun, Jeong, Kim, Kang, and Kim]{yun2019GTN}
Yun, S., Jeong, M., Kim, R., Kang, J., and Kim, H.~J.
\newblock Graph transformer networks.
\newblock In \emph{Advances in Neural Information Processing Systems 32: Annual
  Conference on Neural Information Processing Systems 2019}, pp.\
  11960--11970, 2019.

\bibitem[Zhao et~al.(2021)Zhao, Li, Wen, Wang, Liu, Sun, Xie, and
  Ye]{zhao2021gophormer}
Zhao, J., Li, C., Wen, Q., Wang, Y., Liu, Y., Sun, H., Xie, X., and Ye, Y.
\newblock Gophormer: Ego-graph transformer for node classification.
\newblock \emph{arXiv preprint arXiv:2110.13094}, 2021.

\end{thebibliography}
\bibliographystyle{icml2022}

\appendix

\section{Derivations for Attention Heads}
\label{appendix:ah}
\subsection{Gradient of Softmax}
\label{appendix:softmax}
We first consider the softmax function given by:
$$
p_{ij} = \frac{\exp(q_{ij})}{\alpha_j} \quad \alpha_j = \sum_{i} \exp(q_{ij})
$$
where $q_{ij}$ is a shortcut notation for $q(x_i,x'_j)$. We can derive its gradient with respect to the tokens $x_i$ and $x'_j$:
\begin{align*}
\frac{\partial p_{kj}}{\partial x_i}
  &= \frac{1}{\alpha_j^2} \Big(\exp(q_{kj})\frac{\partial q_{kj}}{\partial x_i} \alpha_j - \exp(q_{kj}) \exp(q_{ij}) \frac{\partial q_{ij}}{\partial x_i}\Big)\\
 &= p_{kj} \frac{\partial q_{kj}}{\partial x_i} - p_{kj} p_{ij} \frac{\partial q_{ij}}{\partial x_i}\\
 &= p_{kj} [\delta_{ik} - p_{ij}] \frac{\partial q_{ij}}{\partial x_i}
\intertext{and}
\frac{\partial p_{il}}{\partial x'_j}
  &= \frac{1}{\alpha_l^2} \Big(\exp(q_{il})\frac{\partial q_{il}}{\partial x'_j} \alpha_l \\
  & \hspace{1.4cm} - \exp(q_{il}) \delta_{jl} \sum_k \exp(q_{kl}) \frac{\partial q_{kl}}{\partial x'_j}\Big)\\
  &= p_{il} \frac{\partial q_{il}}{\partial x'_j} - p_{ij} \delta_{jl} \sum_k p_{kj} \frac{\partial q_{kj}}{\partial x'_j}\\
  &= \delta_{jl} p_{ij} \Big(\frac{\partial q_{ij}}{\partial x'_j} - \mathbb{E}_j  \Big[ \frac{\partial q_{:j}}{\partial x'_j}\Big]\Big)
\end{align*}

\subsection{Gradient Propagation Rule for Attention Heads}
\label{appendix:gradprop}

We recall that the attention head stated in Section \ref{section:attention} produces the output
$$
y_j = \sum_i x_i p_{ij}.
$$
We then state the multivariate chain rule for propagating gradients into tokens of the first sequence and resolve local derivatives, making use in particular of the preliminary derivations of Appendix \ref{appendix:softmax}:
\begin{align*}
\frac{\partial f}{\partial x_i} &= \sum_j  \Big(\frac{\partial y_j}{\partial x_i} + \sum_k \frac{\partial p_{kj}}{\partial x_i} \frac{\partial y_j}{\partial p_{kj}}\Big) \frac{\partial f}{\partial y_j}\\
&= \sum_j p_{ij}\frac{\partial f}{\partial y_j} + \sum_{jk}  p_{kj} [\delta_{ik} - p_{ij}] \frac{\partial q_{ij}}{\partial x_i} x_k^\top \frac{\partial f}{\partial y_j}\\
&= \sum_j  p_{ij}\frac{\partial f}{\partial y_j}\ +  \sum_j p_{ij}\frac{\partial q_{ij}}{\partial x_i} (x_i - \mathbb{E}_j [x])^\top \frac{\partial f}{\partial y_j}
\end{align*}
Similarly, we get for the second sequence:
\begin{align*}
\frac{\partial f}{\partial x'_j} &= \sum_l  \sum_i \frac{\partial p_{il}}{\partial x'_j} \frac{\partial y_l}{\partial p_{il}} \frac{\partial f}{\partial y_l}\\
&= \sum_l \sum_i  \delta_{jl} p_{ij} \Big(\frac{\partial q_{ij}}{\partial x'_j} - \mathbb{E}_j  \Big[ \frac{\partial q_{:j}}{\partial x'_j}\Big]\Big) x_i^\top \frac{\partial f}{\partial y_l}\\
&= \sum_i p_{ij} \Big(\frac{\partial q_{ij}}{\partial x'_j} - \mathbb{E}_j  \Big[ \frac{\partial q_{:j}}{\partial x'_j}\Big]\Big) x_i^\top \frac{\partial f}{\partial y_j}\\
&= \mathbb{E}_j\Big[\Big(\frac{\partial q_{:j}}{\partial x'_j} - \mathbb{E}_j  \Big[ \frac{\partial q_{:j}}{\partial x'_j}\Big]\Big) x^\top \Big] \frac{\partial f}{\partial y_j}.
\end{align*}

\subsection{Relevance Conservation}

Computing the relevance for the first sequence
\begin{align*}
\sum_i \mathcal{R}(x_i) &= \sum_i x_i^\top\frac{\partial f}{\partial x_i}\\
&= \sum_j \mathcal{R}(y_j) +  \sum_{ij} p_{ij} x_i^\top \frac{\partial q_{ij}}{\partial x_i}  (x_i  - \mathbb{E}_j [x])^\top \frac{\partial f}{\partial y_j}\\
&= \sum_j \mathcal{R}(y_j) + \sum_j \mathbb{E}_j \big[q_{:j} \cdot (x - \mathbb{E}_j[x])^\top \big]  \frac{\partial f}{\partial y_j}
\end{align*}
and for the second sequence
\begin{align*}
\sum_j \mathcal{R}(x'_j) &= \sum_j {x'}_j^\top\frac{\partial f}{\partial x'_j}\\
&= \sum_j {x'}_j^\top \mathbb{E}_j\Big[\Big(\frac{\partial q_{:j}}{\partial x'_j} - \mathbb{E}_j  \Big[ \frac{\partial q_{:j}}{\partial x'_j}\Big]\Big) x^\top \Big] \frac{\partial f}{\partial y_j}\\
&= \sum_j \mathbb{E}_j\big[(q_{:j} - \mathbb{E}_j  [ q_{:j}]) x^\top \big] \frac{\partial f}{\partial y_j},
\end{align*}
and summing the two scores, we obtain:
\begin{align*}
&\sum_i \mathcal{R}(x_i) + \sum_j \mathcal{R}(x'_j)\\
& \qquad = \sum_j \mathcal{R}(y_j) + \sum_j \Big(\mathbb{E}_j\big[(q_{:j} - \mathbb{E}_j  [ q_{:j}]) x^\top \big] \\ &\qquad \qquad+ \mathbb{E}_j \big[q_{:j} \cdot (x - \mathbb{E}_j[x])^\top \big] \Big)  \frac{\partial f}{\partial y_j}
\end{align*}
Making the further assumption that the first sequence of tokens and the softmax input both have expected value zero, we obtain the simplified form:
\begin{align*}
&\sum_i \mathcal{R}(x_i) + \sum_j \mathcal{R}(x'_j)\\
&\qquad = \sum_j \mathcal{R}(y_j) + \sum_j 2\,\text{Cov}_j(q_{:j},x)^\top \frac{\partial f}{\partial y_j}
\end{align*}

\section{Derivations for LayerNorm}
\label{appendix:ln}

The core part of LayerNorm can be decomposed into two parts:
\begin{align}
\widetilde{x}_i &=  x_i - \mathbb{E}[x]\\
y_i &= \frac{\widetilde{x}_i}{\sqrt{\epsilon + \mathbb{E}[\widetilde{x}^2]}}
\end{align}
i.e.\ a centering followed by a rescaling, where $\mathbb{E}[\cdot]$ is computed over a uniform distribution, i.e.\ $\mathbb{E}[x] = \frac1N \sum_i x_i$.

\subsection{Centering step}

We first analyze GI relevance conservation in the centering step. For this, we make use of Eq.\ \eqref{eq:girule} of the main paper, and perform a summation over input variables on both sides:
\begin{align}
\sum_i \mathcal{R}(x_i) = \sum_i \sum_j \frac{\partial \widetilde{x}_j}{\partial x_i} \frac{x_i}{\widetilde{x}_j} \mathcal{R}(\widetilde{x}_j)
\label{eq:girule-appendix}
\end{align}
Observing that $\widetilde{x}_j(x)$ is a homogeneous function of degree one, and that consequently $\sum_i x_i \frac{\partial \widetilde{x}_j}{\partial x_i} = \widetilde{x}_j$ for all $j$, 
Eq.\ \eqref{eq:girule-appendix} reduces to
\begin{align}
\sum_i \mathcal{R}(x_i) &= \sum_i \mathcal{R}(\widetilde{x}_i)
\end{align}

\subsection{Rescaling step}

We now analyze conservation in the rescaling part of LayerNorm. The local derivative is given by:
\begin{align*}
\frac{\partial y_j}{\partial \widetilde{x}_i}
&= \frac{\delta_{ij}}{\sqrt{\epsilon + \mathbb{E}[\widetilde{x}^2]}} - \frac{\widetilde{x}_j \frac{\partial}{\partial \widetilde{x}_i} \big(\sqrt{\epsilon + \mathbb{E}[\widetilde{x}^2]}\big)}{\big(\sqrt{\epsilon + \mathbb{E}[\widetilde{x}^2]}\big)^2}\\
&= \frac{\delta_{ij}}{\sqrt{\epsilon + \mathbb{E}[\widetilde{x}^2]}} - \frac{\frac1N \widetilde{x}_i\widetilde{x}_j}{\big(\sqrt{\epsilon + \mathbb{E}[\widetilde{x}^2]}\big)^3}
\end{align*}
We can now analyze conservation by applying Eq.\ \eqref{eq:girule} of the paper:
\begin{align*}
\sum_i \mathcal{R}(\widetilde{x}_i) &=\sum_i \sum_j \frac{\partial y_j}{\partial \widetilde{x}_i} \frac{\widetilde{x}_i}{y_j} \mathcal{R}(y_j)\\
&=\sum_i \sum_j \bigg( \frac{\delta_{ij}}{\sqrt{\epsilon + \mathbb{E}[\widetilde{x}^2]}} \\
& \hspace{2.3cm} - \frac{\frac1N \widetilde{x}_i\widetilde{x}_j}{\big(\sqrt{\epsilon + \mathbb{E}[\widetilde{x}^2]}\big)^3} \bigg) \frac{\widetilde{x}_i}{y_j} \mathcal{R}(y_j)\\
&= \sum_i \sum_j \big( \delta_{ij} y_i - \frac1N y_i y_i y_j\big) \cdot \frac{1}{y_j} \mathcal{R}(y_j)\\
&= \sum_j \big(1 - \mathbb{E}[y^2]\big) \mathcal{R}(y_j)\\
&= \Big(1 - \frac{\mathbb{E}[\widetilde{x}^2]}{\epsilon + \mathbb{E}[\widetilde{x}^2]} \Big) \sum_j \mathcal{R}(y_j)
\end{align*}
Hence, after combining with the centering step, we obtain the conservation equation:
\begin{align}
\sum_i \mathcal{R}(x_i) &= \Big(1 - \frac{\mathrm{Var}[x]}{\epsilon + \mathrm{Var}[x]} \Big) \sum_i \mathcal{R}(y_i)
\end{align}
where conservation holds only approximately for large values of $\epsilon$.

\section{Implementation Details}\label{app:implementation}

 For SST-2 and IMDB sentiment classification, the embeddings module and the tokenizer are initialized from pre-trained BERT-Transformers (textattack/bert-base-uncased-\{sst-2/imdb\}). 
 We report the following test accuracies after optimization: $81.3\%$ for SST-2,  $88.1\%$ on IMDB. For training, we use batchsizes of $bs=32$ and optimize the model parameters using the AdamW optimizer with a learning rate of $lr=2e{-}5$ for a maximal number of $T=20$ epochs or until early stopping for decreasing validation performance is reached. For the other NLP tasks we follow the same settings, except that we initialize the embedding model and the tokenizer from a HuggingFace pre-trained BERT-Transformer (bert-base-uncased).
 
 For the BACE and MNIST-Superpixels datasets, the recent Graphormer model is used. We train a 2-layer graphormer model with a batchsize of $bs=64$ and AdamW as an optimizer with $lr=2e{-}4$. The model is trained for $1000$ epochs for the BACE dataset and 10 epochs for MNIST-Superpixels, achieving an accuracy of  $97.2\%$ for MNIST, and a ROC-ACC of $83.5\%$ for BACE.
 
 \section{Additional Conservation Experiments}
 \label{app:conservation}
\begin{figure}[h]
    \centering
     \includegraphics[width=.3\linewidth]{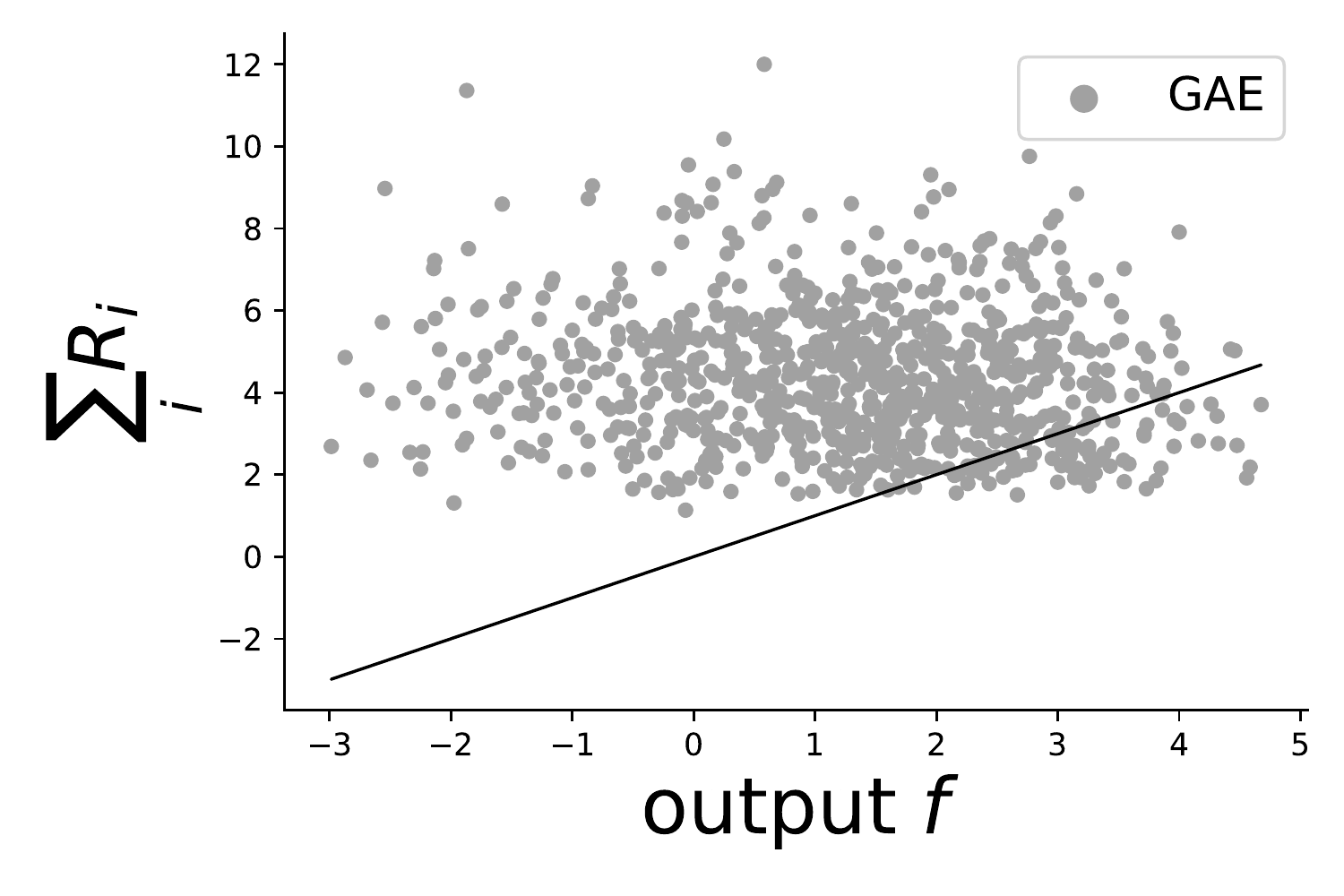}
     \includegraphics[width=.3\linewidth]{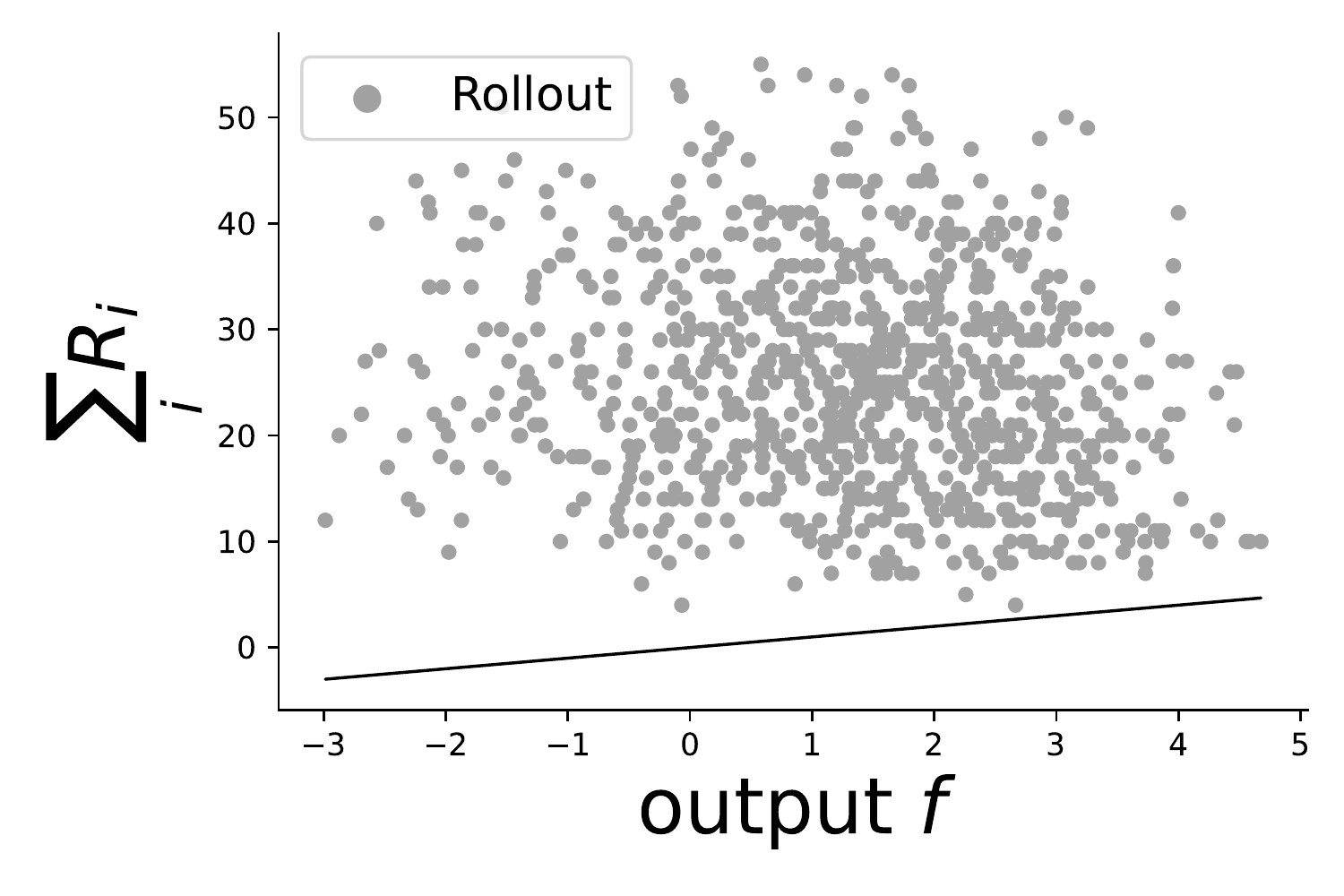}
     \includegraphics[width=.3\linewidth]{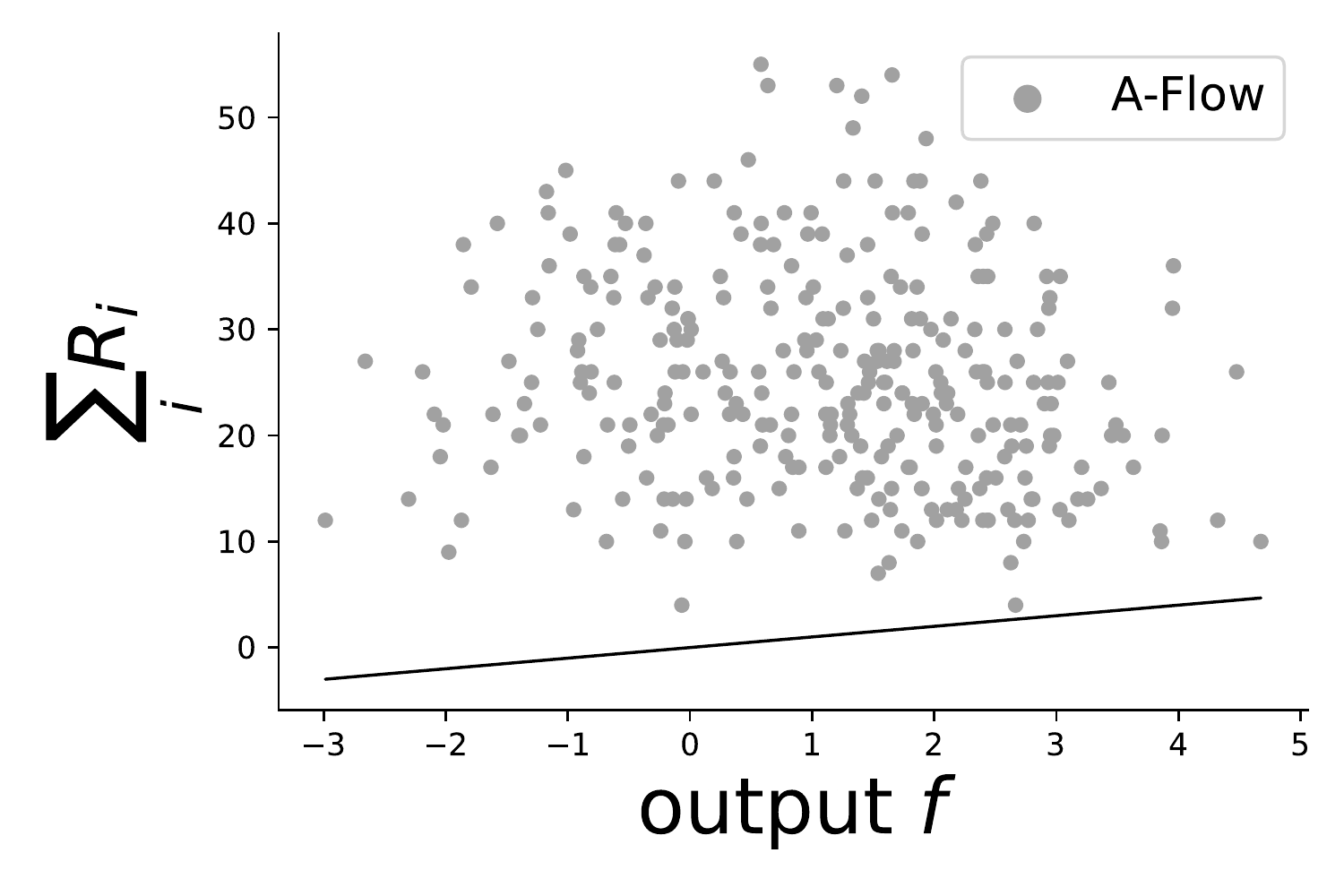}
    \caption{Conservation experiment for additional explainability baselines used in the experiments.}
    \label{fig:baselinesconservation}
\end{figure}

 In this section, we show the conservation experiment results for three additional baselines (GAE, Rollout and A-Flow). Results are shown in Fig.\ \ref{fig:baselinesconservation} for the SST-2 dataset. For A-Flow we use $30\%$ of the dataset. We observe that the conservation does not hold for these baseline explanation methods.
\vspace{-2mm}

\section{Runtime Analysis}
 \label{app:timeanalysis}
In this section, we report the time needed for each explanation method to produce an explanation.
Wall-clock elapsed time for running each method averaged over all of the samples on the SST-2 dataset (results are reported in seconds): random $6e{-}5$, A-Last $0.0002$, Rollout $0.0004$, GAE $0.016$, GI $0.017$, LRP(AH) $0.014$, LRP(AH + LN) $0.012$. (In comparison, prediction takes $0.024$.) Our method is competitive with SOTA explanation techniques.
\onecolumn
\section{Additional Qualitative Results on SST-2}\label{app:sentiment}

\begin{tabular}{ll} 
  \centering
  \includegraphics[width=0.35\textwidth]{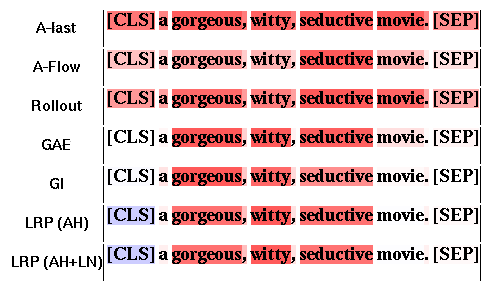} & \includegraphics[width=0.40\textwidth]{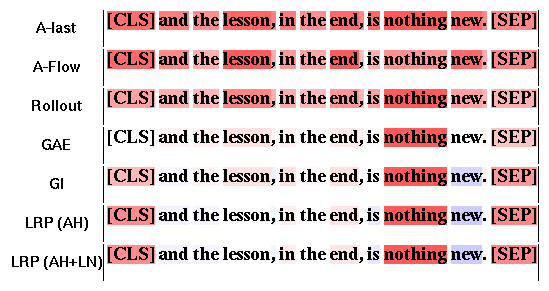}   \\
  \includegraphics[width=0.35\textwidth]{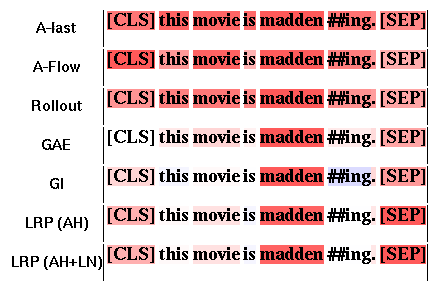}  &  \includegraphics[width=0.35\textwidth]{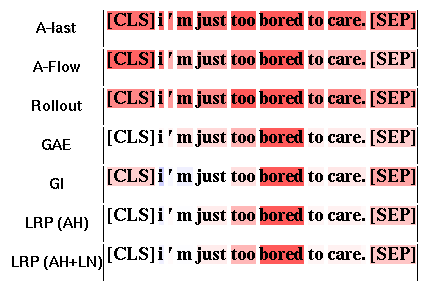}  \\
  \includegraphics[width=0.35\textwidth]{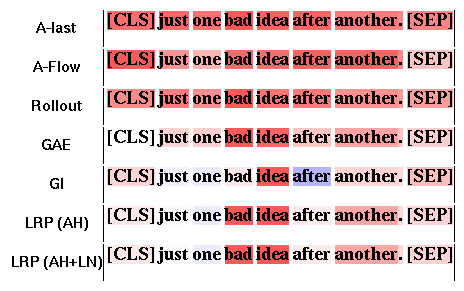}  &  \includegraphics[width=0.5\textwidth]{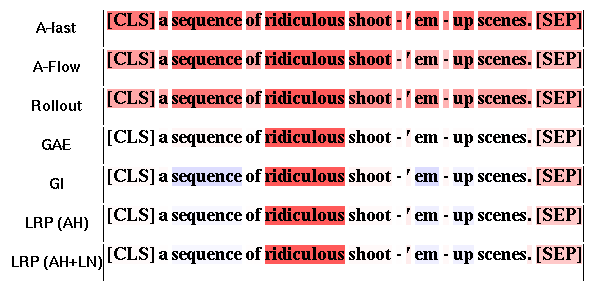} \\
  \includegraphics[width=0.35\textwidth]{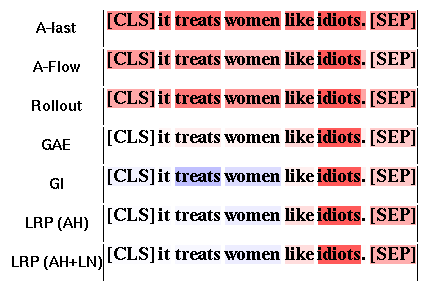} & 
  \includegraphics[width=0.5\textwidth]{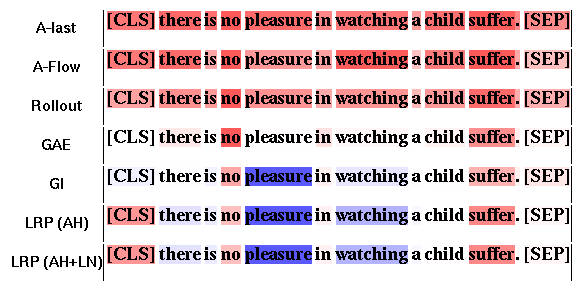}
\end{tabular}
\newpage
\section{Additional Qualitative Results on MNIST Superpixels}\label{app:mnist}

    \begin{tabular}{@{}cccc}
        A-Last & Rollout & GAE \\

        \includegraphics[width=0.14\textwidth]{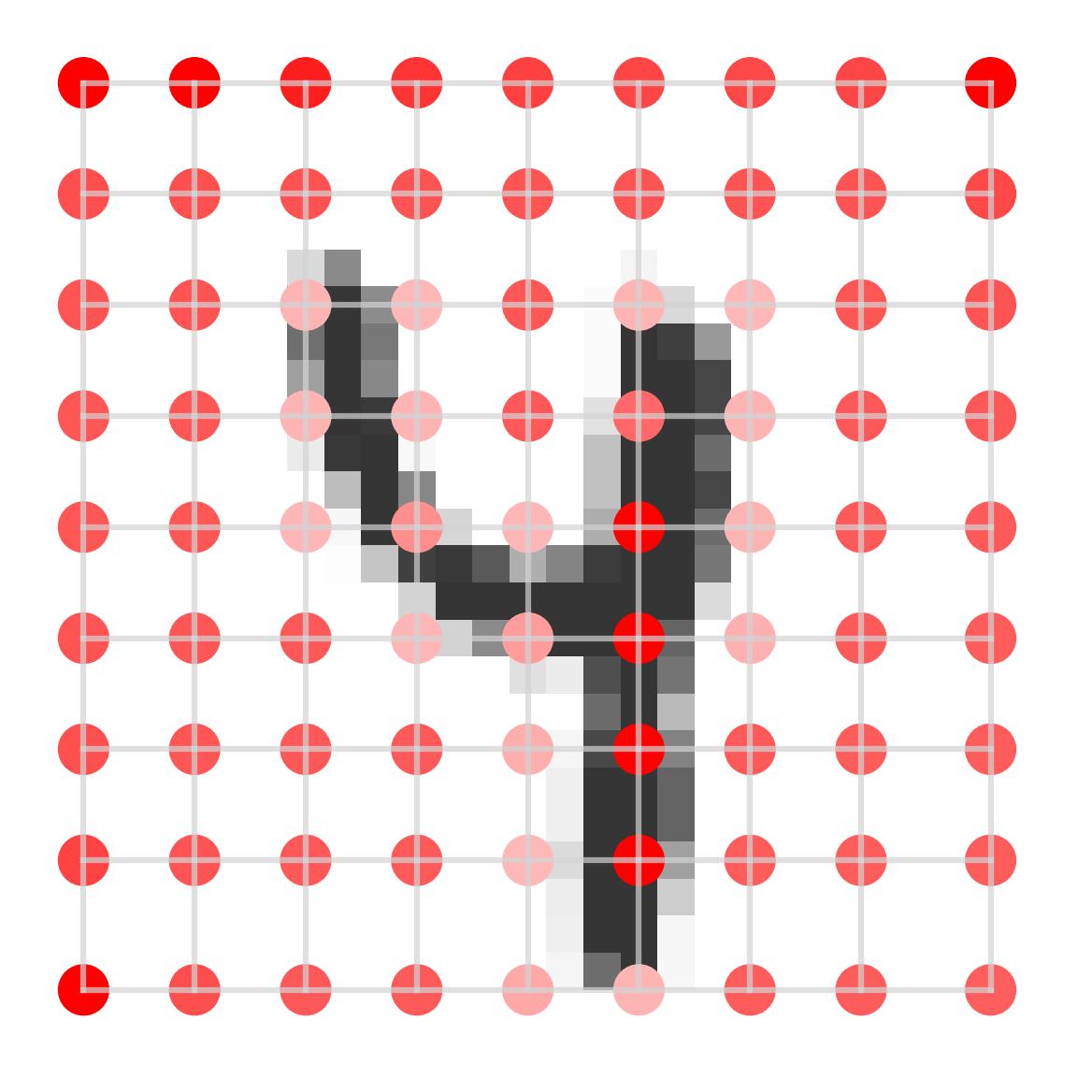}
        &
        \includegraphics[width=0.14\textwidth]{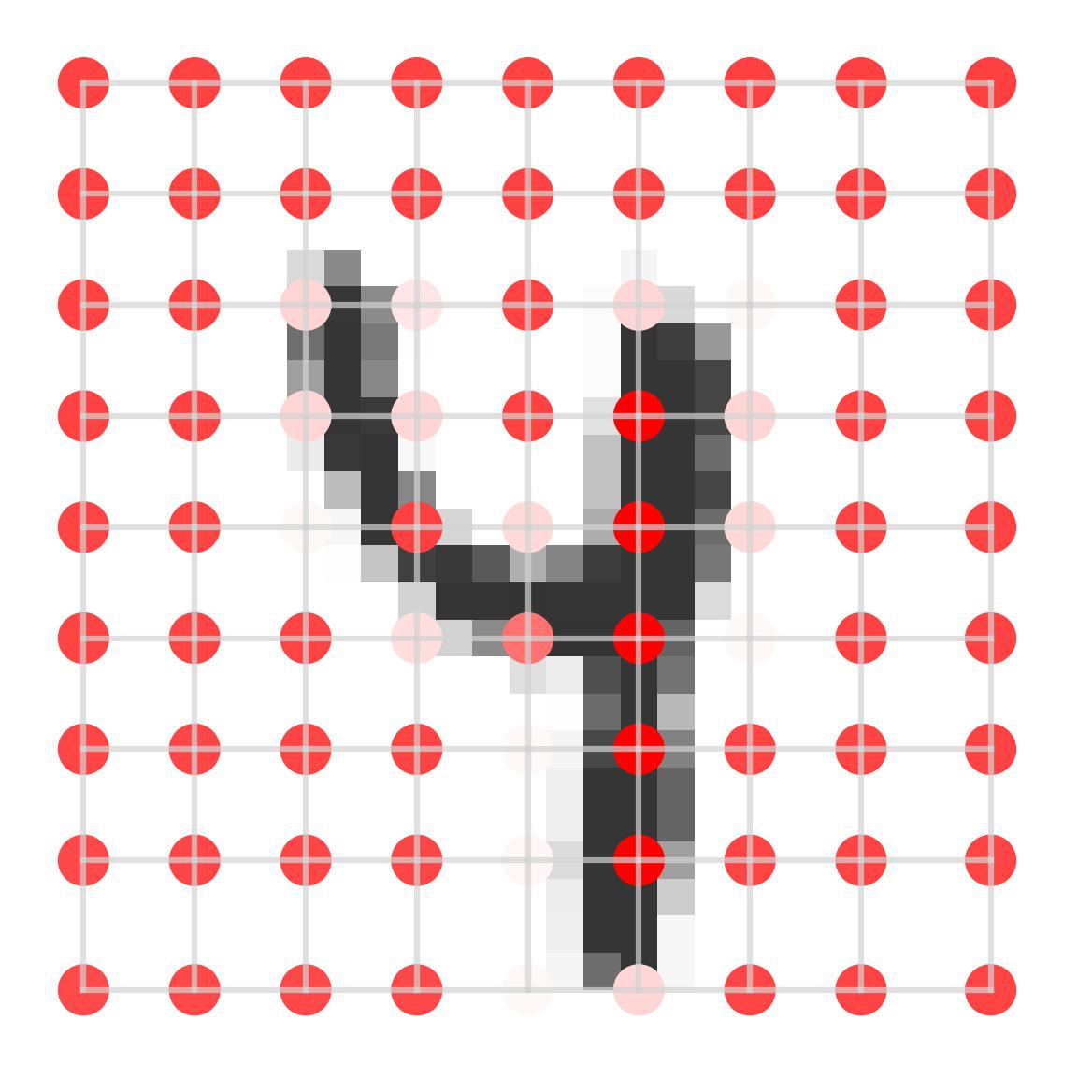}
        &
        \includegraphics[width=0.14\textwidth]{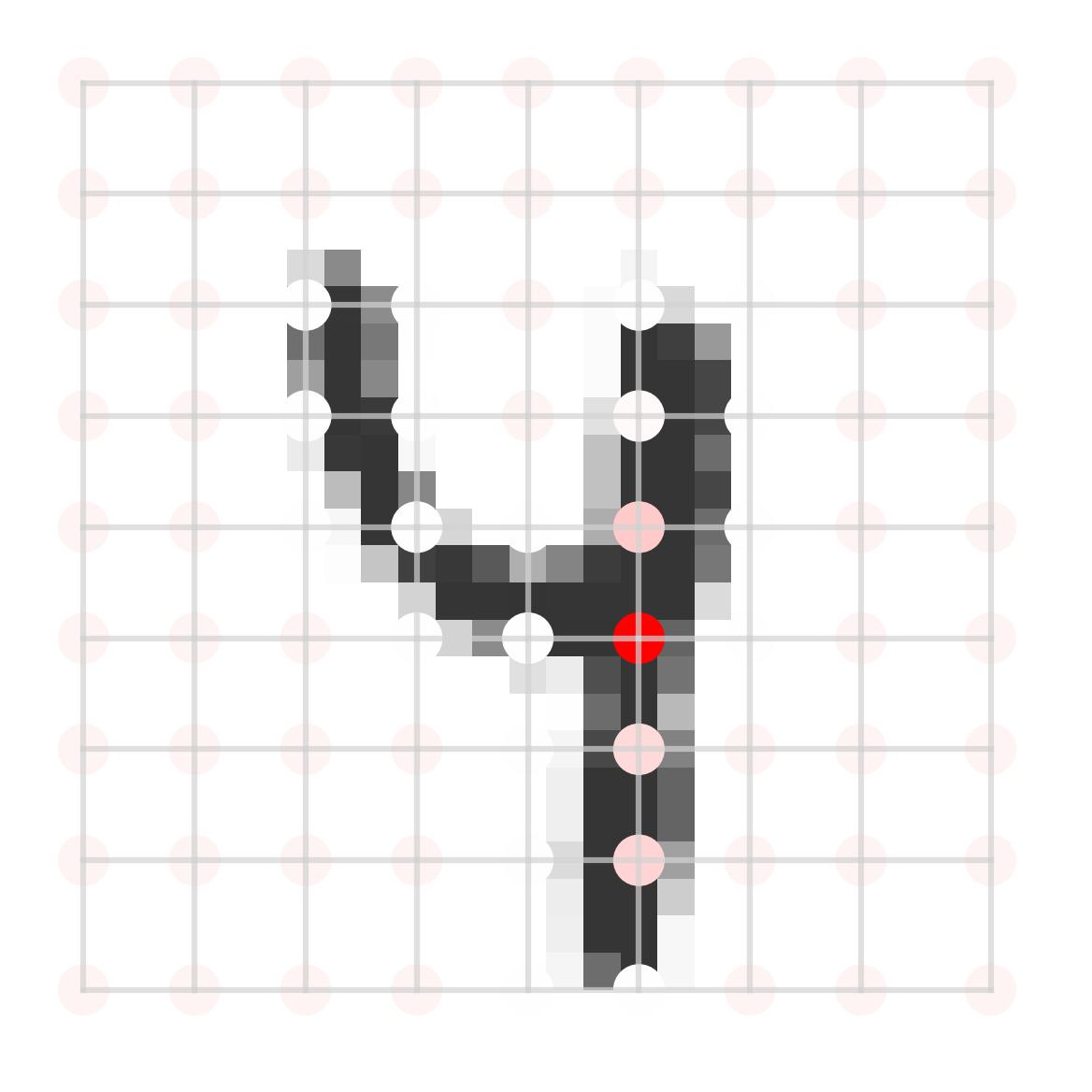}   
         \\
         GI  & LRP (AH) & LRP (AH + LN)   \\
        \includegraphics[width=0.14\textwidth]{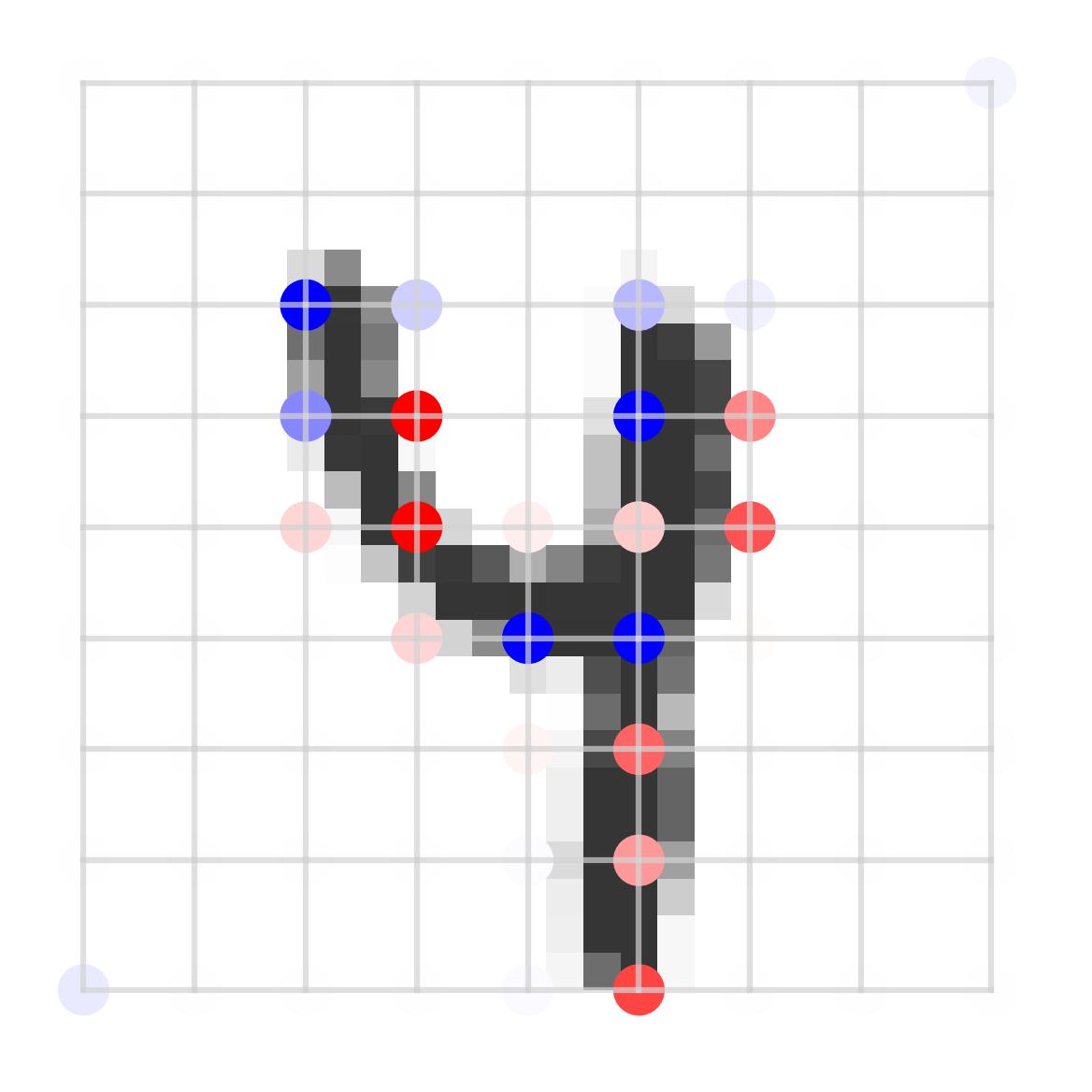}
        &
        \includegraphics[width=0.14\textwidth]{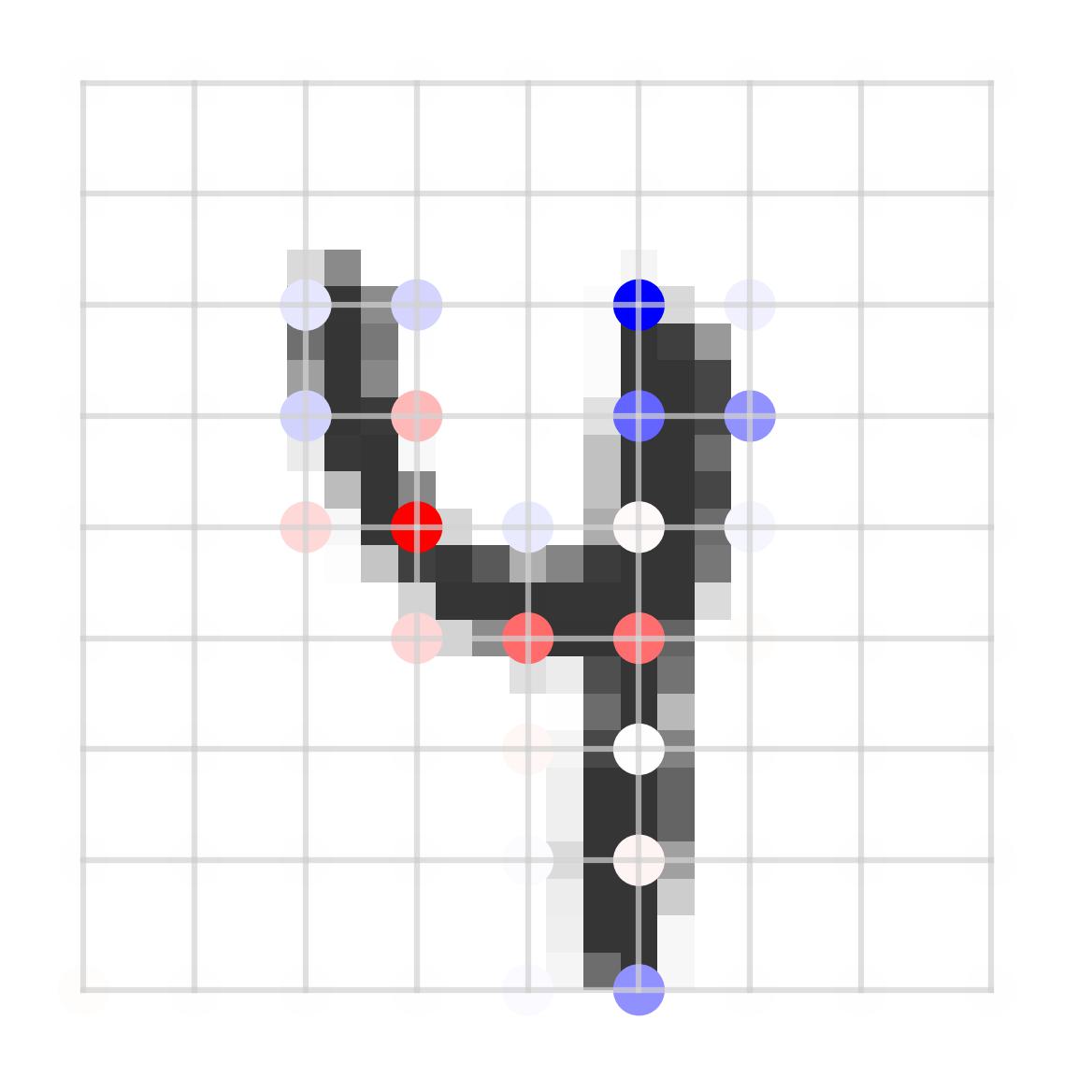}
        &
        \includegraphics[width=0.14\textwidth]{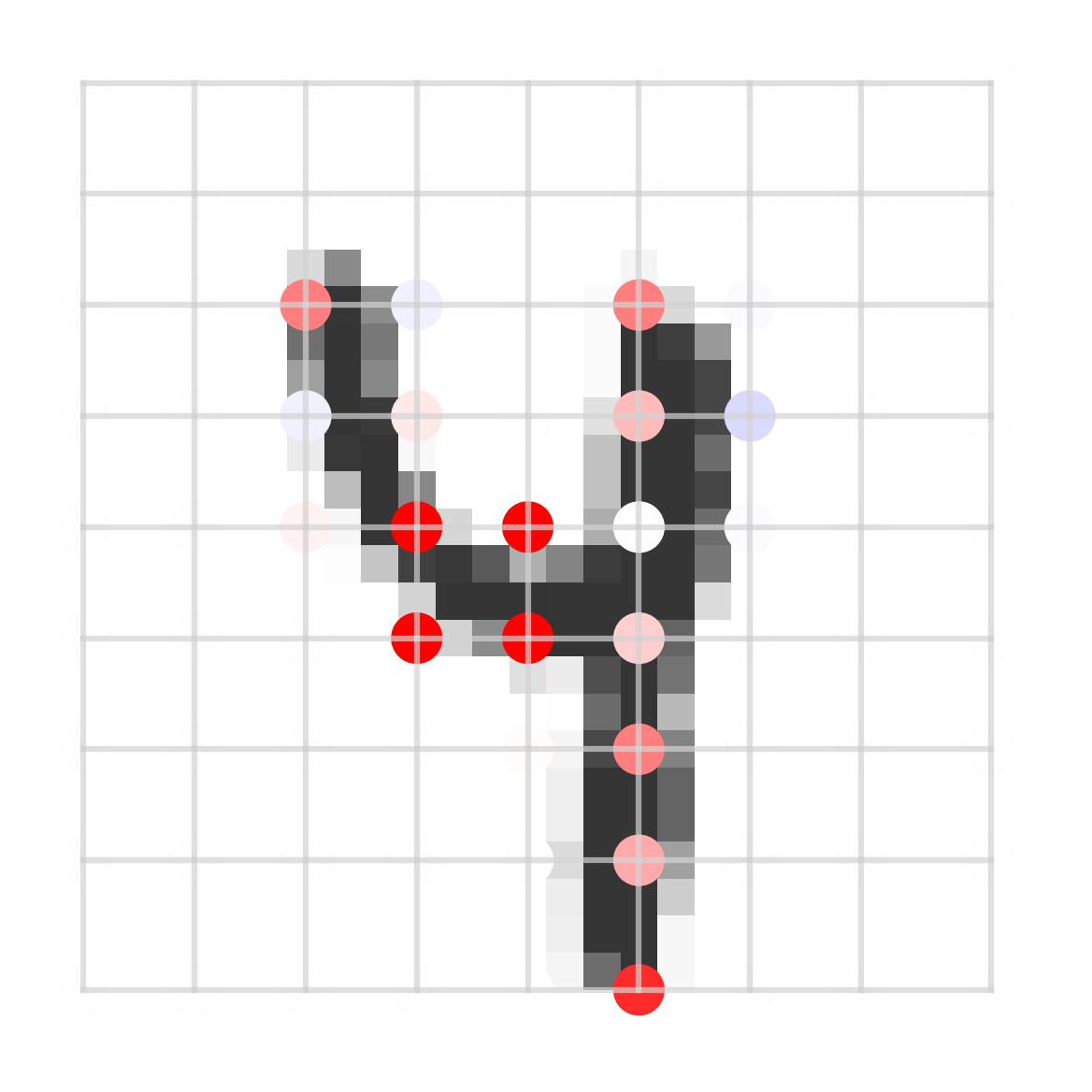}
        &
        \\
        &&& \\

        A-Last & Rollout & GAE \\

        \includegraphics[width=0.14\textwidth]{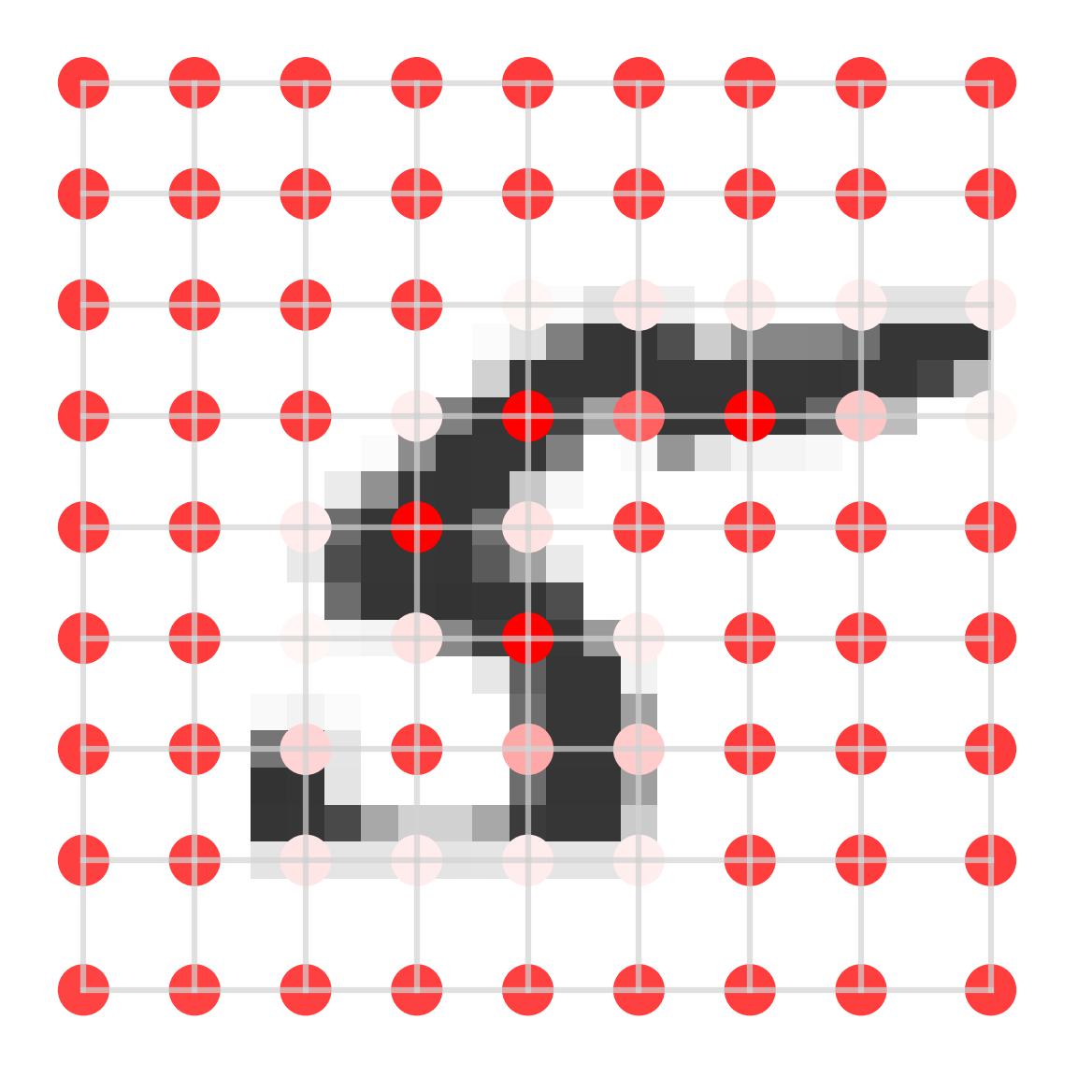}
        &
        \includegraphics[width=0.14\textwidth]{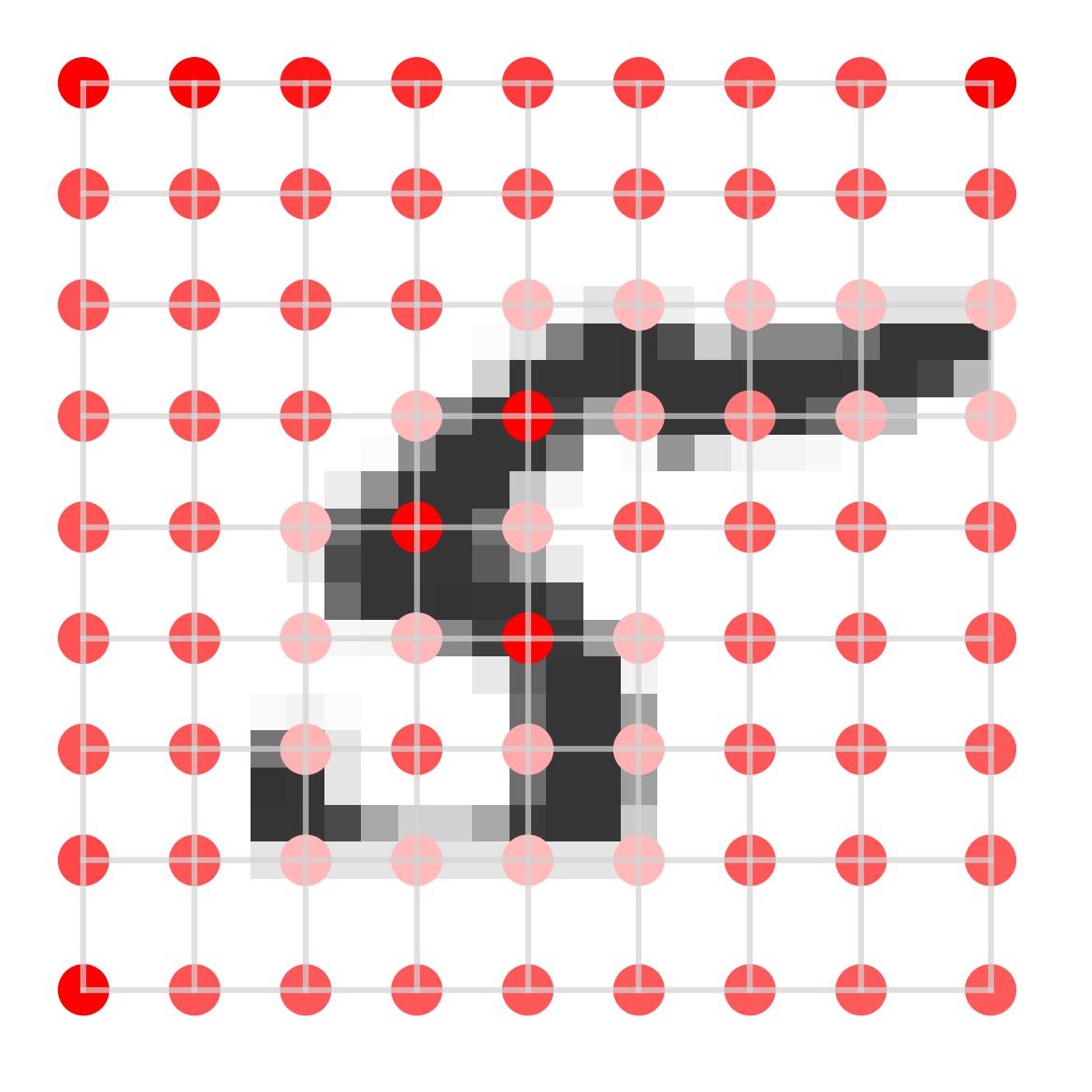}
        &
        \includegraphics[width=0.14\textwidth]{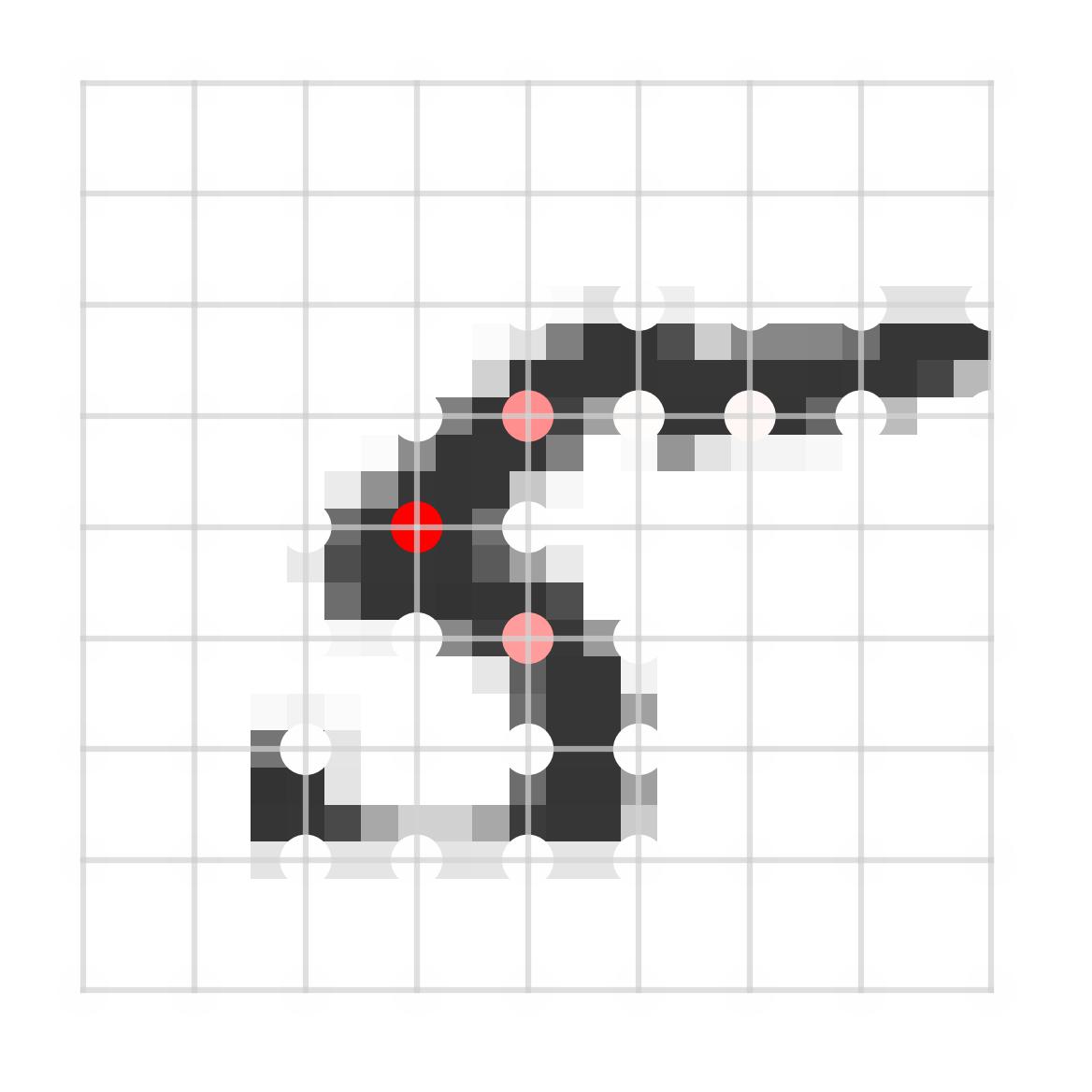}
         \\
         GI &  LRP (AH) & LRP (AH + LN)   \\
        \includegraphics[width=0.14\textwidth]{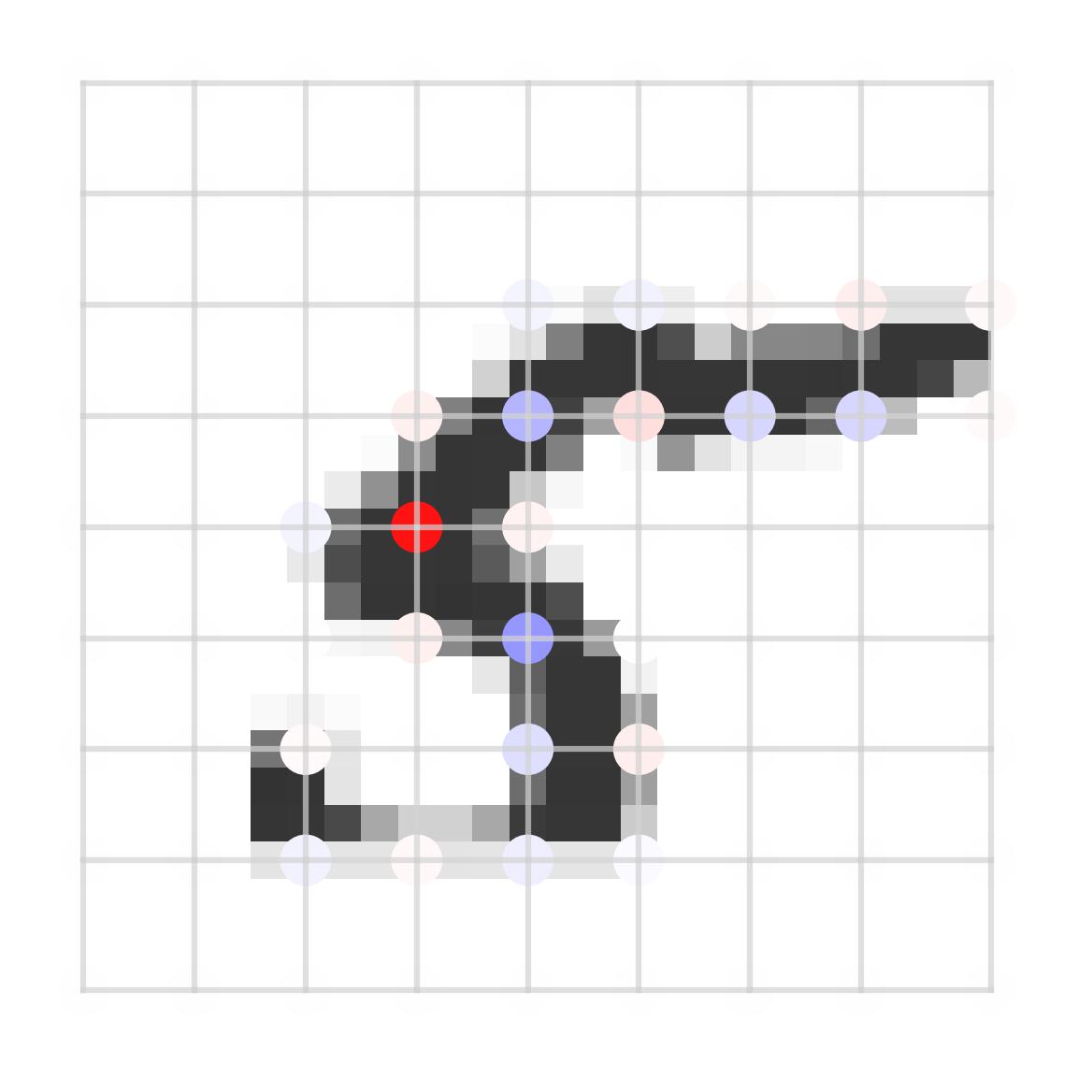}
        &
        \includegraphics[width=0.14\textwidth]{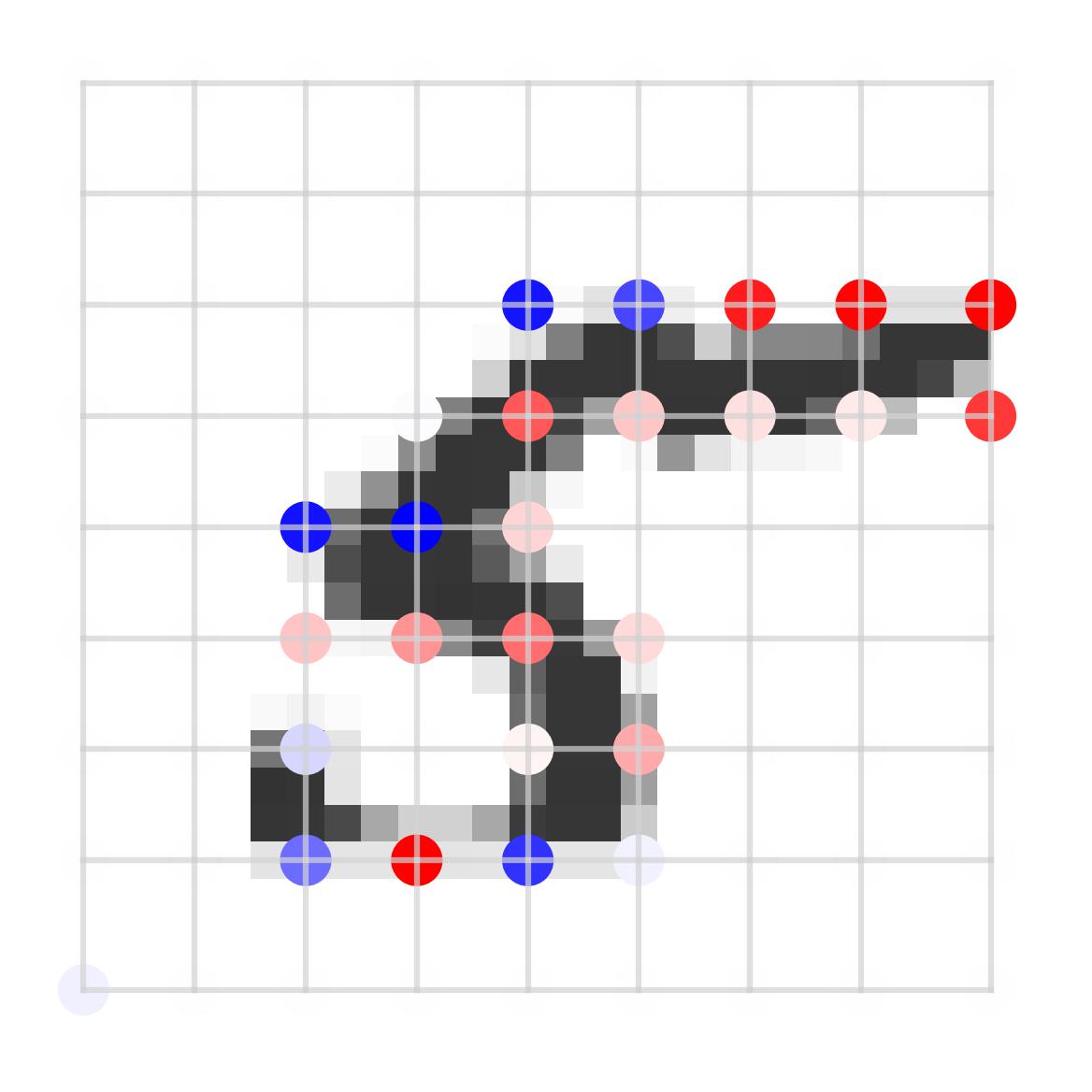} &
        \includegraphics[width=0.14\textwidth]{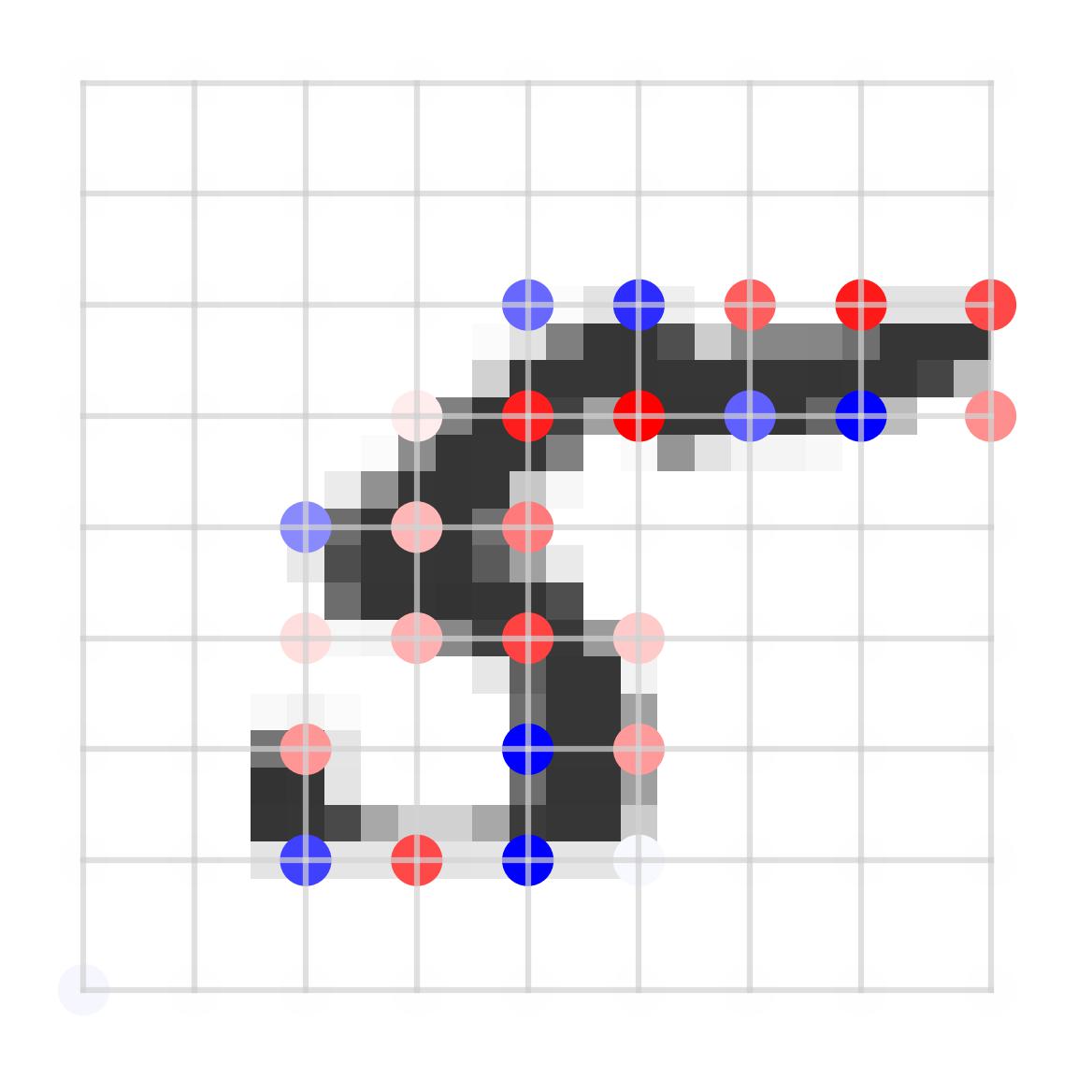} &
        \\
           &&& \\
         A-Last & Rollout & GAE \\

        \includegraphics[width=0.14\textwidth]{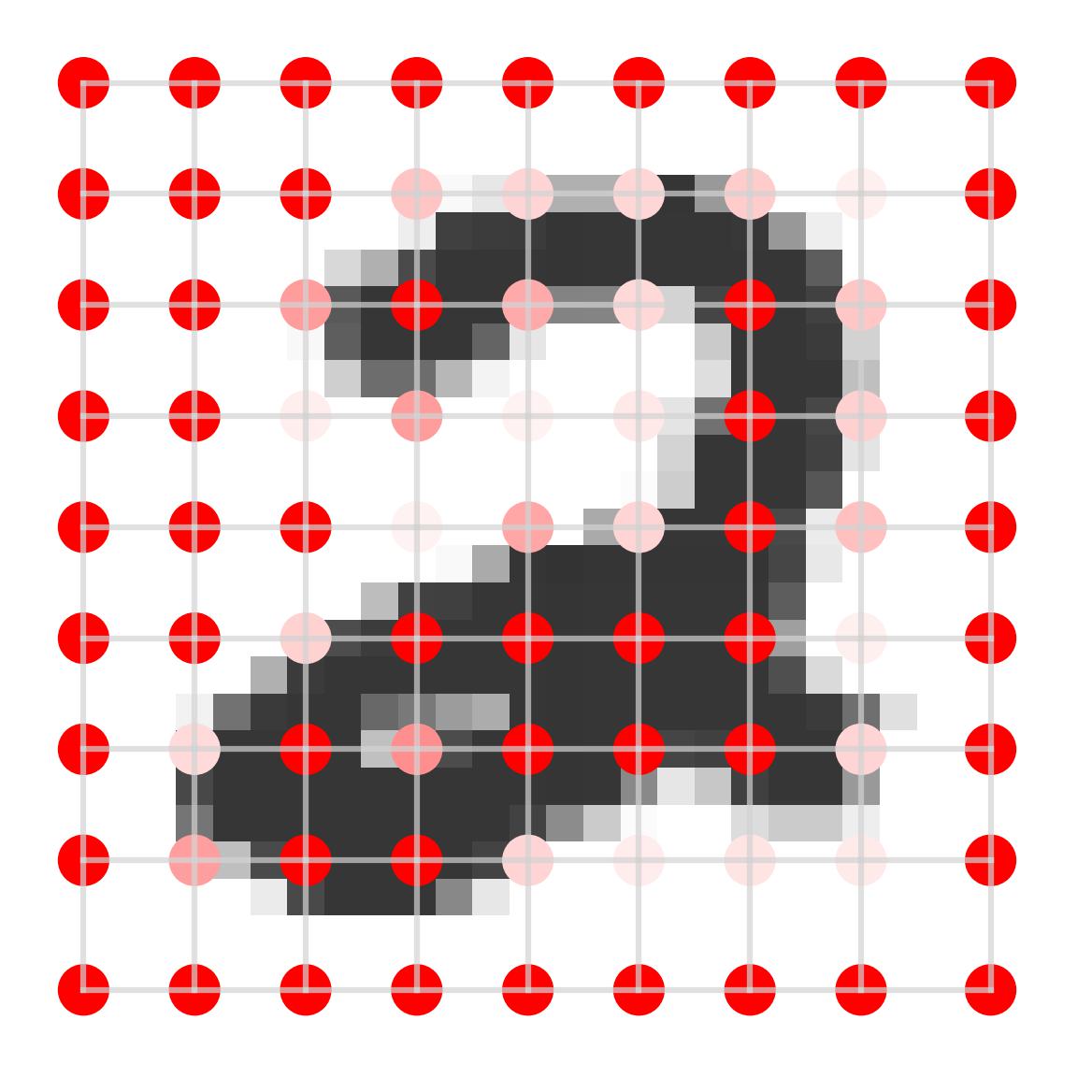}
        &
        \includegraphics[width=0.14\textwidth]{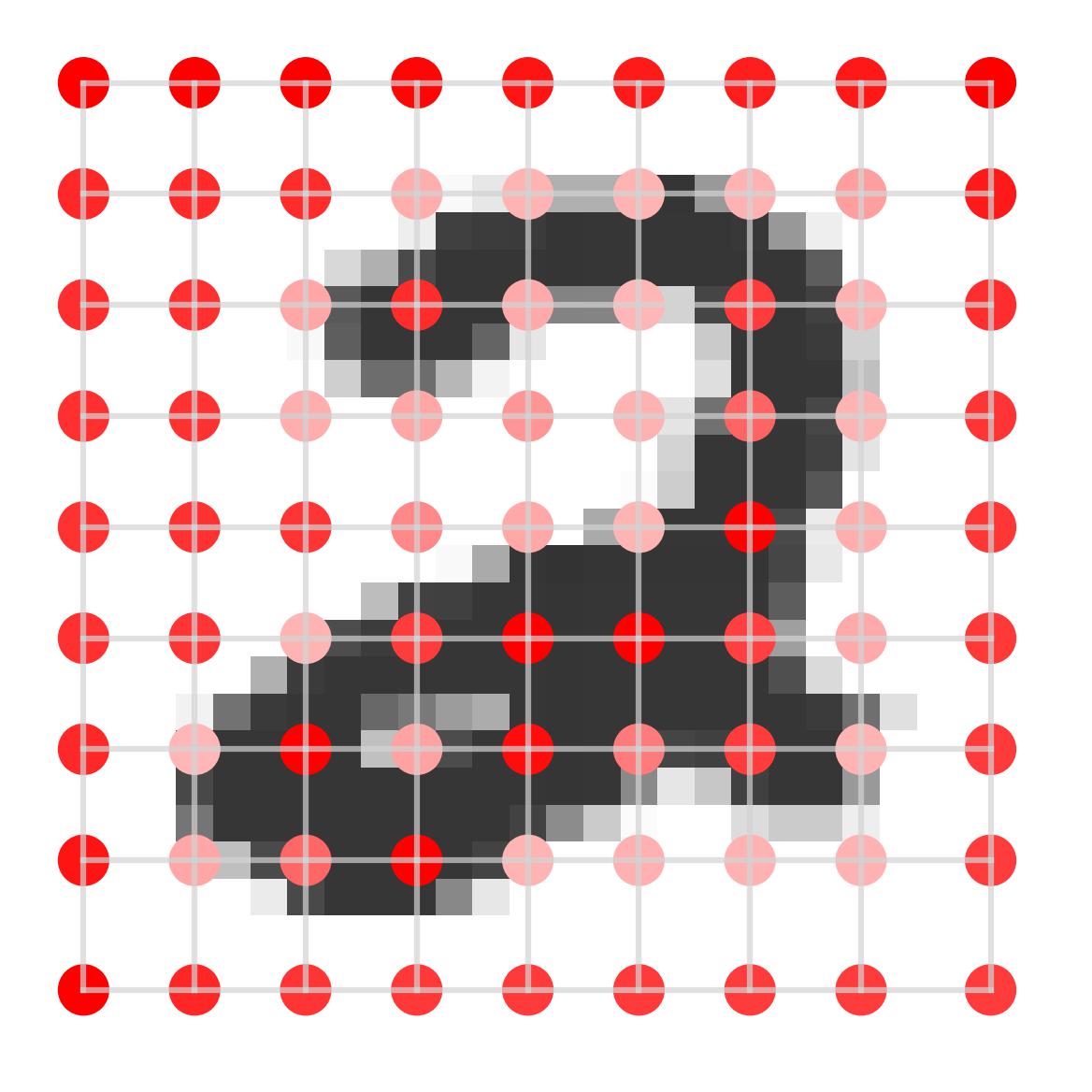}
        &
        \includegraphics[width=0.14\textwidth]{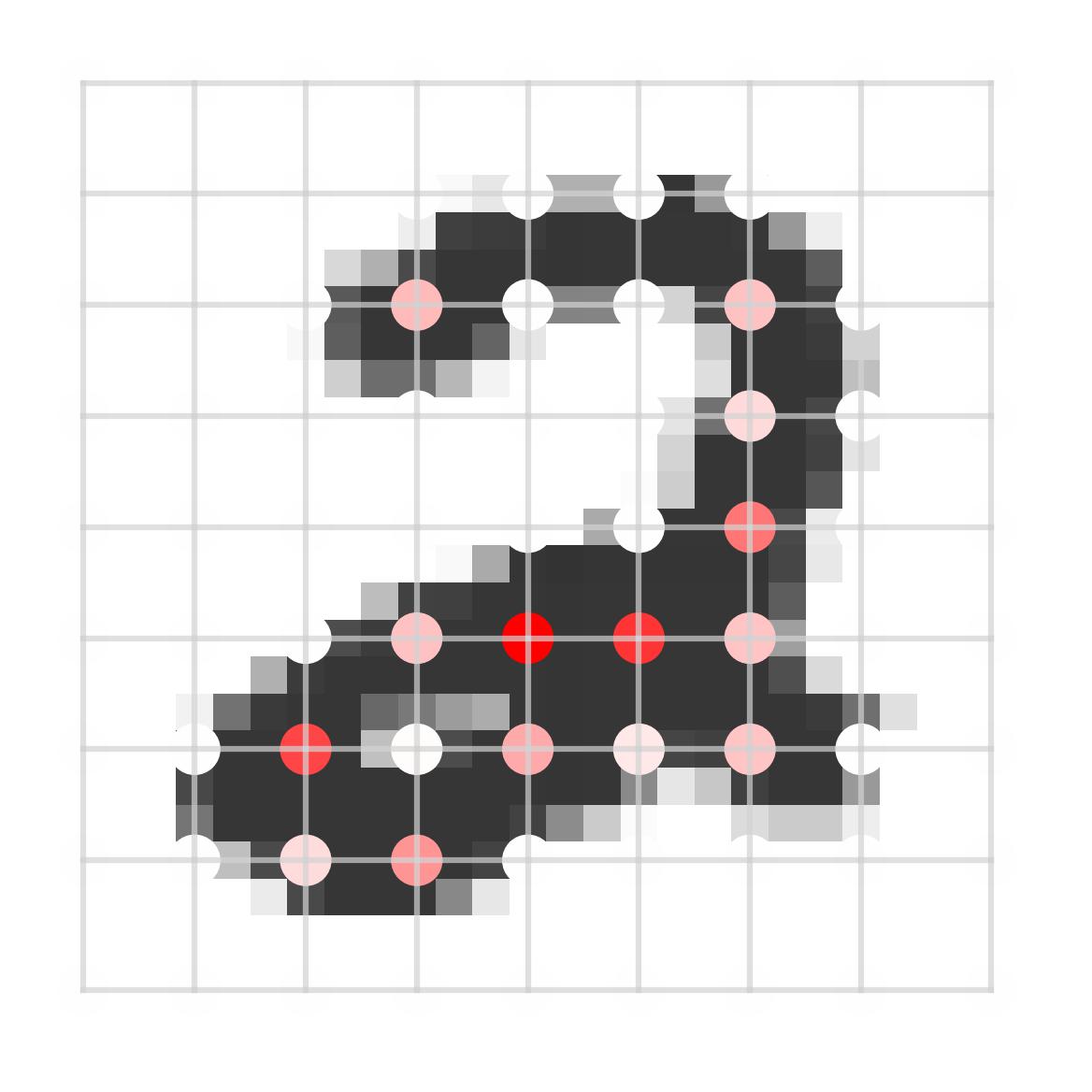}
         \\
         GI &  LRP (AH) & LRP (AH + LN)   \\
        \includegraphics[width=0.14\textwidth]{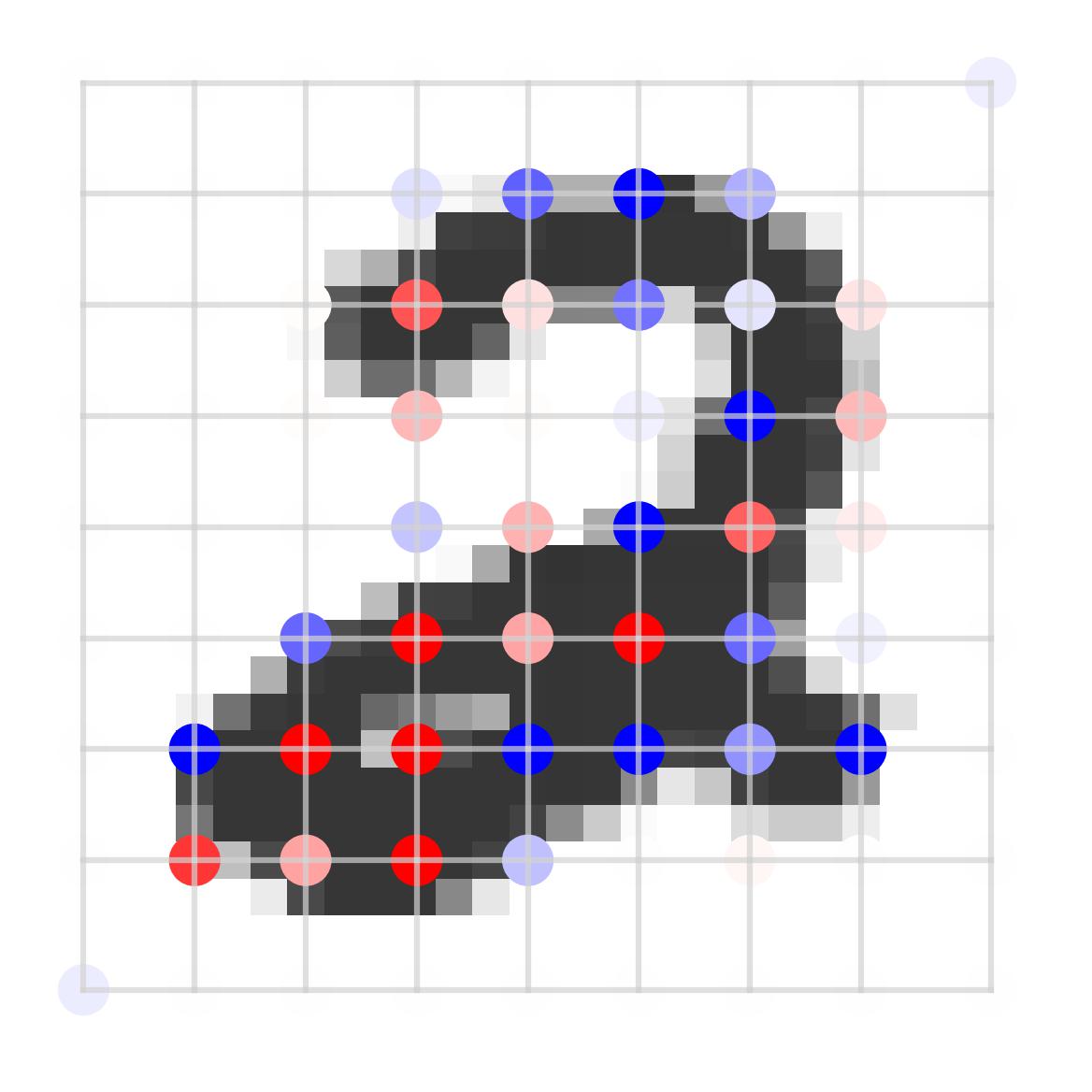}
        &
        \includegraphics[width=0.14\textwidth]{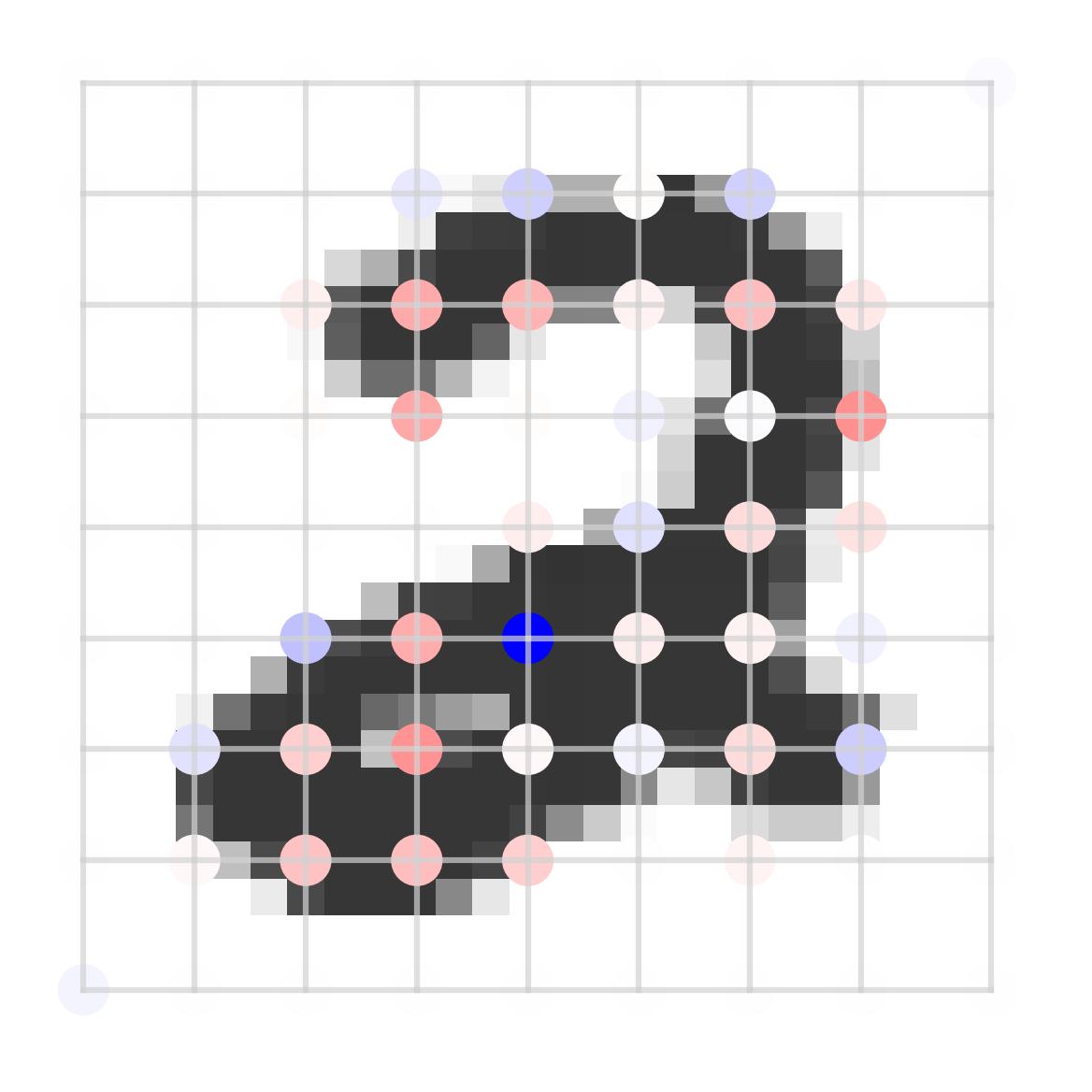}
 &
         \includegraphics[width=0.14\textwidth]{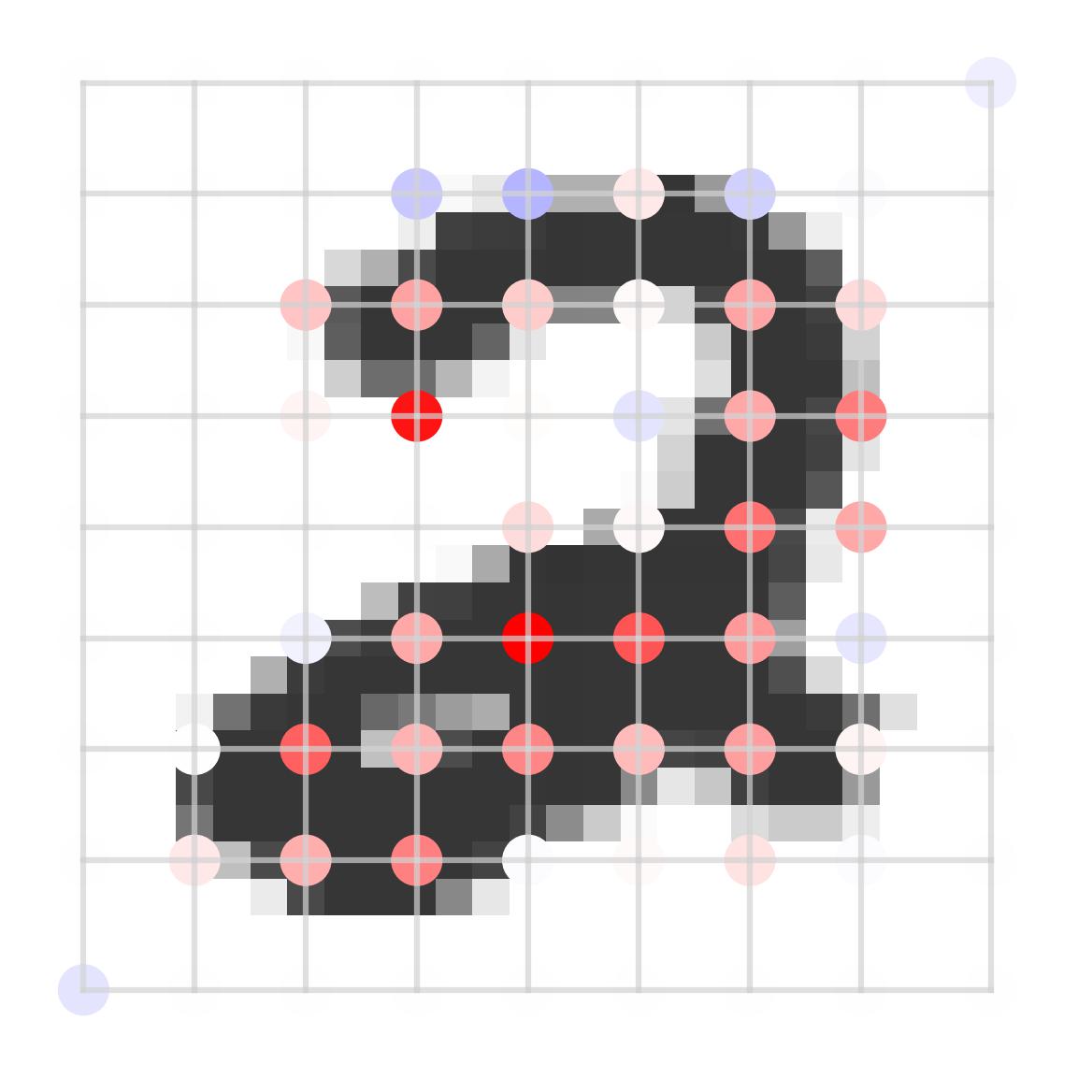}
\\
    \end{tabular} 
        \begin{tabular}{@{}cccc}
        A-Last & Rollout & GAE \\

        \includegraphics[width=0.14\textwidth]{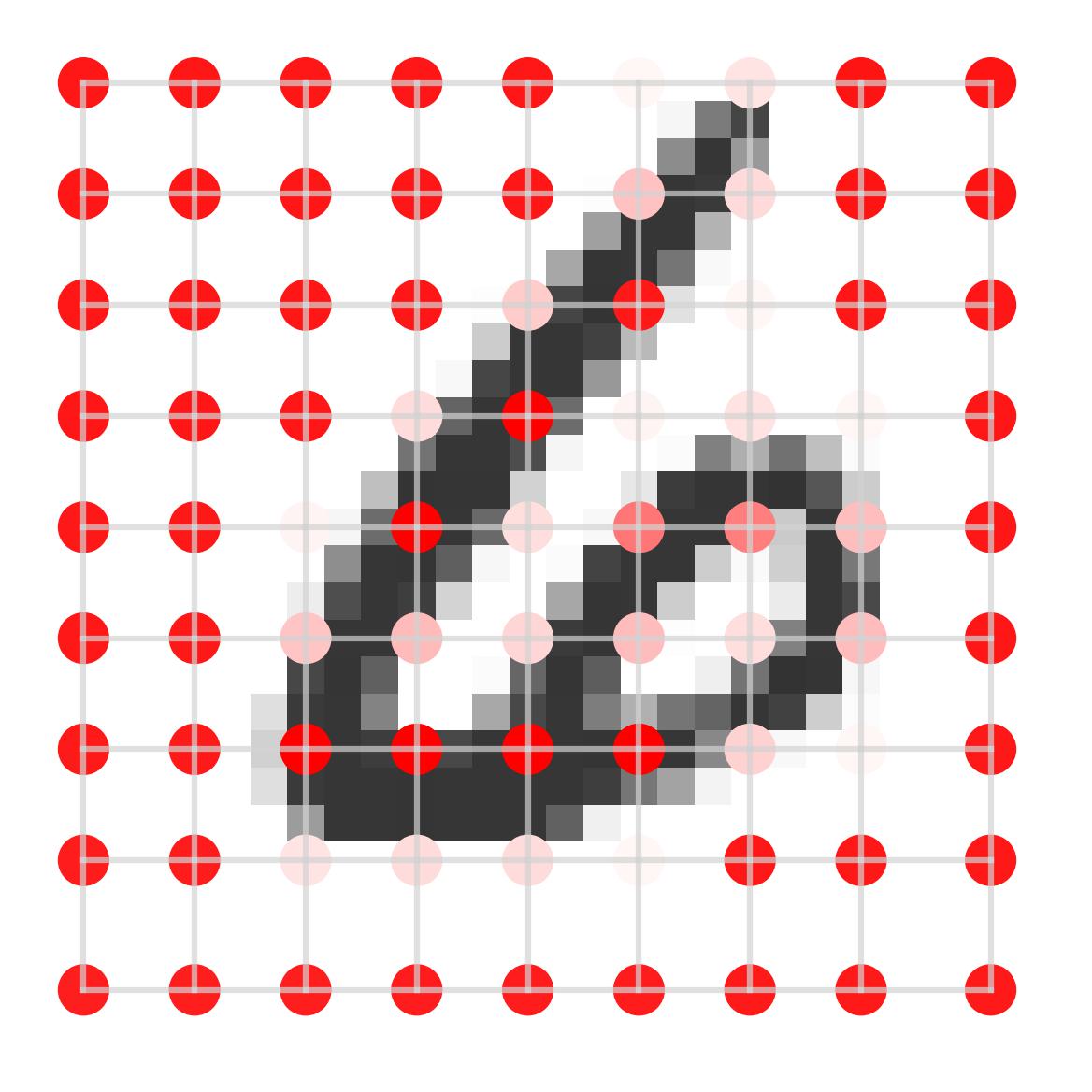}
        &
        \includegraphics[width=0.14\textwidth]{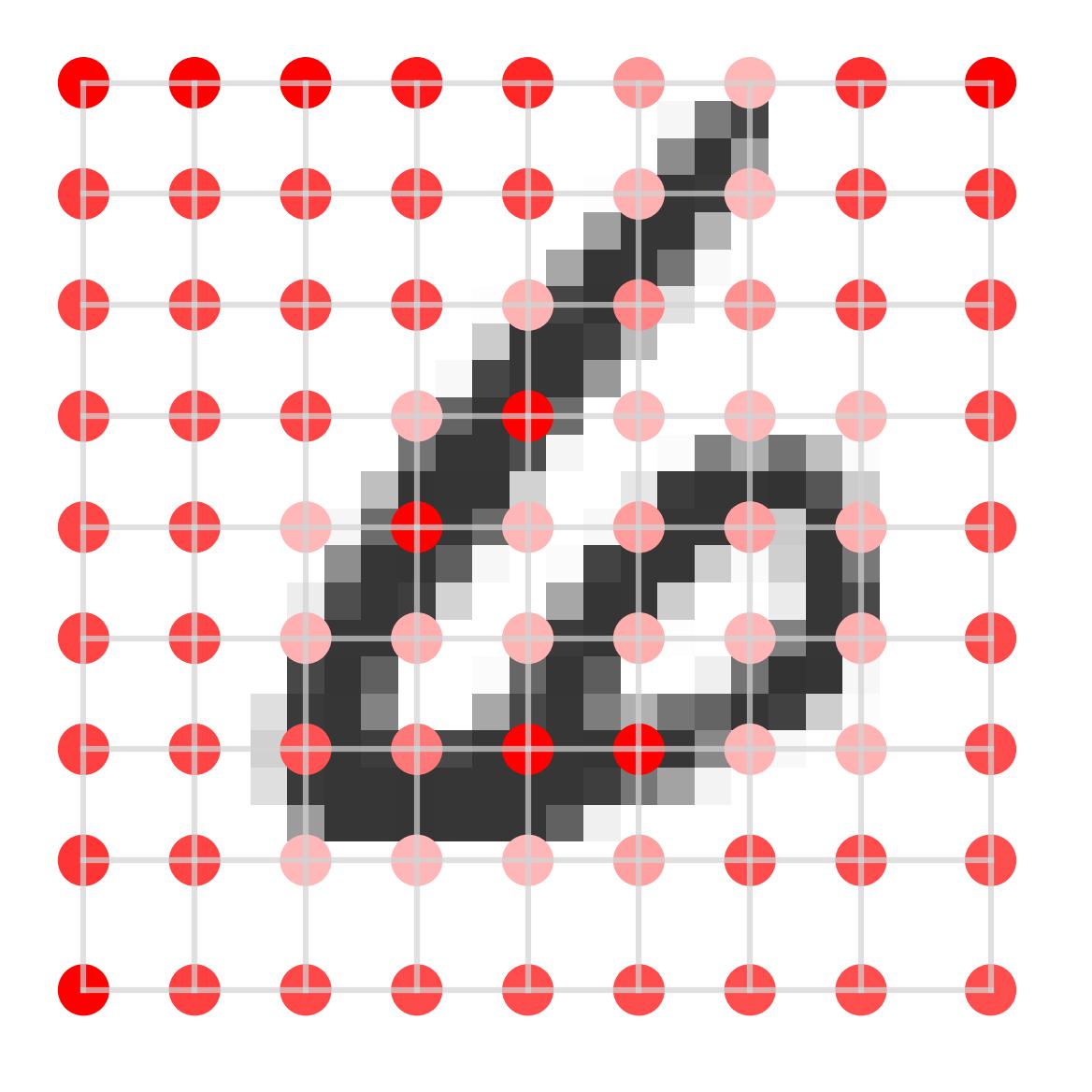}
        &
        \includegraphics[width=0.14\textwidth]{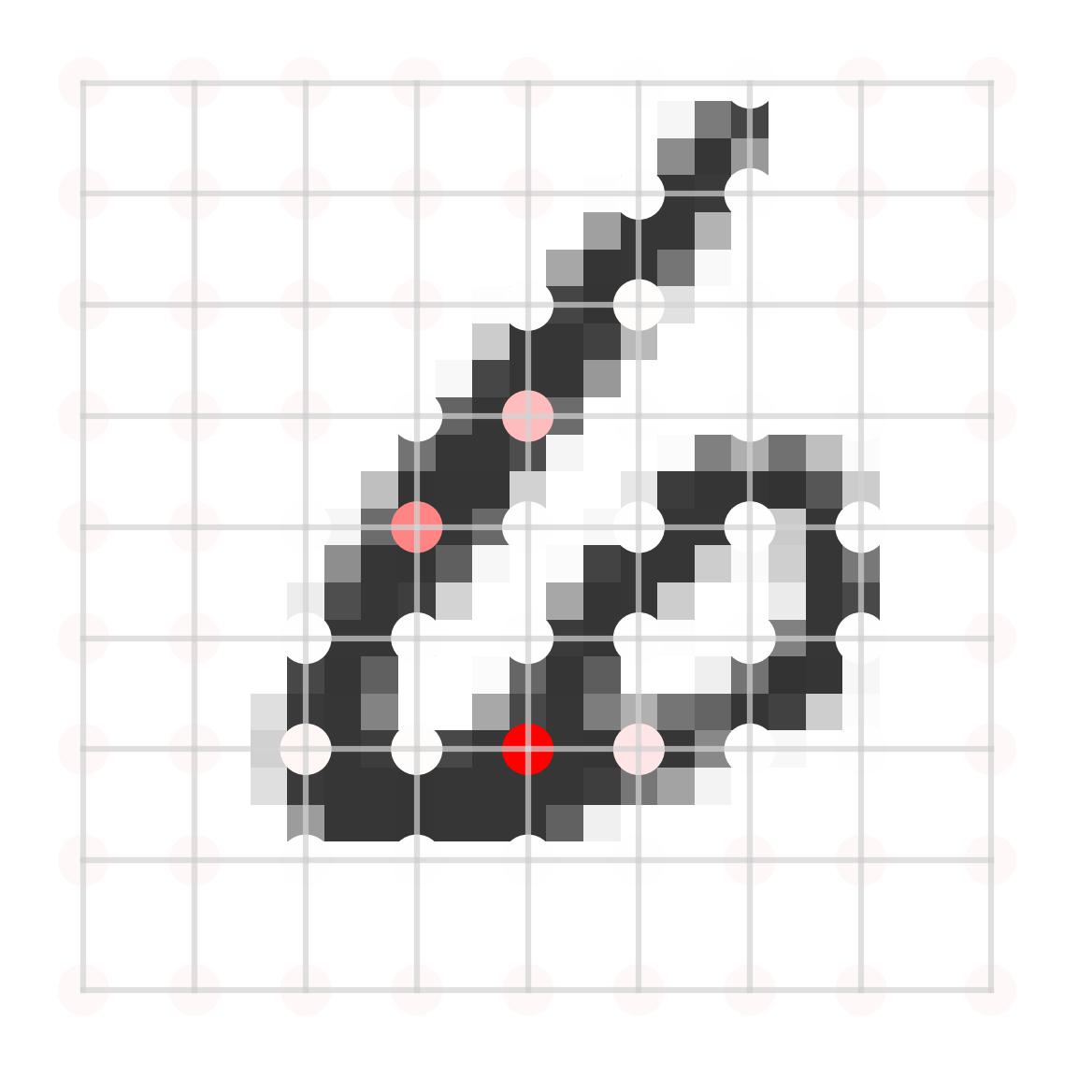}
         \\
         GI &  LRP (AH) & LRP (AH + LN)&   \\
        \includegraphics[width=0.14\textwidth]{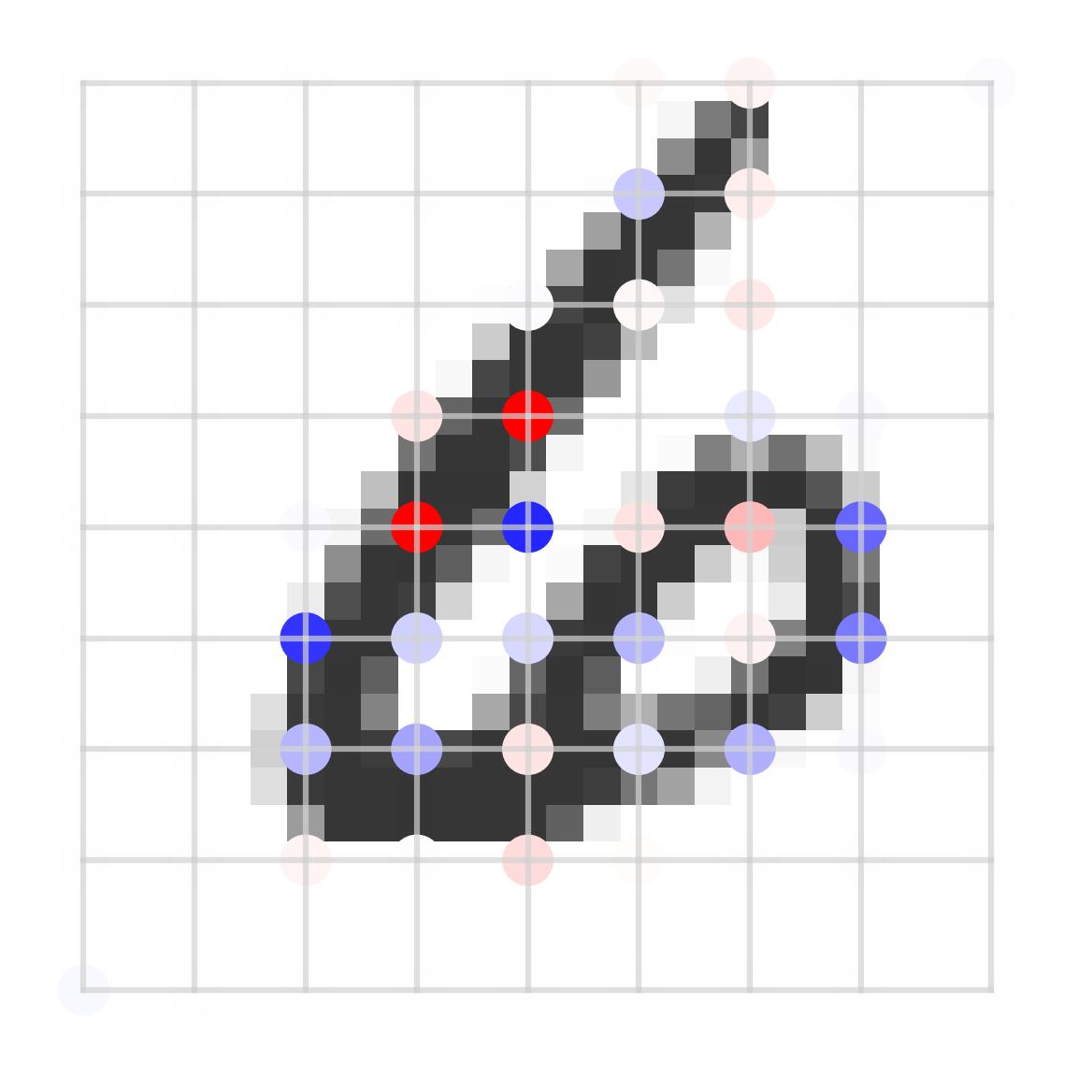}
        &
        \includegraphics[width=0.14\textwidth]{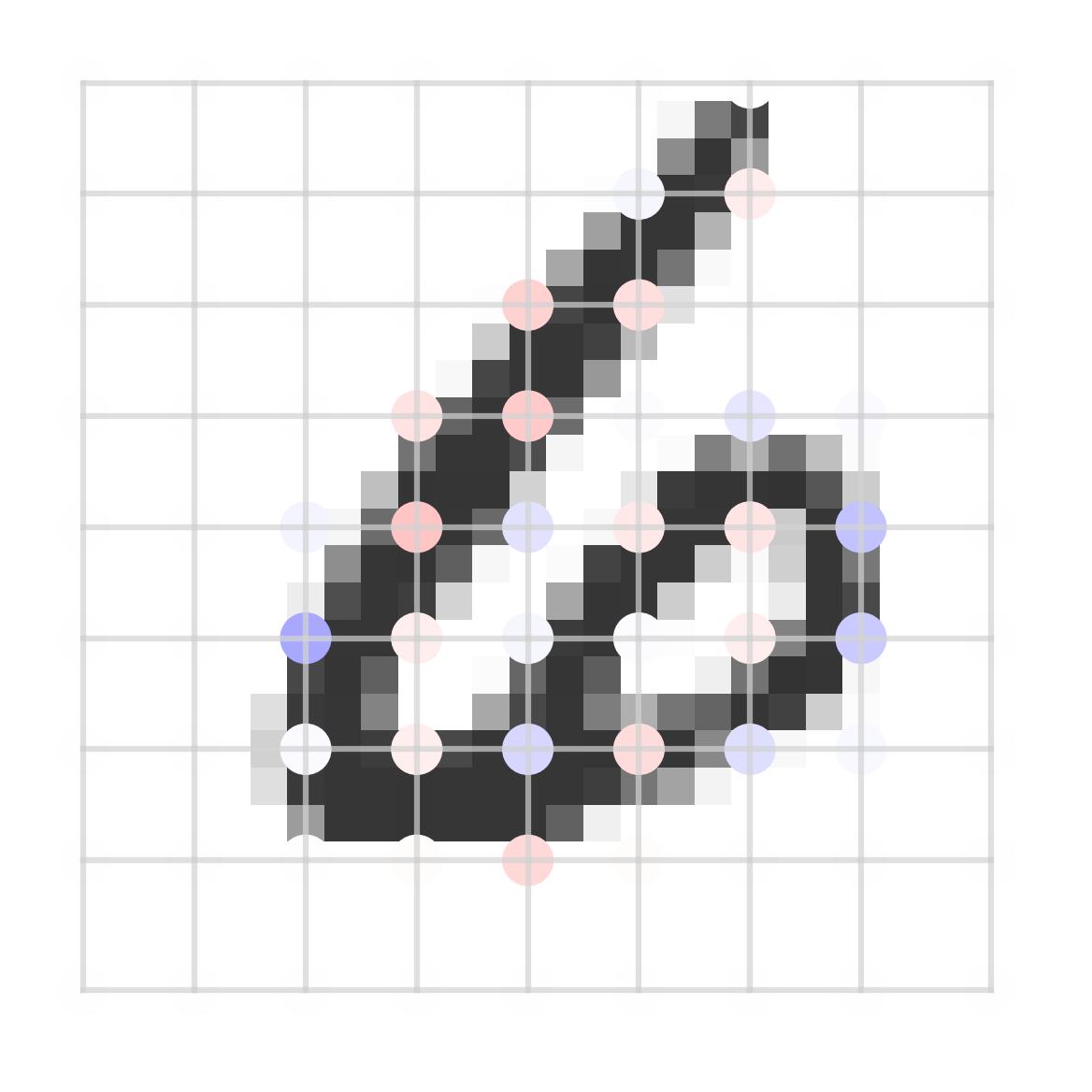}
        &
        \includegraphics[width=0.14\textwidth]{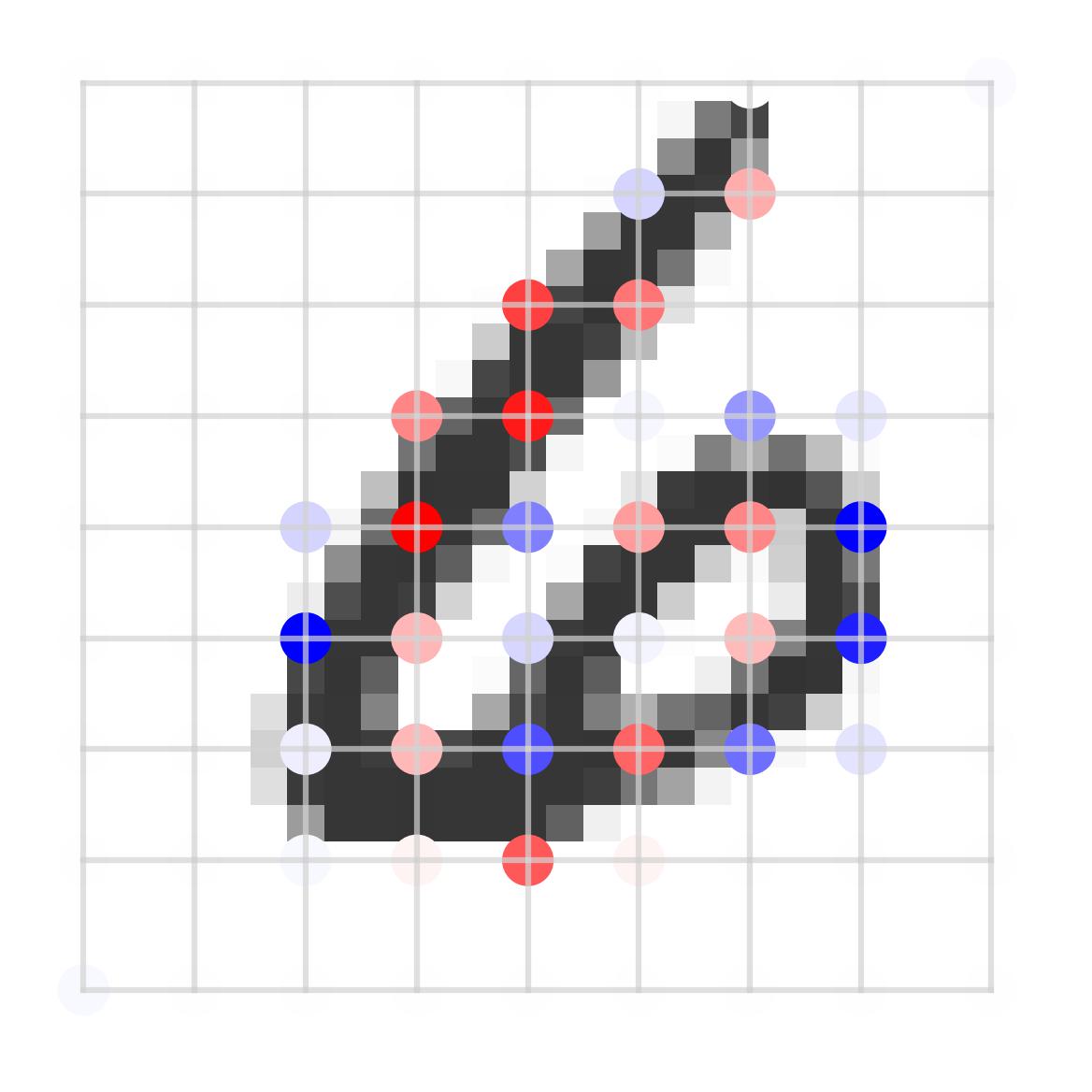}
        &
        \\
        &&& \\
         A-Last & Rollout & GAE \\

        \includegraphics[width=0.14\textwidth]{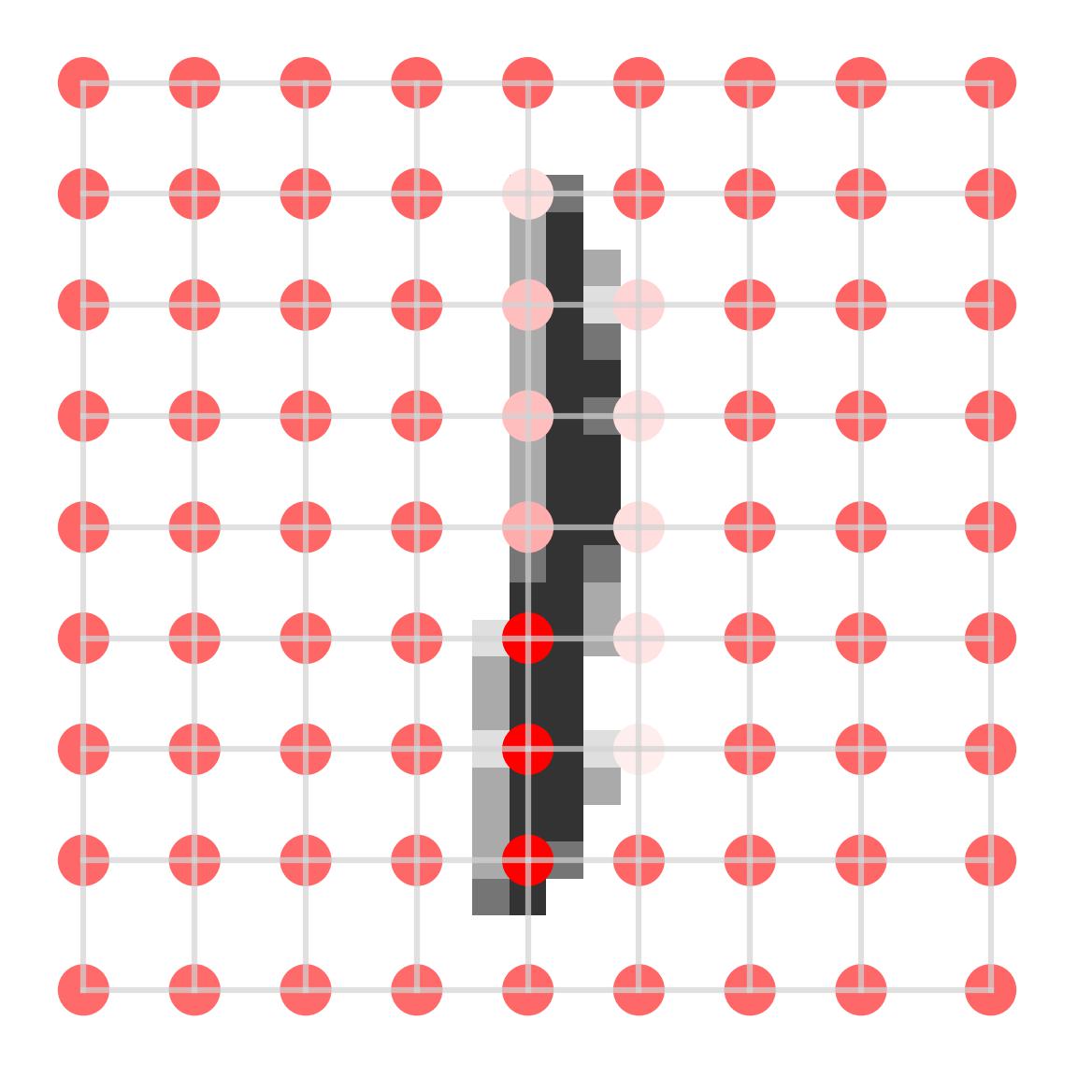}
        &
        \includegraphics[width=0.14\textwidth]{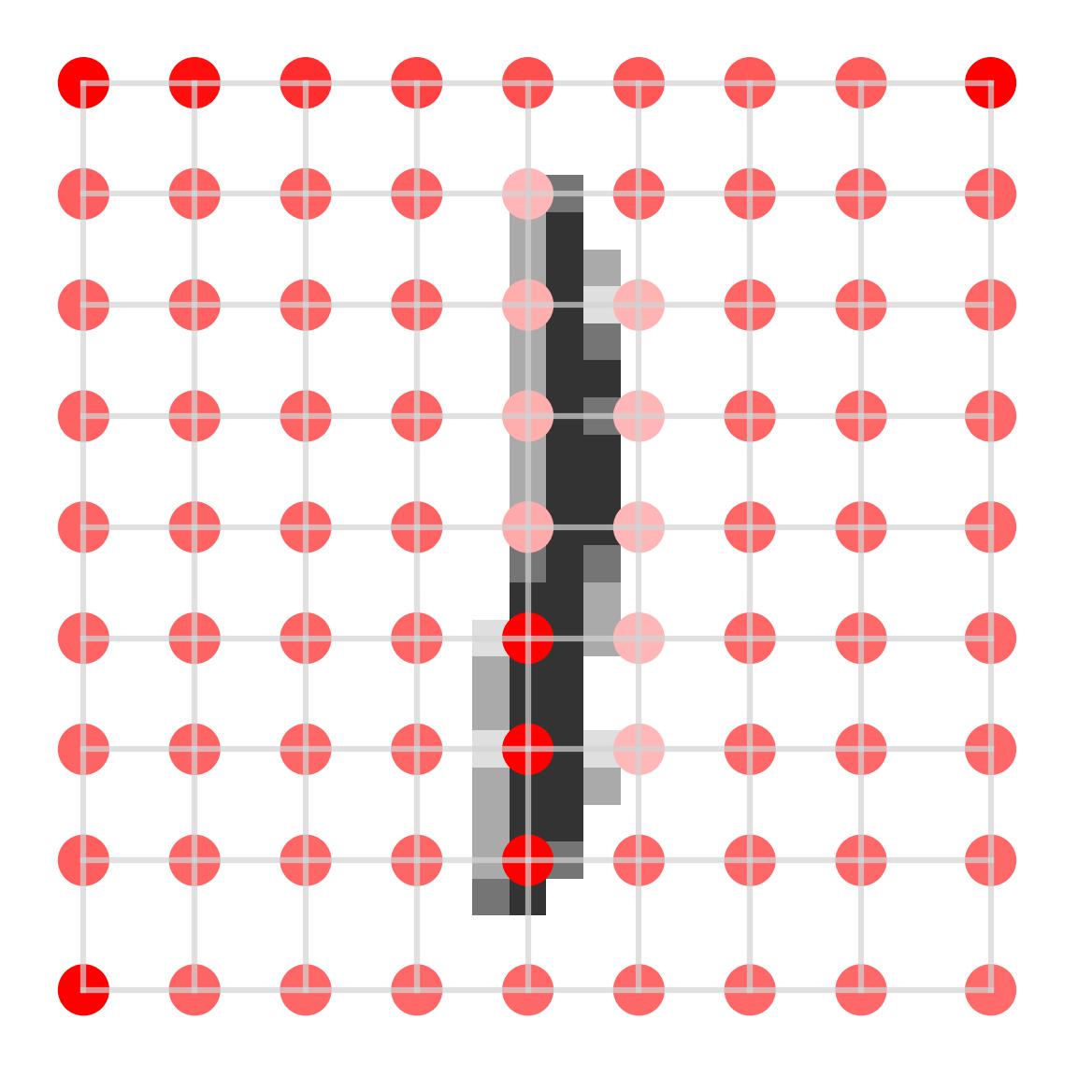}
        &
        \includegraphics[width=0.14\textwidth]{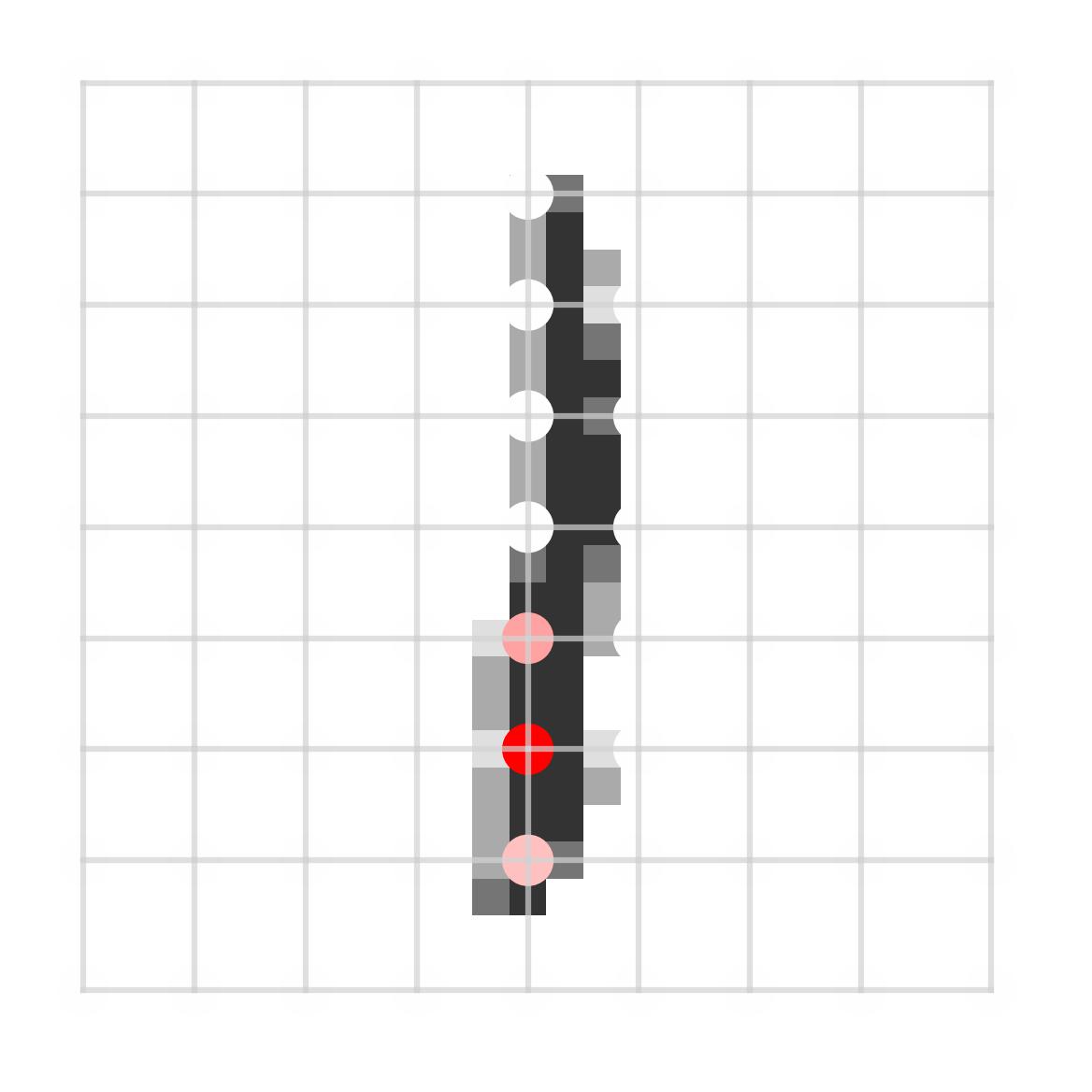}
         \\
         GI &  LRP (AH) & LRP (AH + LN)&   \\
        \includegraphics[width=0.14\textwidth]{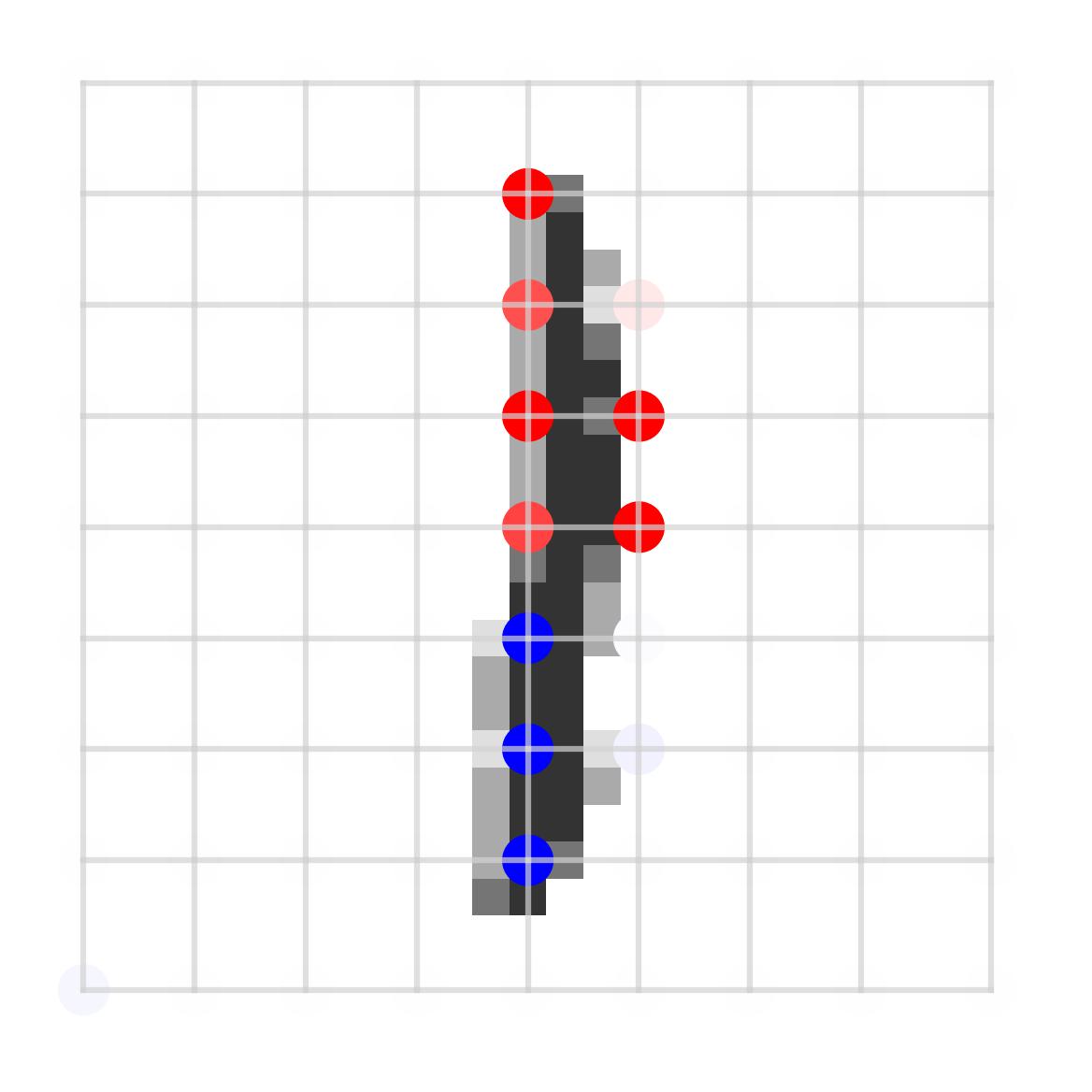}
        &
        \includegraphics[width=0.14\textwidth]{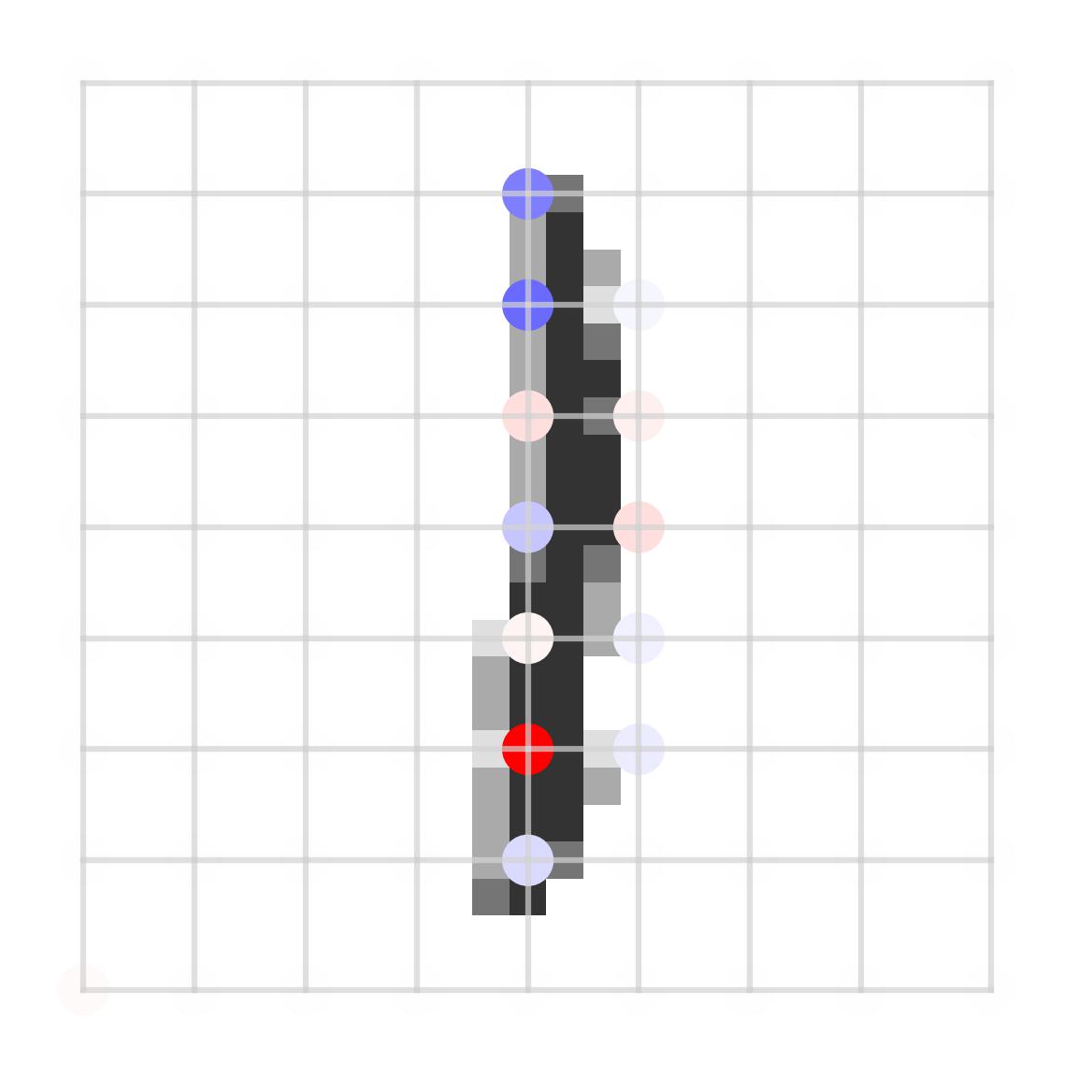}
        &
        \includegraphics[width=0.14\textwidth]{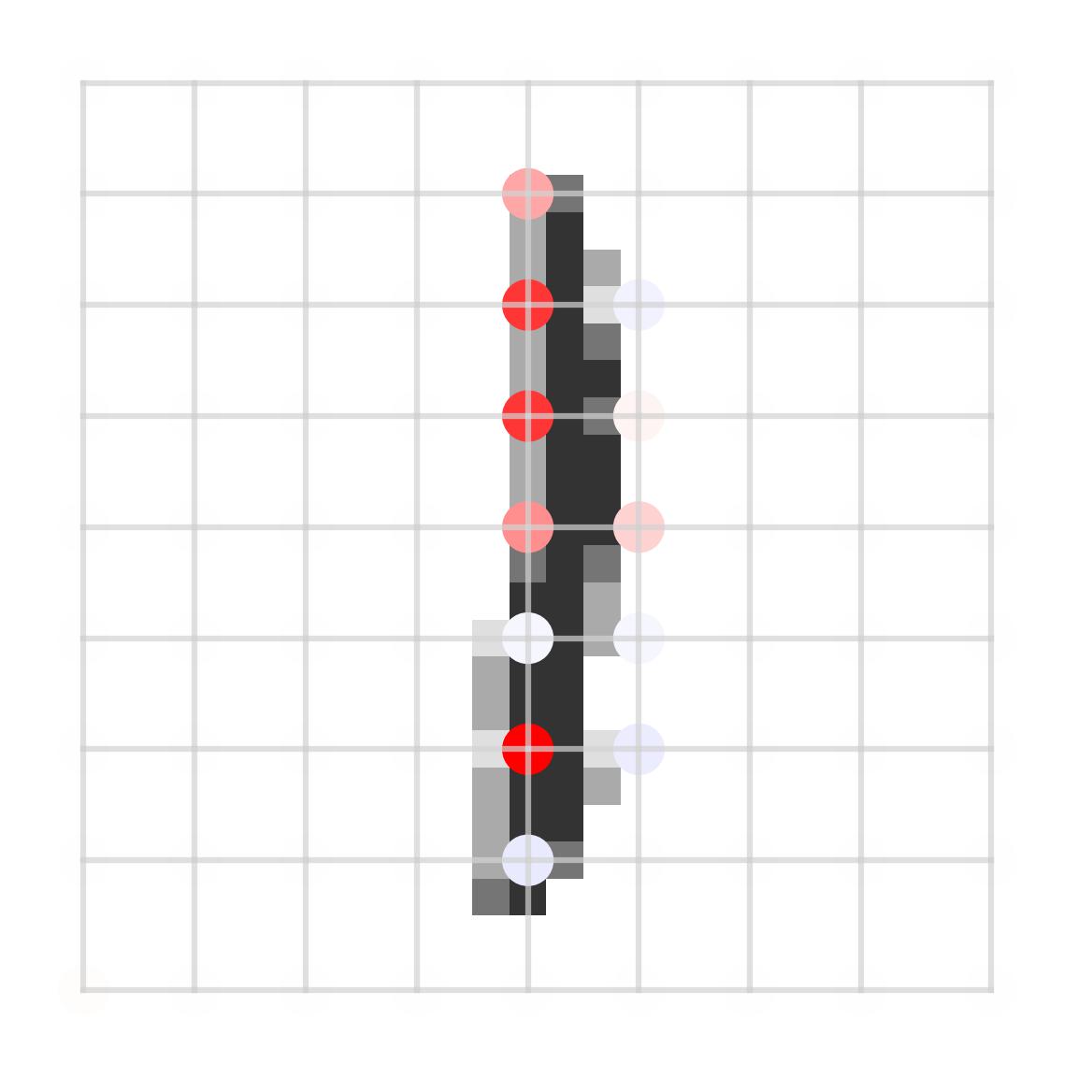}
        &
        \\
          &&& \\
         A-Last & Rollout & GAE \\
        \includegraphics[width=0.14\textwidth]{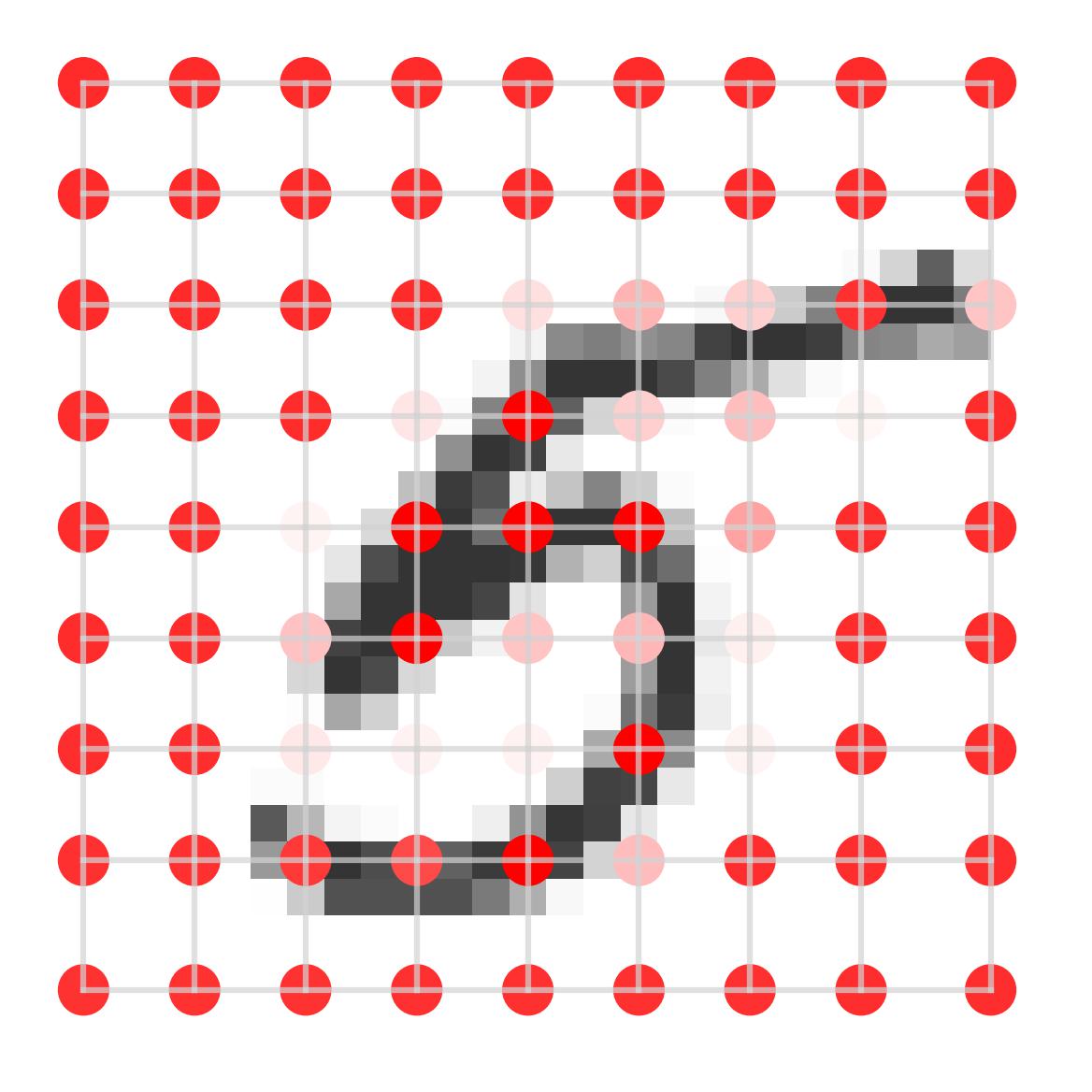}
        &
        \includegraphics[width=0.14\textwidth]{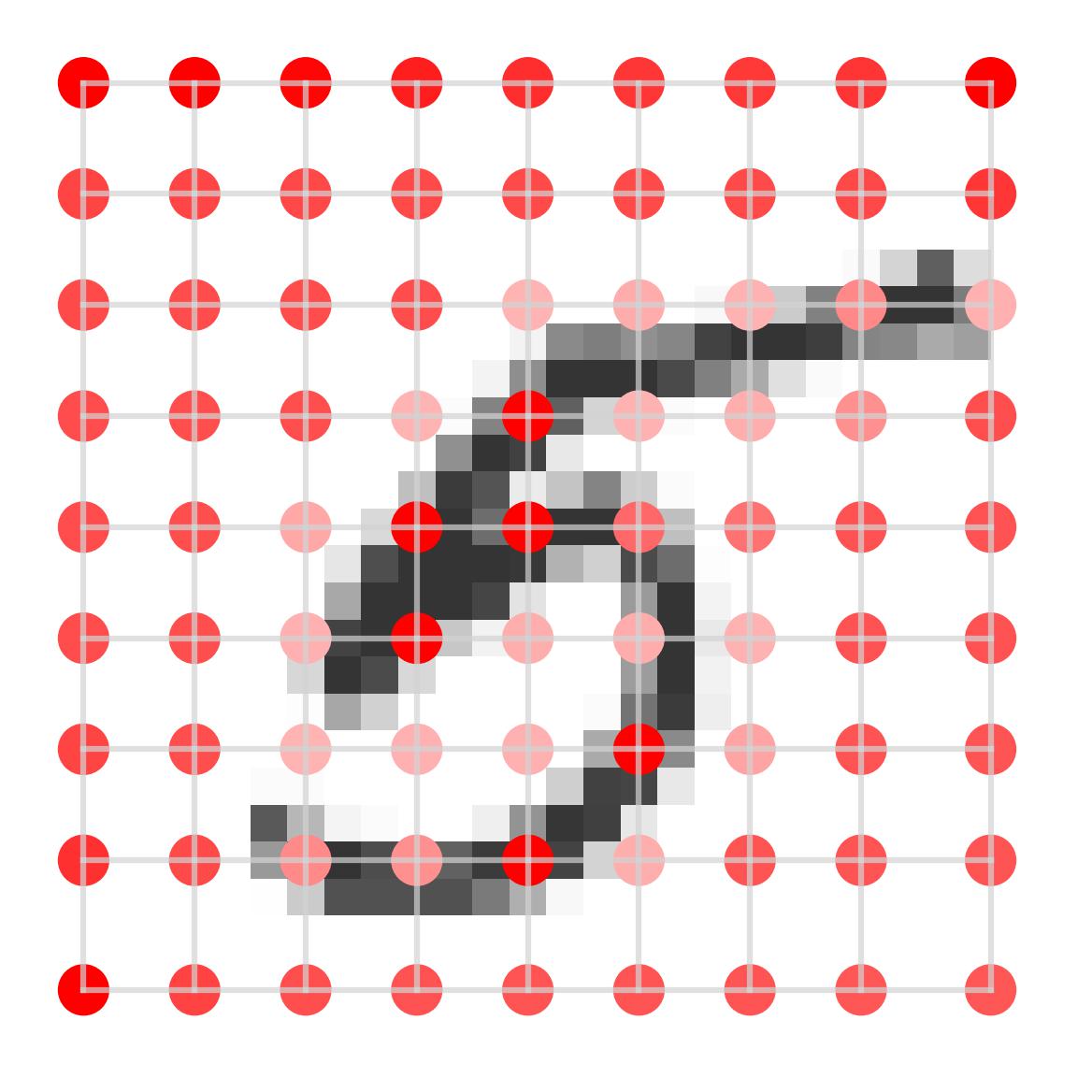}
        &
        \includegraphics[width=0.14\textwidth]{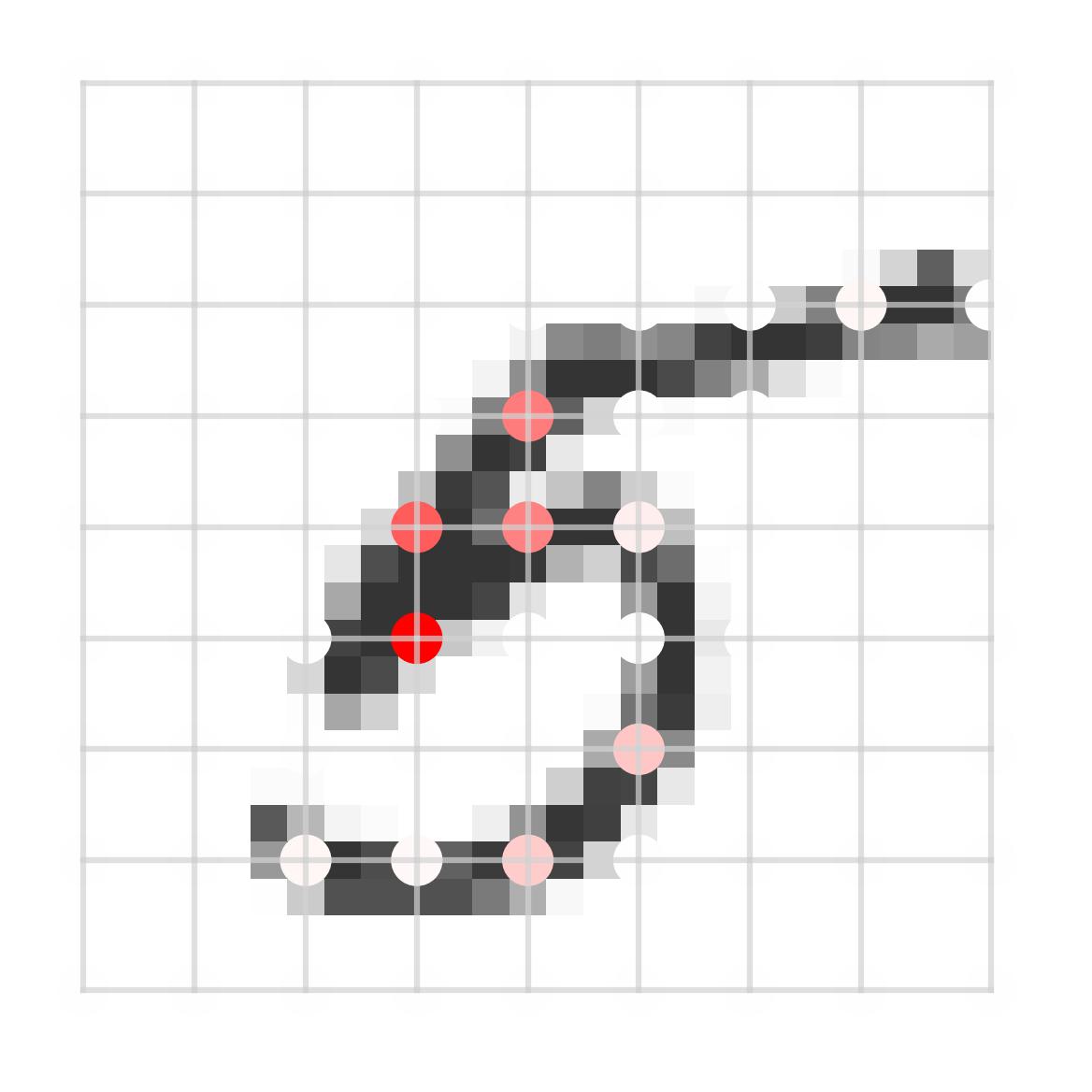}
         \\
         GI &  LRP (AH) & LRP (AH + LN) &  \\
        \includegraphics[width=0.14\textwidth]{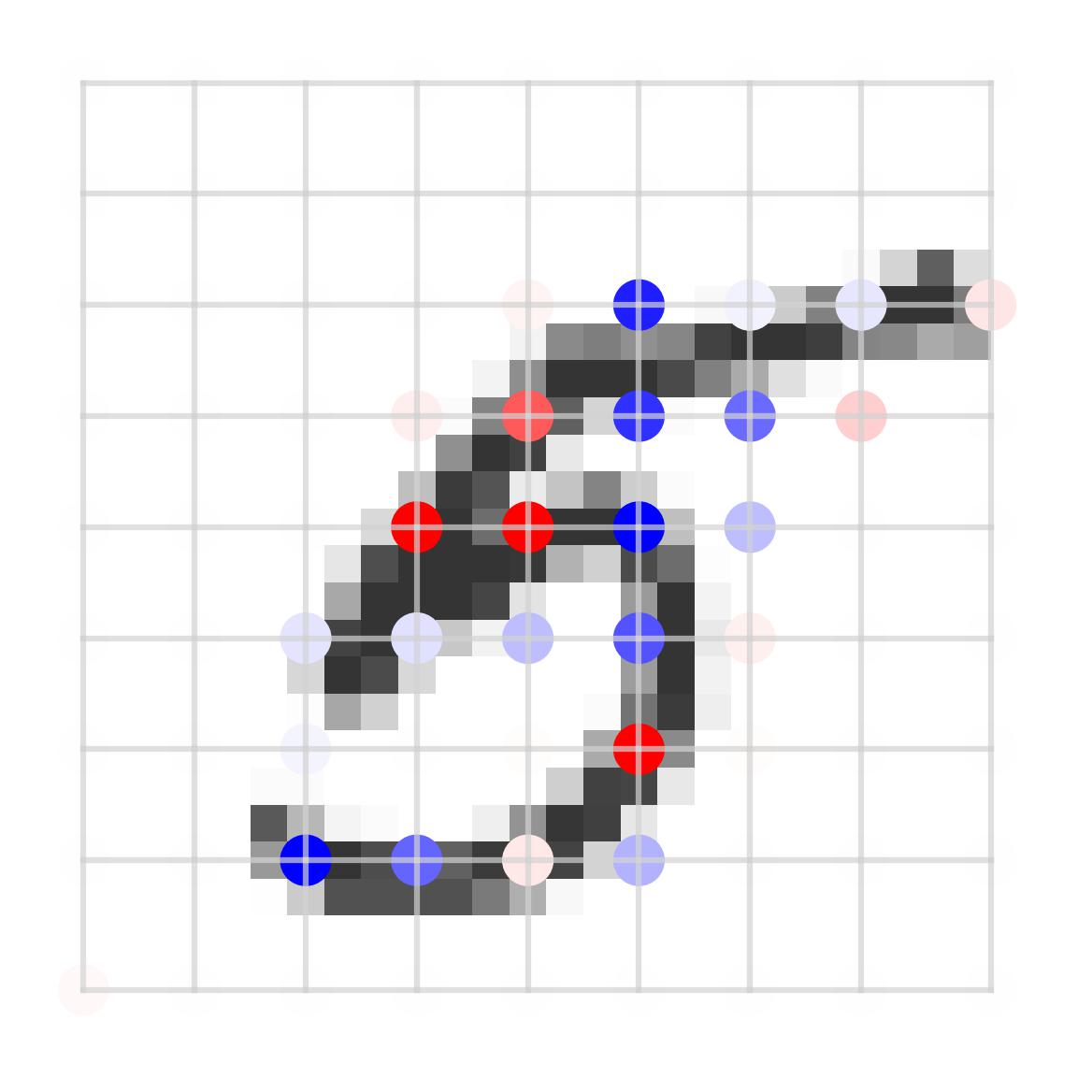}
        &
        \includegraphics[width=0.14\textwidth]{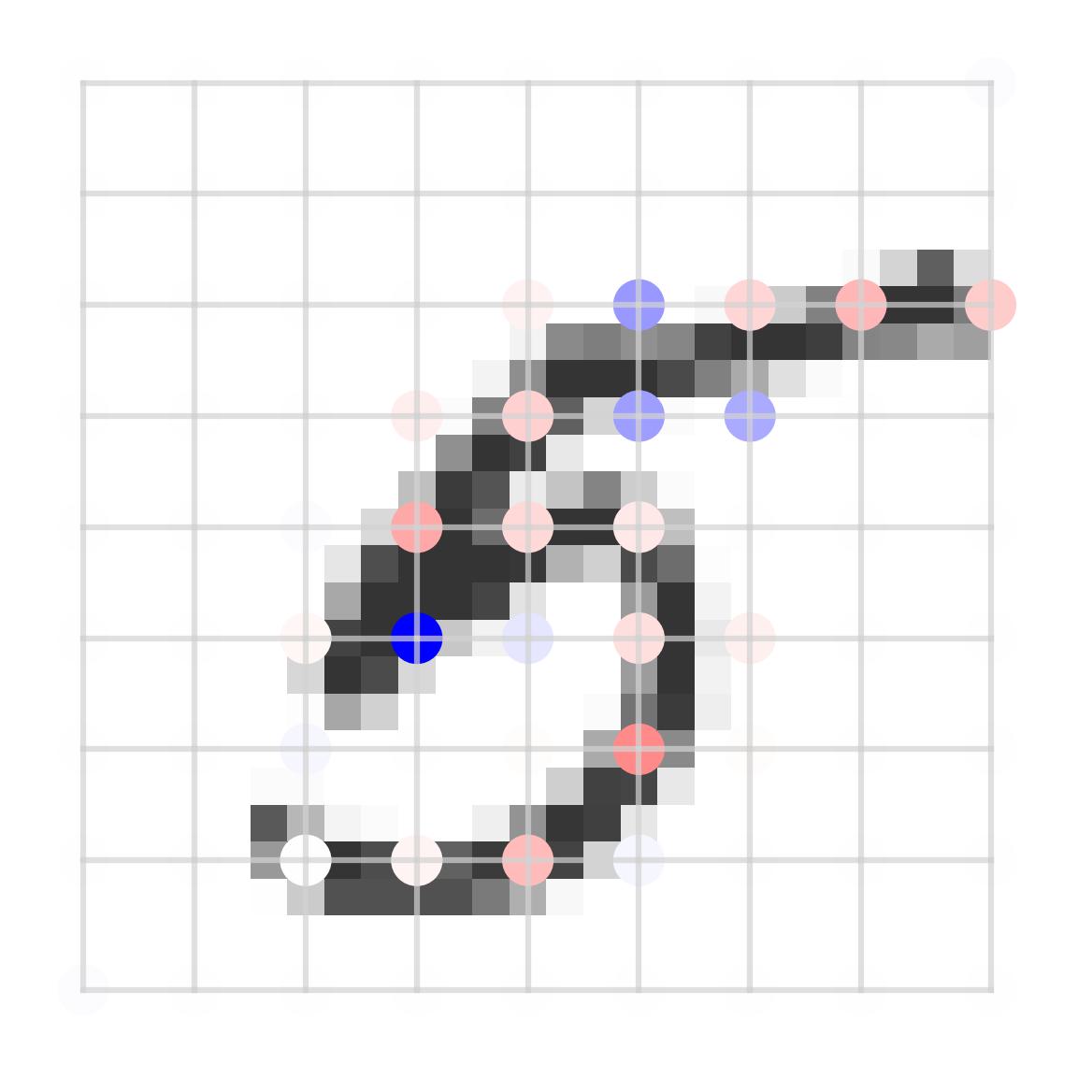}
        &
        \includegraphics[width=0.14\textwidth]{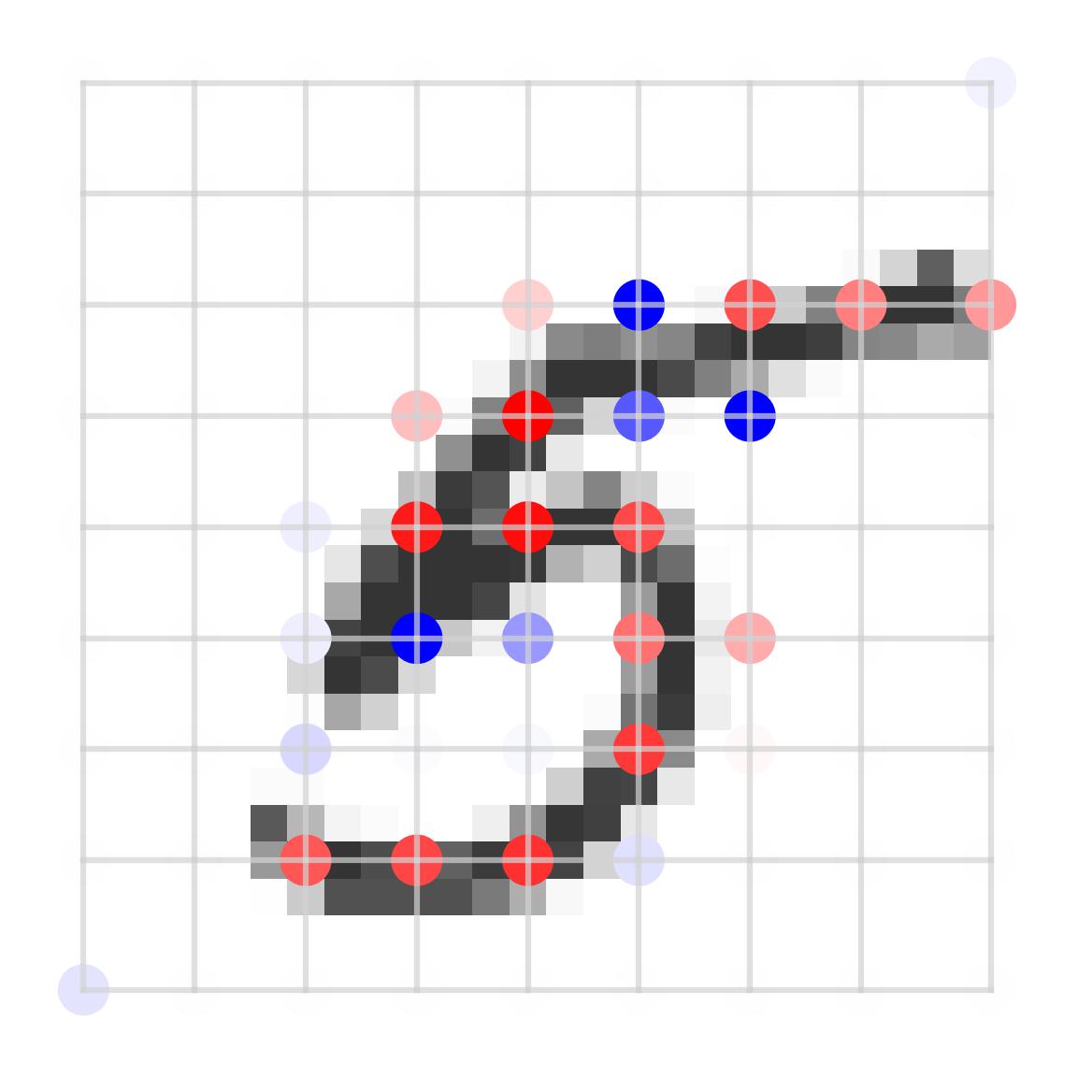} \\
    \end{tabular} \\


\newpage

\twocolumn{
\section{Perturbation Experiments}\label{app:curves}
\begin{tabular}{@{}cc}
 IMDB &  Twitter-Emotions \\
 
 \includegraphics[width=\linewidth]{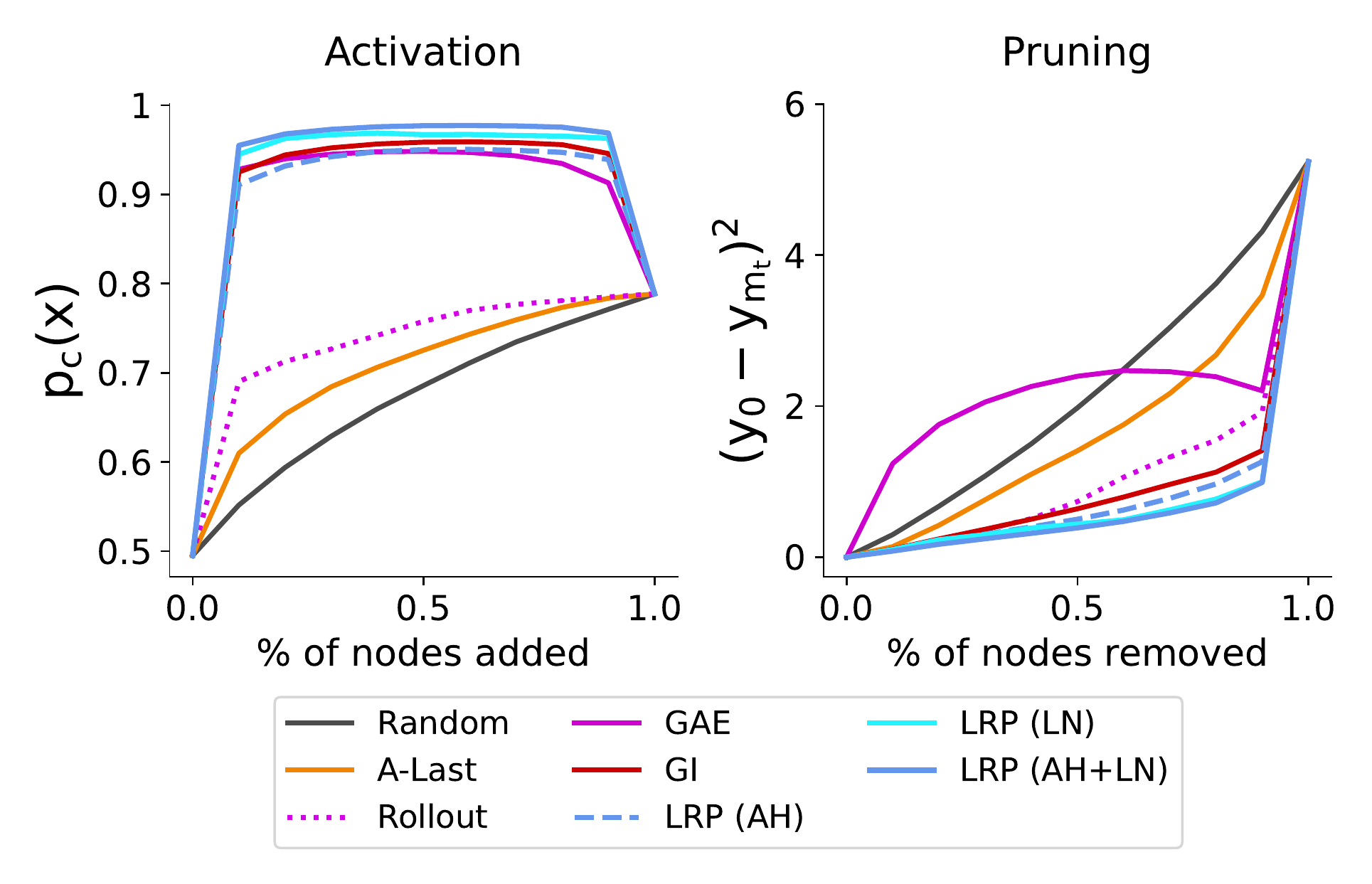} & 
 \includegraphics[width=\linewidth]{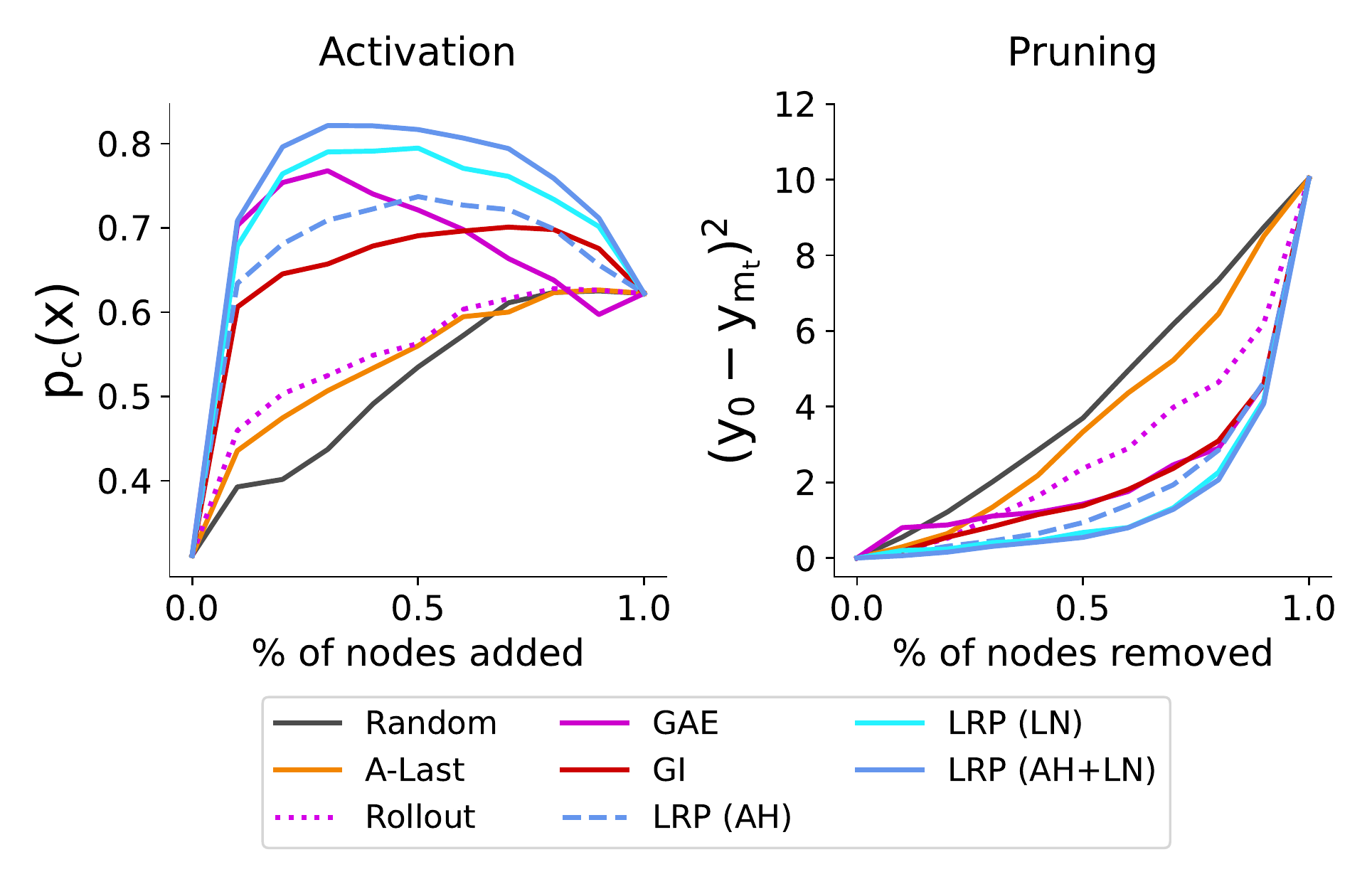} \\
 
Twitter-Sentiment & Twitter-Hate \\
\includegraphics[width=\linewidth]{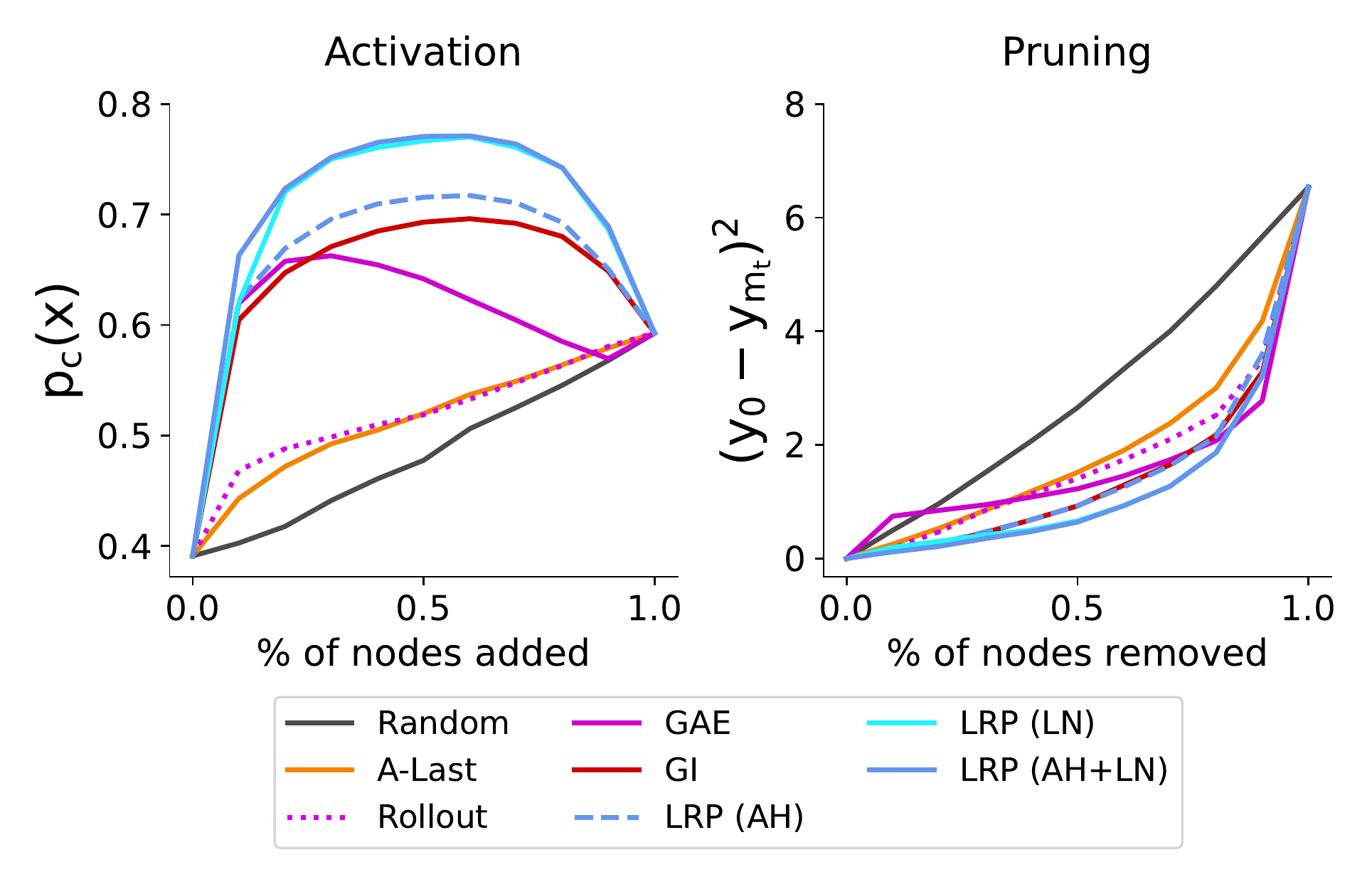} & 
\includegraphics[width=0.5\textwidth]{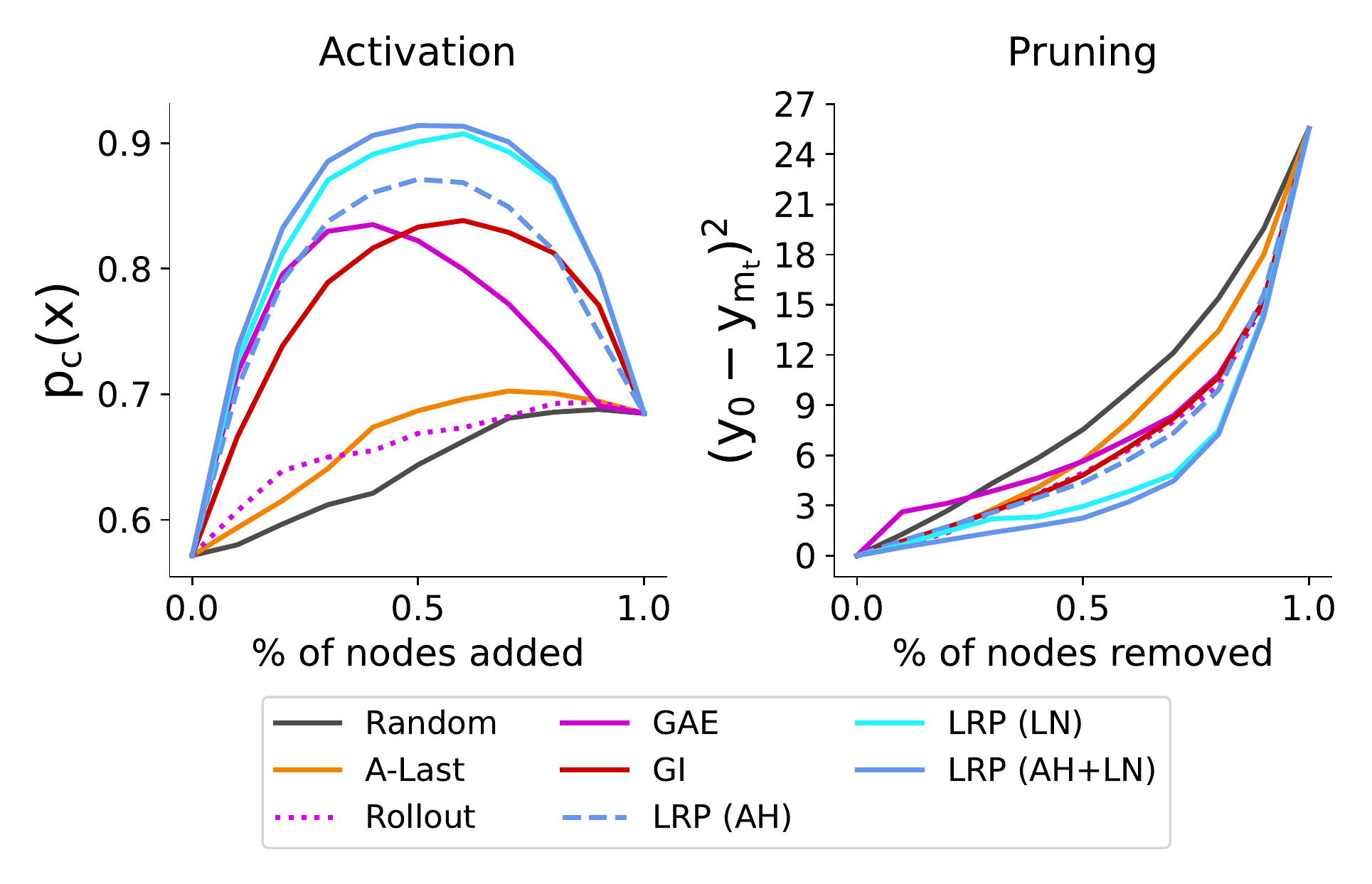} \\
Semaine & Meld-S \\
\includegraphics[width=\linewidth]{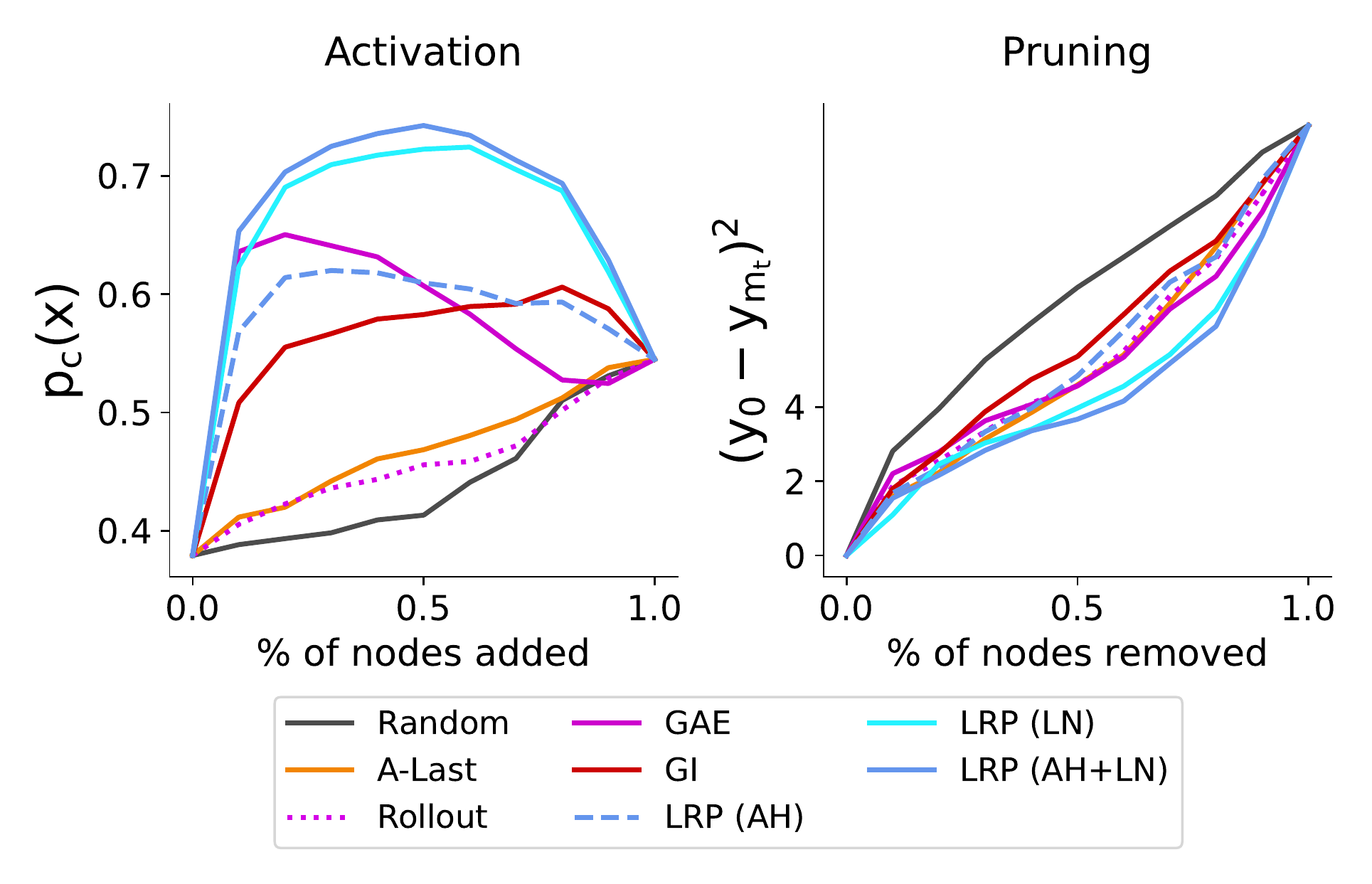} &
\includegraphics[width=\linewidth]{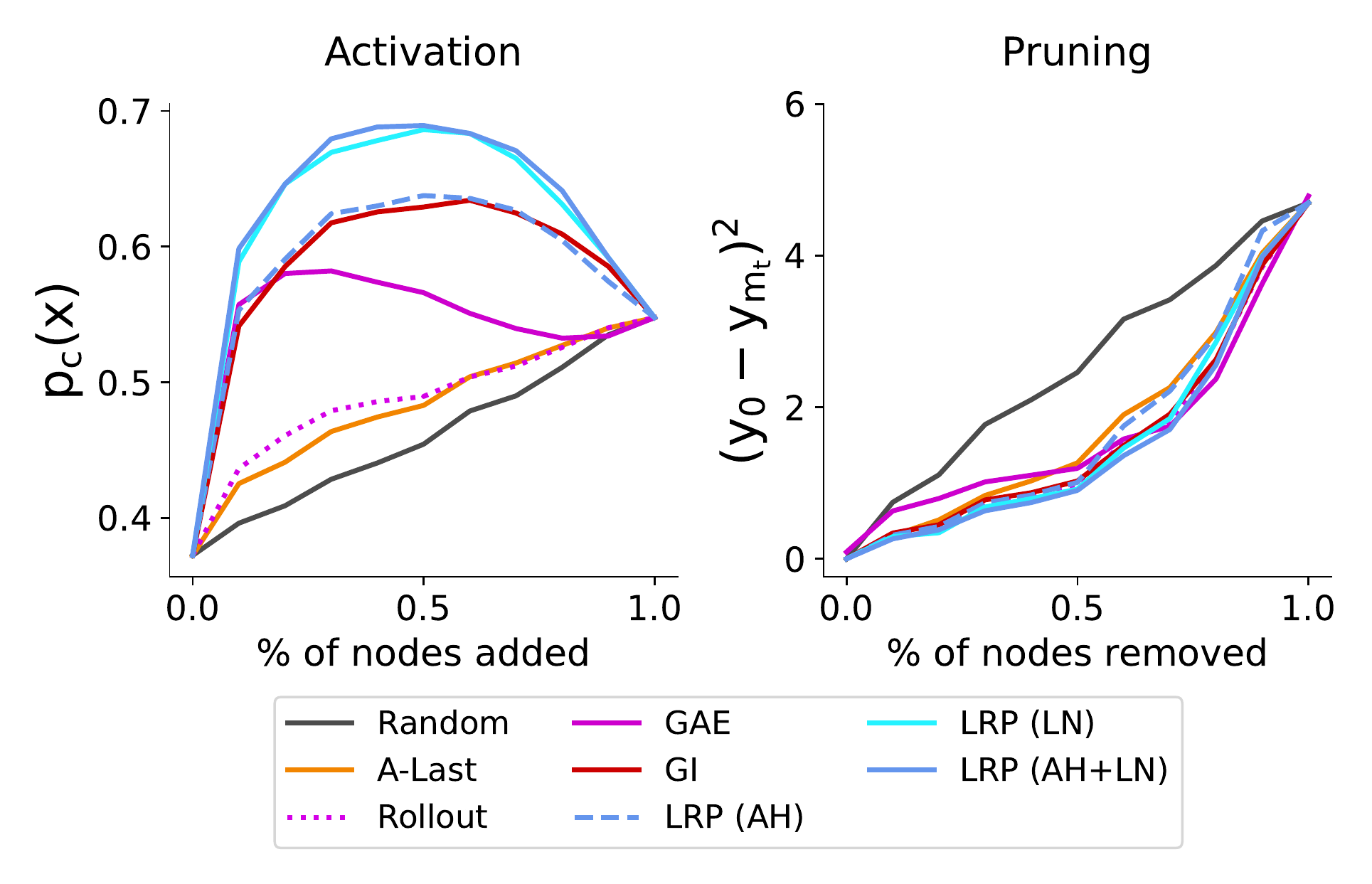}\\
\end{tabular}
}

\onecolumn

\end{document}